\documentclass[10pt]{article} 
\usepackage[accepted]{tmlr}

\usepackage{amsmath,amsfonts,bm}

\def\eqref#1{equation~\ref{#1}}

\def\Algref#1{Algorithm~\ref{#1}}

\def\1{\bm{1}}

\DeclareMathAlphabet{\mathsfit}{\encodingdefault}{\sfdefault}{m}{sl}
\SetMathAlphabet{\mathsfit}{bold}{\encodingdefault}{\sfdefault}{bx}{n}

\def\gA{{\mathcal{A}}}

\def\gD{{\mathcal{D}}}

\def\gT{{\mathcal{T}}}

\def\gW{{\mathcal{W}}}
\def\gX{{\mathcal{X}}}

\def\sN{{\mathbb{N}}}

\def\sP{{\mathbb{P}}}

\def\sR{{\mathbb{R}}}

\newcommand{\E}{\mathbb{E}}

\newcommand{\R}{\mathbb{R}}

\DeclareMathOperator*{\argmax}{arg\,max}

\usepackage{amsmath,amsfonts,bm}

\def\gA{{\mathcal{A}}}

\def\gD{{\mathcal{D}}}

\def\gT{{\mathcal{T}}}

\def\gW{{\mathcal{W}}}
\def\gX{{\mathcal{X}}}

\def\sN{{\mathbb{N}}}

\def\sP{{\mathbb{P}}}

\def\sR{{\mathbb{R}}}

\usepackage{dsfont}
\usepackage{amsmath}
\usepackage{amssymb}
\usepackage{mathtools}
\usepackage{amsthm}
\usepackage{xspace}
\newcommand{\ie}{{\em{i.e.,}}\xspace}

\newcommand{\eg}{{\em{e.g.,}}\xspace}

\newcommand{\wrt}{w.r.t.\xspace}

\newcommand{\ind}{\mathds{1}}
\def\1{\mathbf 1}
\usepackage{subfig}
\usepackage{multirow}
\usepackage{caption}
\usepackage{hyperref}
\usepackage{cleveref}
\usepackage{url}
\usepackage{algorithm}
\let\classAND\AND
\let\AND\relax
\usepackage{algorithmic}

\let\AND\classAND
\AtBeginEnvironment{algorithmic}{\let\AND\algoAND}
\newcommand{\CIFAR}{\texttt{CIFAR-10}\xspace}

\usepackage{booktabs} 
\usepackage{array}
\usepackage{colortbl}
\title{Identify Ambiguous Tasks Combining Crowdsourced Labels \\ by Weighting Areas
Under the Margin}

\author{\name Tanguy Lefort \email tanguy.lefort@umontpellier.fr \\
      \addr IMAG, Univ. Montpellier, CNRS, LIRMM, INRIA
      \AND
      \name Benjamin Charlier \email benjamin.charlier@umontpellier.fr \\
      \addr IMAG, Univ. Montpellier, CNRS
      \AND
      \name Alexis Joly \email alexis.joly@inria.fr\\
      \addr LIRMM, INRIA
      \AND
    \name Joseph Salmon \email joseph.salmon@umontpellier.fr \\
    \addr IMAG, Univ. Montpellier, CNRS, IUF
      }

\begin{document}

\maketitle

\begin{abstract}
In supervised learning — for instance in image classification — modern massive datasets are commonly labeled by a crowd of workers. The obtained labels in this crowdsourcing setting are then aggregated for training, generally leveraging a per-worker trust score.
Yet, such workers oriented approaches discard the tasks' ambiguity.
Ambiguous tasks might fool expert workers, which is often harmful for the learning step.
In standard supervised learning settings -- with one label per task -- the Area Under the Margin (AUM) was tailored to identify mislabeled data.
We adapt the AUM to identify ambiguous tasks in crowdsourced learning scenarios, introducing the Weighted Areas Under the Margin (WAUM). 
The WAUM is an average of AUMs weighted according to task-dependent scores. 
We show that the WAUM can help discarding ambiguous tasks from the training set, leading to better generalization performance.
We report improvements over existing strategies for learning with a crowd, both on simulated settings, and on real datasets such as \texttt{CIFAR-10H} (a crowdsourced dataset with a high number of answered labels), \texttt{LabelMe} and \texttt{Music} (two datasets with few answered votes).
\end{abstract}

\section{Introduction}

\begin{figure}[t!]
    \centering
    \includegraphics[width=.9\textwidth]{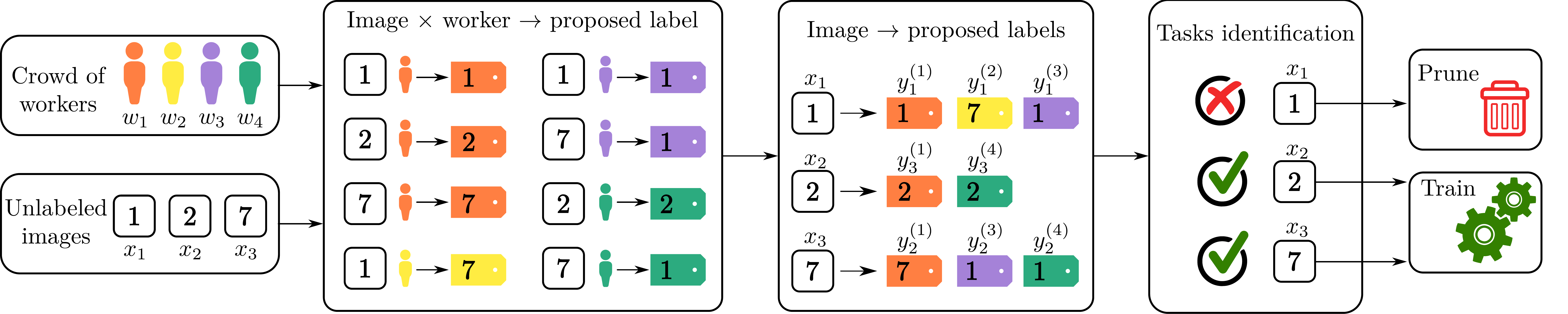}
    \caption{Learning with crowdsourcing labels: from label collection with a crowd to training on a pruned dataset. High ambiguity from either crowd workers or tasks intrinsic difficulty can lead to mislabeled data and harm generalization performance. To illustrate our notation, here the set of tasks annotated by worker $w_3$ is $\gT(w_3)=\{1,3\}$ while the set of workers annotating task $x_3$ is $\gA(x_3)=\{1,3,4\}$.}
    \label{fig:crowdsourcing_scheme}
\end{figure}

Crowdsourcing labels for supervised learning has become quite common in the last two decades, notably for image classification datasets.
Using a crowd of workers is fast, simple (see \Cref{fig:crowdsourcing_scheme}) and less expensive than using experts.
Furthermore, aggregating crowdsourced labels instead of working directly with a single one enables modeling the sources of possible ambiguities and directly taking them into account at training \citep{aitchison2020statistical}.
With deep neural networks nowadays common in many applications, both the architectures and data quality have a direct impact on the model performance \citep{muller2019does, northcutt_pervasive_2021} and on calibration \citep{guo_calibration_2017}.
Yet, depending on the crowd and platform's control mechanisms, the quality of the labels might be low, with possibly many mislabeled instances \citep{muller2019}, hence, degrading generalization \citep{snow_cheap_2008}.

Popular label aggregation schemes take into account the uncertainty related to workers' abilities: for example by estimating confusions between classes, or using a latent variable representing each worker trust \citep{dawid_maximum_1979, pmlr-v22-kim12, sinha2018fast, camilleri2019extended}.
This leads to scoring workers without taking into account the inherent difficulty of the tasks at stake.
Inspired by the Item Response Theory (IRT) introduced in \citep{birnbaum1968some}, the authors of \citep{whitehill_whose_2009} have combined both the task difficulty and the worker's ability in a feature-blind fashion for label aggregation.
Other feature-blind aggregation strategies exist using (rank-one) matrix completion techniques \citep{ma2020adversarial,ma2020gradient} or pairwise co-occurrences \citep{ibrahim2019crowdsourcing}. Both rely on the work by \citet{dawid_maximum_1979} and take into account worker abilities but neglect the task difficulty.
All the feature-blind strategies only leverage the labels but discard the associated features to evaluate workers performance.
For instance, GLAD \citep{whitehill_whose_2009} estimates a task difficulty without the actual task: its estimation only relies on the collected labels and not on the tasks themselves (in image-classification settings, this means the images are not considered for evaluating the task difficulty). Neglecting such task difficulty might become critical when the number of labels collected per task is small.


\begin{figure}[t]
    \centering
    \hfill
    \subfloat[Label \texttt{airplane} is easy to identify (unanimity among workers).]{\includegraphics[width=0.3083\textwidth]{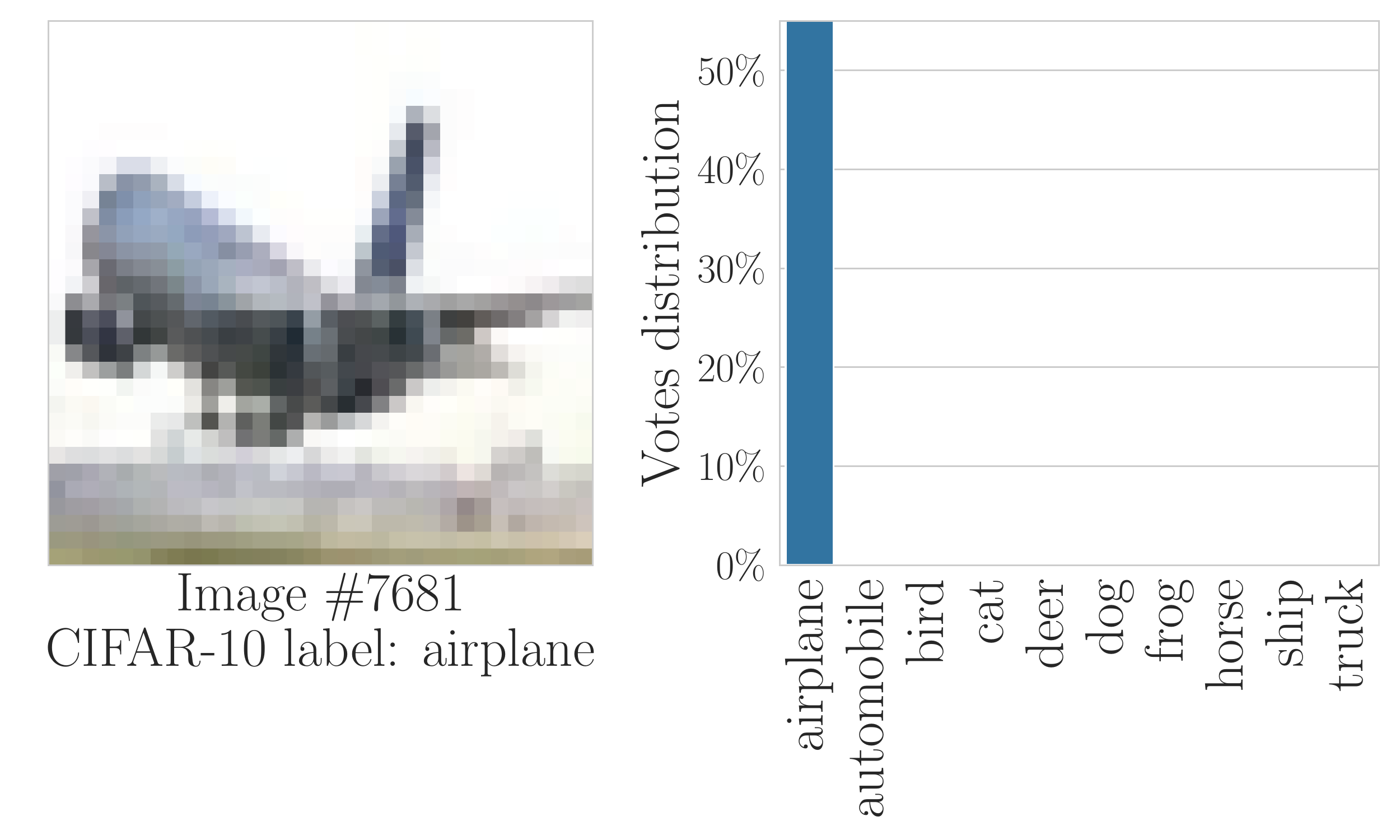}}
    \hfill
    \subfloat[Label \texttt{deer} is meaningless here, and workers are confused with all other labels.]{\includegraphics[width=0.3083\textwidth]{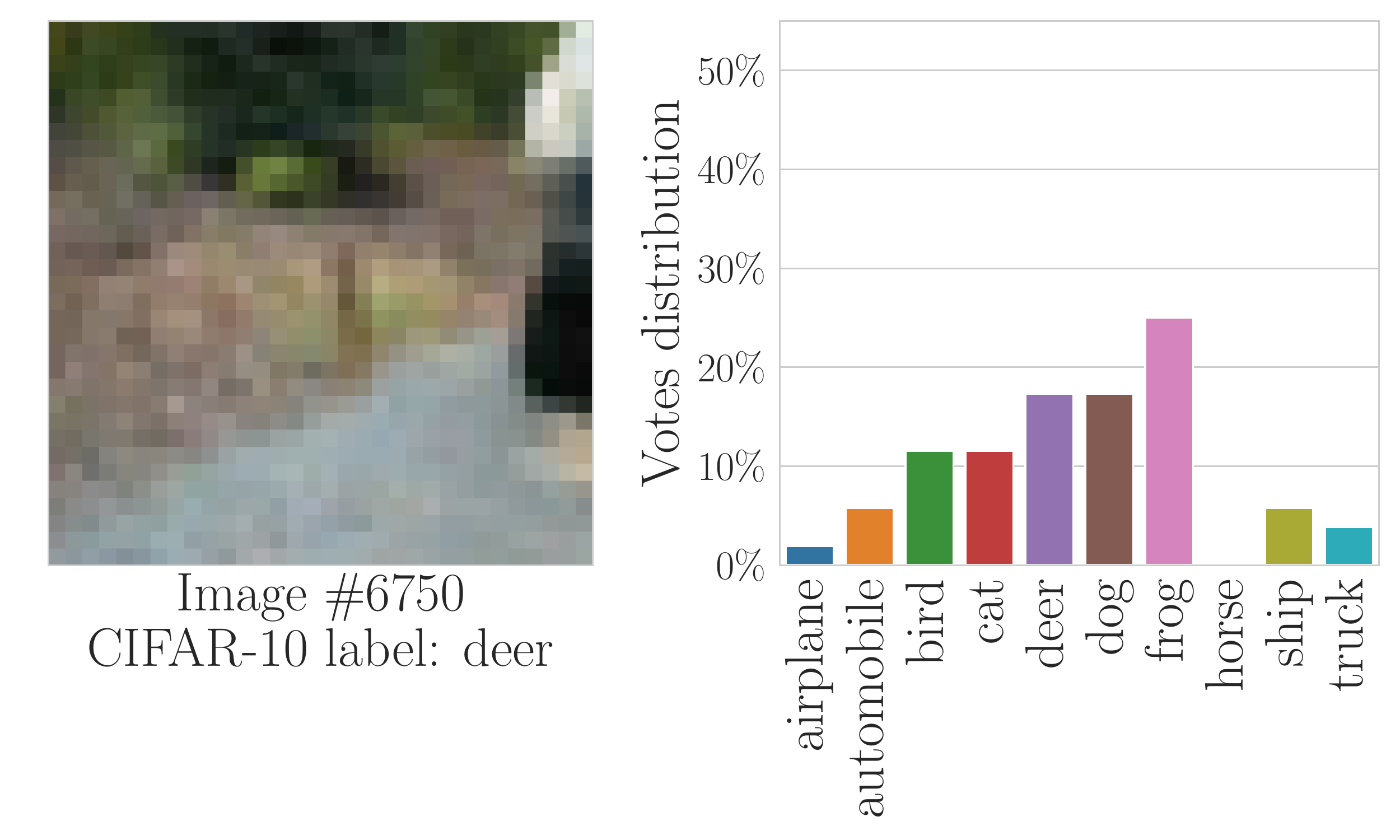}\label{subfig:example_deer}}
    \hfill
    \subfloat[Label \texttt{cat} often confused with horns of a wild \texttt{deer}]{\includegraphics[width=0.3083\textwidth]{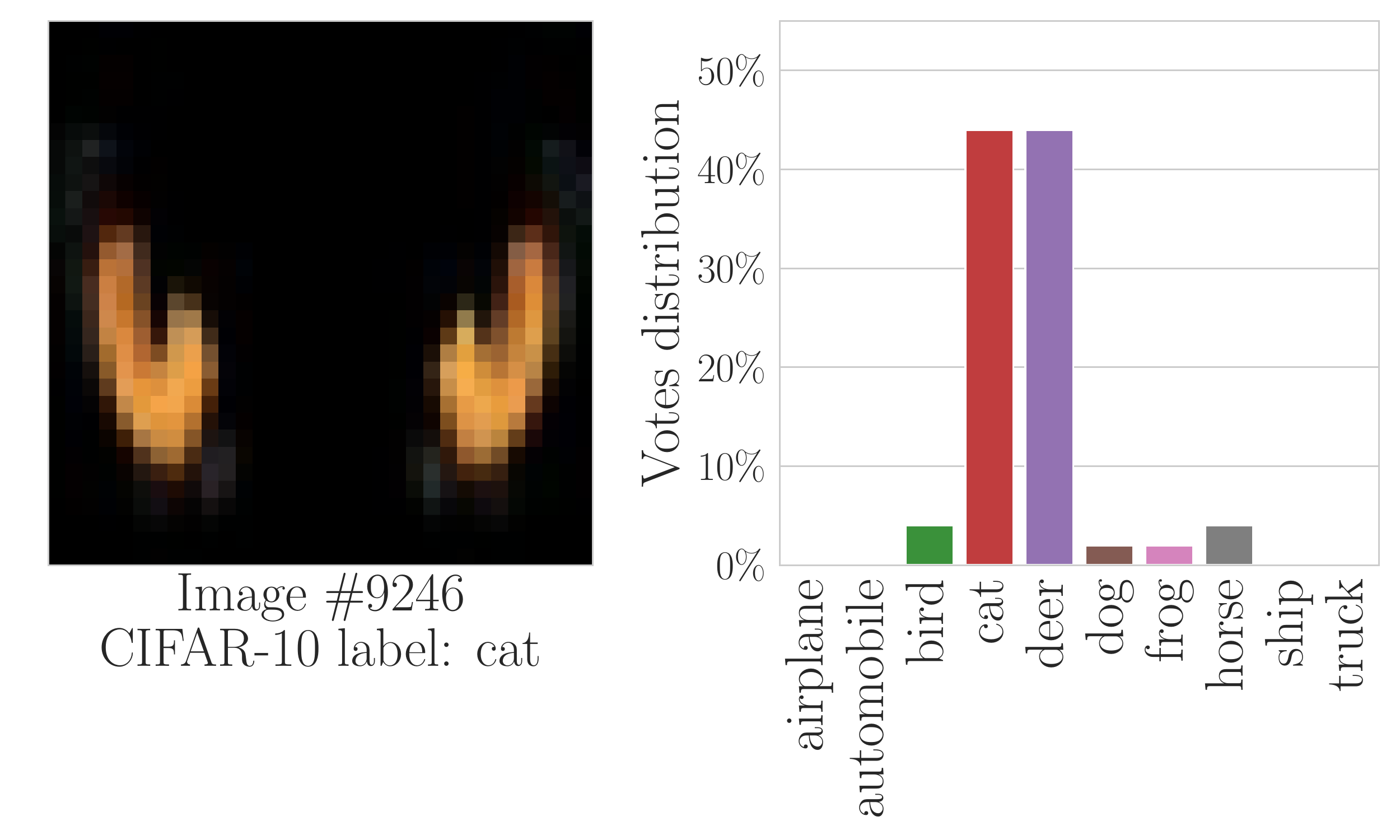}\label{subfig:example_cat}}
    \hfill
    \caption{Three images from \texttt{CIFAR-10H} dataset \citep{peterson_human_2019}, with the empirical distribution of workers' labels (soft labels): the \texttt{airplane} image (a) is easy, while the landscape (b) is ambiguous due to the image's poor quality. The last image (c) looks like a black cat face often perceived as the horns of a \texttt{deer}.
    }
    \label{fig:example}%
\end{figure}

In this work, we aim at identifying ambiguous tasks from their associated features, hence discarding hurtful tasks (such as the ones illustrated on \Cref{subfig:example_deer} and \Cref{subfig:example_cat}).
Recent works on data-cleaning in supervised learning \citep{han2019deep, pleiss_identifying_2020, northcutt_confident_2021} have shown that some images might be too corrupted or too ambiguous to be labeled by humans.
Hence, one should not consider these tasks for label aggregation or learning since they might reduce generalization power; see for instance \citep{pleiss_identifying_2020}.
Throughout this work, we consider the ambiguity of a task with the informal definition proposed by \citet{angelova2004data} that fit standard learning frameworks: \emph{``Difficult examples are those which obstruct the learning process or mislead the learning algorithm or those which are impossible to reconcile with the rest of the examples''}.
This definition links back to with how \citet{pleiss_identifying_2020} detect corrupted samples using the area under the margin (AUM) during the training steps of a machine learning classifier.
However, it is important to notice that, in this context, the task ambiguity is inherent to the classifier architecture, and thus might not exactly overlap with human-level difficulty.  

In this work, we combine task difficulty scores with worker abilities scores, but we measure the task difficulty by incorporating feature information.
We thus introduce the Weighted Area Under the Margin ($\mathrm{WAUM}$), a generalization to the crowdsourcing setting of the Area Under the Margin ($\mathrm{AUM}$) by \citep{pleiss_identifying_2020}.
The $\mathrm{AUM}$ is a confidence indicator in an assigned label defined for each training task.
It is computed as an average of margins over scores obtained along the learning steps.
The $\mathrm{AUM}$ reflects how a learning procedure struggles to classify a task to an assigned label\footnote{See the Linear SVC in \Cref{fig:threecircles_workers} to visualize how the $\mathrm{AUM}$ is connected to the classical margin from the kernel literature.}.
The $\mathrm{AUM}$ is well suited when training a neural network (where the steps are training epochs) or other iterative methods.
For instance, it has led to better network calibration \citep{park2022calibration} using MixUp strategy \citep{zhang2017mixup}, \ie mixing tasks identified as simple and difficult by the $\mathrm{AUM}$.
The $\mathrm{WAUM}$, our extension of the $\mathrm{AUM}$, aims at identifying harmful data points in crowdsourced datasets, so one can prune ambiguous tasks that degrade the generalization.
It is a weighted average of workers $\mathrm{AUM}$, where the weights reflect trust scores based on task difficulty and workers' ability.

\section{Related Work}

Inferring a learning consensus from a crowd is a challenging task.
In this work, we do not consider methods with prior knowledge on the workers, since most platforms do not provide this information\footnote{For instance, by default Amazon Mechanical Turk \url{https://www.mturk.com/} does not provide it.}.
Likewise, we do not rely on ground-truth knowledge for any tasks.
Hence, trapping-set or control-items-based algorithms like ELICE or CLUBS \citep{khattak_toward_2017} do not match our framework.
Some algorithms rely on self-reported confidence: they directly ask workers their answering confidence and integrate it into the model \citep{albert2012combining, oyama2013accurate, hoang2021tournesol}.
We discard such cases for several reasons.
First, self-reported confidence might not be beneficial without a reject option \citep{li2017does}.
Second, workers have a tendency to be under or overconfident, raising questions on how to present self-evaluation and assessing own scores \citep{draws2021checklist}.

To reach a consensus in the labeling process, the most common aggregation step is majority voting (MV), where one selects the label most often answered.
MV does not infer any trust score on workers and does not leverage workers' abilities.
MV is also very sensitive to under-performing workers \citep{gao2013minimax, zhou2015regularized}, to biased workers \citep{kamar2015identifying}, to spammers \citep{raykar_ranking_2011}, or lack of experts for hard tasks \citep{james1998majority, gao2013minimax, germain2015risk}.
Closely related to MV, naive soft (NS) labeling goes beyond \emph{hard labels} (also referred to as \emph{one-hot labels}) by computing the frequency of answers per label, yield a distribution over labels, often referred to as \emph{soft-labels}.
In practice, training a neural network with soft labels improves calibration \citep{guo_calibration_2017} \wrt using hard labels.
However, both MV and NS are sensitive to spammers (\eg workers answer \emph{all} tasks randomly) or workers' biases (\eg workers who answer \emph{some} tasks randomly).
Hence, the noise induced by workers' labeling might not be representative of the actual task difficulty \citep{jamison2015noise}.

Another class of methods leverages latent variables, defining a probabilistic model on workers' responses.
The most popular one, proposed by \citep{dawid_maximum_1979} (DS), estimates a single confusion matrix per worker, as a measure of workers' expertise.
The underlying model assumes that a worker answers according to a multinomial distribution, yielding a joint estimation procedure of the confusion matrices and the soft labels through Expectation-Maximization (EM).
Variants of the DS algorithm include accelerated \citep{sinha2018fast}, sparse \citep{servajean2017crowdsourcing}, and clustered versions \citep{imamura2018analysis} among others.

Since DS only models workers' abilities, \citep{whitehill_whose_2009} have introduced the Generative model of Labels, Abilities, and Difficulties ($\mathrm{GLAD}$) to exploit task difficulties to improve confusion estimation.
While DS estimates a matrix of pairwise label confusion per worker, $\mathrm{GLAD}$ considers also an EM procedure to estimate one ability score per worker, and one difficulty score per task.
It is inspired by the IRT \citep{birnbaum1968some}, modeling the workers' probability to answer the true label with a logistic transform of the product of these scores.
Following IRT, the difficulty is inferred as a latent variable given the answers: as for DS, the underlying tasks are discarded.
Finally, following deep learning progresses, end-to-end strategies have emerged that do not produce aggregated labels but allow to train classifiers from crowdsourced labels.
\citet{rodrigues2018deep} introduced $\mathrm{CrowdLayer}$ adding a new layer inside the network mimicking confusion matrices per worker.
Later, \citet{chu2021learning} have generalized this setting with $\mathrm{CoNAL}$, adding an element encoding global confusion.

Here, we propose the $\mathrm{WAUM}$ to combine the information from a confusion matrix per worker and a measure of relative difficulty between tasks.
It refines the judging system and identifies data points harming generalization that should be pruned.
Data pruning has been shown to improve generalization by removing mislabeled data \citep{Angelova_AbuMostafa_Perona05, pleiss_identifying_2020}, possibly dynamically along the learning phase \citep{ada_pruning} or by defining a forgetfulness score \citep{NEURIPS2021_ac56f8fe}.
\citet{sorscher2022beyond} have highlighted that data pruning strategies are highly impacted by the labeling in supervised settings and we confirm its relevance to the crowdsourcing framework. It is also a flexible tool that can be combined with most existing methods, using the pruning as a preliminary step.

\begin{figure}[thb]
    \centering
    \hfill
    \subfloat[\texttt{CIFAR-10H} dataset.]{\includegraphics[width=0.3083\textwidth]{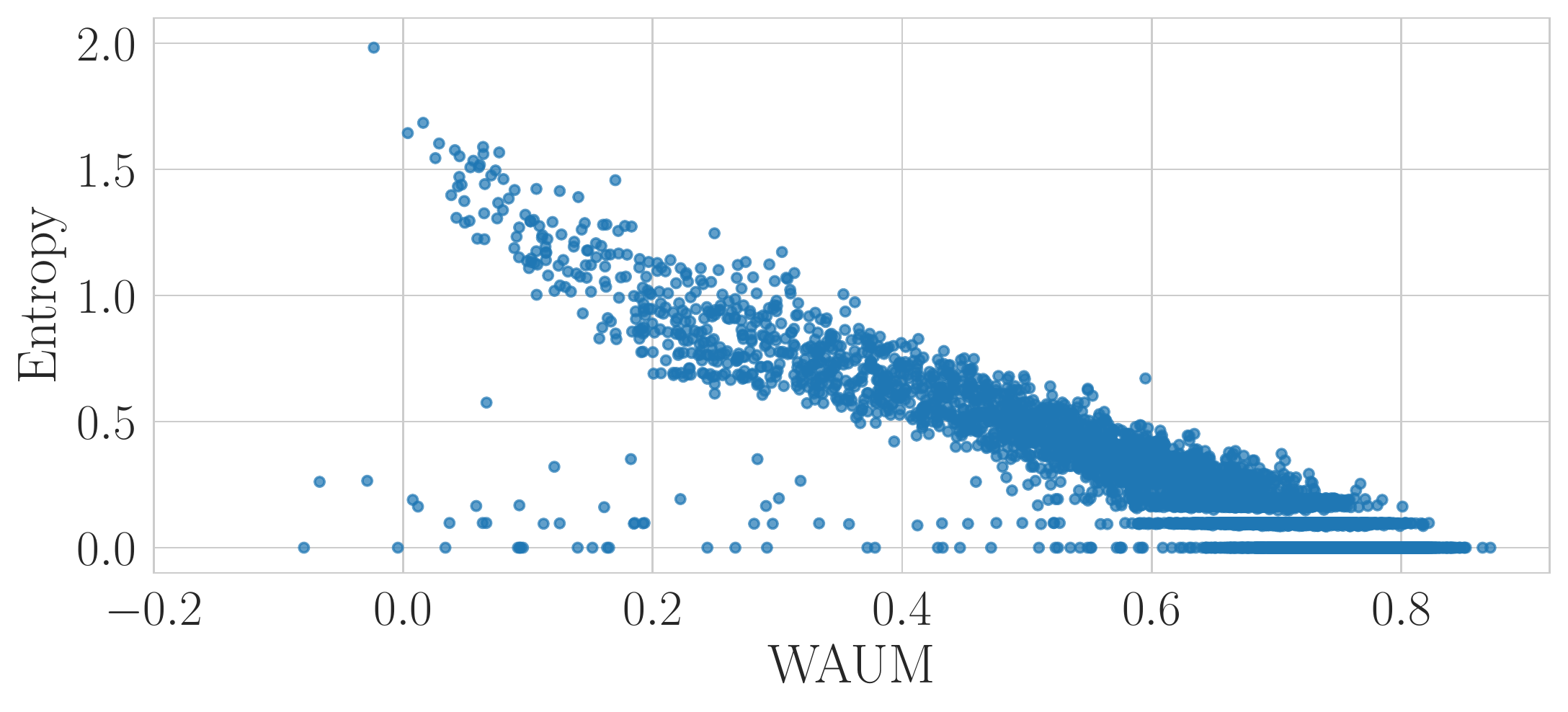}}
    \hfill
    \subfloat[\texttt{LabelMe} dataset.]{\includegraphics[width=0.3083\textwidth]{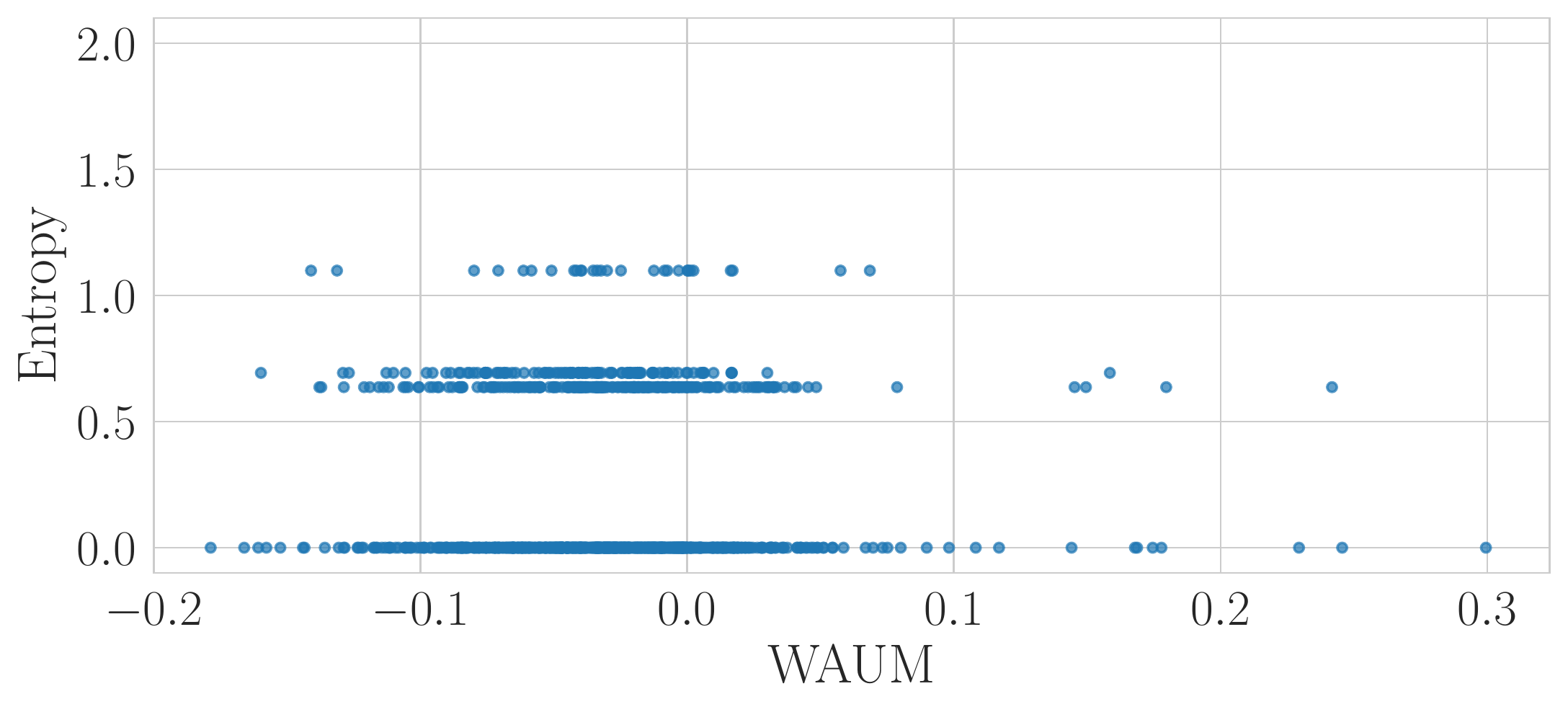}}
    \hfill
    \subfloat[\texttt{Music} dataset.]{\includegraphics[width=0.3083\textwidth]{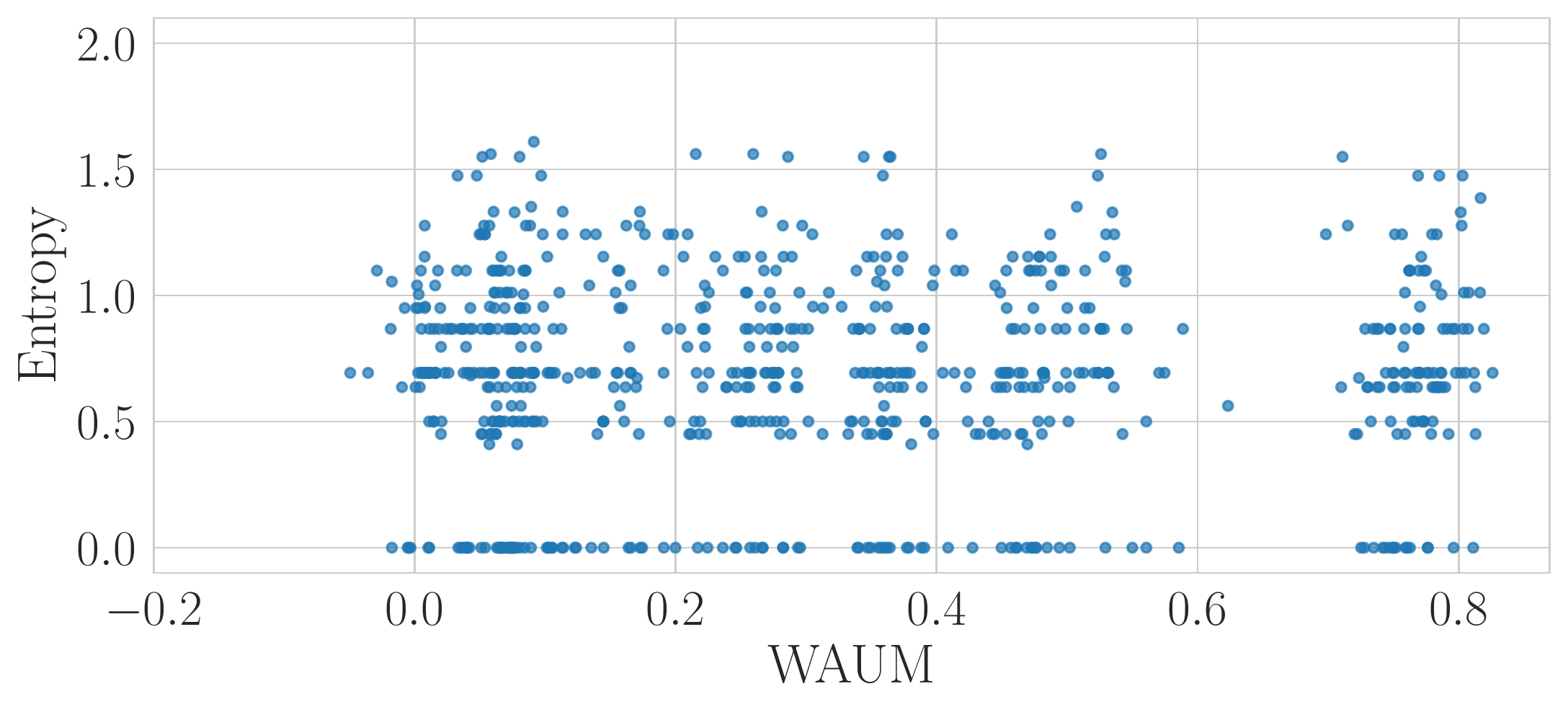}}
    \hfill
    \caption{Entropy of votes vs. $\mathrm{WAUM}$ for \texttt{CIFAR-10H}, \texttt{LabelMe}, and \texttt{Music}, each point representing a task/image. When large amounts of votes per task are available, $\mathrm{WAUM}$ and entropy ranking coincide well, as in (a). Yet, when votes are scarce, as in (b) and (c), entropy becomes irrelevant while our introduced $\mathrm{WAUM}$ remains useful. Indeed, tasks with few votes can benefit from feedback obtained for a similar one. And for the \texttt{LabelMe} dataset in particular, there are only up to three votes available per task, thus only four different values of the entropy possible, making it irrelevant in such cases for modeling task difficulty.}
    \label{fig:entropy_vs_waum}%
\end{figure}

\section{Weighted Area Under the Margin}
\label{sec:diff_aware}

\subsection{Definitions and Notation}
\label{subsec:defWAUM}

\paragraph{General notation.}
We consider classical multi-class learning notation, with input in $\gX$ and labels in $[K]:=\{1,\dots,K\}$.
The set of tasks is written as $\gX_\texttt{train} = \{x_1,\dots,x_{n_\texttt{task}}\}$, and we assume $\{(x_1,y_1^\star),\dots,(x_{n_\texttt{task}},y_{n_\texttt{task}}^\star)\}$ are $n_\texttt{task}$ \emph{i.i.d} tasks and labels,
with underlying distribution denoted by $\sP$.
The true labels $(y_i^\star)_{i \in [n_\texttt{task}]}$ are unobserved but crowdsourced labels are provided by $n_\texttt{worker}$ workers $(w_j)_{j \in [n_\texttt{worker}]}$.
We write $\gA(x_i)=\{j \in [n_\texttt{worker}]: \text{worker } w_j \text{ labeled task } x_i\}$ the \textbf{annotators set} of a task $x_i$ and $\gT(w_j)=\{ i \in [n_\texttt{task}]:\text{worker } w_j \text{ answered task } x_i\}$ the \textbf{tasks set} for a worker $w_j$.
For a task $x_i$ and each $j \in \gA(x_i)$, we denote $\smash{y_i^{(j)}\in[K]}$ the label answered by worker $w_j$.
Given an aggregation strategy \texttt{agg} (such as MV, DS or GLAD), we call estimated soft label $\hat y^{\texttt{agg}}_i$ the obtained label.
Note that for MV, the aggregated label $\hat y^{\mathrm{MV}}_i\in[K]$ and for other strategies, $\hat y^{\texttt{agg}}_i$ lies in the standard simplex $ \Delta_{K-1}=\{p \in \sR^K, \sum_{k=1}^K p_k =1, p_k \geq 0 \}$.
For any set $\mathcal{S}$, we write $|\mathcal{S}|$ for its cardinality.
Examples of annotators set and tasks set are provided in \Cref{fig:crowdsourcing_scheme}.
The training set has task-wise and worker-wise formulations:
\begin{align}\label{eq:training_crowd}
    \mathcal{D}_{\texttt{train}}
    = 
    \bigcup_{i=1}^{n_\texttt{task}}\bigg\{\big( x_i, \big(y_i^{(j)}\big)\big) \text{ for } j\in\gA(x_i)\bigg\} 
    =
    \bigcup_{j=1}^{n_\texttt{worker}}\underbrace{\bigg\{\big( x_i, \big(y_i^{(j)}\big)\big) \text{ for } i \in\gT(w_j)\bigg\}}_{\mathcal{D}_{\texttt{train}}^{(j)}}
    \enspace.
\end{align}

\paragraph{DS model notation.}
The Dawid and Skene (DS) model \citep{dawid_maximum_1979} aggregates answers and evaluates the workers' confusion matrix to observe where their expertise lies.
The confusion matrix of worker $w_j$ is denoted by $\pi^{(j)}\in\sR^{K\times K}$ and reflects individual error-rates between pairs of labels: $\smash{\pi^{(j)}_{\ell, k}}=\sP(y_i^{(j)}=k|y_i^\star=\ell)$ represents the probability that  worker $w_j$ gives label $k$ to a task whose true label is $\ell$.
The model assumes that the probability for a task $x_i$ to have true label $y_i^\star=\ell$ follows a multinomial distribution with probabilities $\smash{\pi^{(j)}_{\ell,\cdot}}$ for each worker, independently of $\gX_\texttt{train}$ (feature-blind).
In practice, DS estimates are obtained thanks to the EM algorithm to output estimated confusion matrices $(\pi^{(j)})_{j\in[n_{\texttt worker}]}$.
The full likelihood is given in \Cref{eq:ds}, \Cref{app:ds}.
Once DS confusion matrices are estimated, it is possible to use the diagonal terms as weights in a majority voting strategy. We denote this Weighted DS vote by $\mathrm{WDS}$, and give more details in \Cref{app:WDS}.
Essentially, the $\mathrm{WDS}$ strategy produces soft labels as NS, and also takes into account the estimated worker ability to recognize a task whose true label is indeed the voted one.

\subsection{Ambiguous tasks identification with the AUM}
To identify labeling errors and evaluate task difficulties, \citet{pleiss_identifying_2020} have introduced the $\mathrm{AUM}$ in the standard learning setting (\ie when $|\gA(x_i)|=1$ for all $i\in[n_\texttt{task}]$).
Given a training task and a label $(x,y)$, let $z^{(t)}(x)\in\sR^K$ be the logit score vector at epoch $t\leq T$ when learning a neural network (where $T$ is the number of training epochs).
We use the notation $\smash{z^{(t)}_{[1]}(x)\geq \dots\geq z^{(t)}_{[K]}(x)}$ for sorting $\smash{(z^{(t)}_{1}(x), \dots, z^{(t)}_{K}(x))}$ in non-increasing order.
Let us denote $\smash{\sigma^{(t)}(x):=\sigma(z^{(t)}(x))}$ the softmax output of the scores at epoch $t$.
Sorting the probabilities in decreasing order such that $\smash{\sigma^{(t)}_{[1]}(x)\geq \dots\geq\sigma^{(t)}_{[K]}(x)}$, the $\mathrm{AUM}$ reads:
\begin{align}\label{eq:Margin_WAUM}
    \mathrm{AUM}\left(x, y; \mathcal{D}_{\texttt{train}}\right)
    = \!\! \frac{1}{T}\sum_{t=1}^T \!\! \big[\sigma^{(t)}_{y} (x)- \sigma^{(t)}_{[2]}(x)\big]
    \enspace.
\end{align}
We write $\mathrm{AUM}\left(x, y\right)$ instead of $\mathrm{AUM}\left(x, y; \mathcal{D}_{\texttt{train}}\right)$ when the training set is clear from the context.
\citet{pleiss_identifying_2020} use an average of margins over logit scores, while we rather consider the average of margin after a softmax step in \Cref{eq:Margin_WAUM},
to temper scaling issues, as advocated by \citet{ju2018relative} in ensemble learning.
Moreover, we consider the margin introduced by \citet{yang2020consistency} since the corresponding hinge loss has better theoretical properties than the one used in the original $\mathrm{AUM}$, especially in top-$k$ settings\footnote{For top-$k$, consider $\sigma^{(t)}_{[k+1]}(x)$ instead of $\sigma^{(t)}_{[2]}(x)$ in \eqref{eq:Margin_WAUM}.} \citep{lapin2016loss, yang2020consistency,Garcin_Servajean_Joly_Salmon22}.

During the training phase, the $\mathrm{AUM}$ keeps track of the difference between the score assigned to the proposed label and the score assigned to the second-largest one.
It has been introduced to detect mislabeled observations in a dataset: the higher the AUM, the more confident the prediction is in the assigned label. Hence, the lower the $\mathrm{AUM}$, the more likely the label is wrong.
Finally, note that the $\mathrm{AUM}$ computation depends on the chosen neural network and on its initialization: pre-trained architectures could be used, yet any present bias would transfer to the $\mathrm{AUM}$ computation.

To generalize the $\mathrm{AUM}$ from \Cref{eq:Margin_WAUM} to the crowdsourcing setting, a difficulty lies in the term $\sigma_y^{(t)}(x)$ as, in this context, the label $y$ is unknown, as one observes several labels per task.
A naive adaptation of the $\mathrm{AUM}$ would be to use the majority voting strategy in order to recover a hard label to be used in \Cref{eq:Margin_WAUM}. We denote such a strategy by $\mathrm{AUMC}$ ($\mathrm{AUM}$ for Crowdsourced data). 
More formally, this writes as:
\begin{equation}\label{eq_aumc}
    \mathrm{AUMC}\left(x_i,\left\{y_i^{(j)}\right\}_{j\in\gA(x_i)};\gD_\texttt{train}\right) = \frac{1}{T}\sum_{t=1}^T \left[\sigma_{\hat y_i^{\mathrm{MV}}}^{(t)}(x_i) - \sigma_{[2]}^{(t)}(x_i)\right] \enspace.
\end{equation}
This naive approach can be refined by taking into account the whole distribution of labels, and not simply its mode (with MV).

\subsection{WAUM and data pruning}
\label{subsec:WAUM}

The $\mathrm{AUM}$ is defined in a standard supervised setting with (hard) labels.
The naive adaptation $\mathrm{AUMC}$ defined at \eqref{eq_aumc} does not take into account the fact that workers may have different abilities.
We now adapt the $\mathrm{AUM}$ to crowdsourced frameworks to improve the identification of difficult tasks.
Let $s^{(j)}(x_i)\in [0,1]$ be a trust factor in the answer of worker $w_j$ for task $x_i$.
The $\mathrm{WAUM}$ is then defined as:
\begin{align}
    \label{eq:WAUM}
    \mathrm{WAUM}(x_i)
     &
    = \tfrac{
    \displaystyle \sum_{j\in\gA(x_i)}  \!\!\!
    s^{(j)}(x_i) \mathrm{AUM}\big(x_i, y_i^{(j)}\big)
    }
    {\displaystyle\sum_{j'\in\gA(x_i)}
    s^{(j')}(x_i)}
    \enspace.
\end{align}
It is a weighted average of $\mathrm{AUM}$s over each worker's answer with a per task weighting score $s^{(j)}(x_i)$ based on workers' abilities.
This score considers the impact of the $\mathrm{AUM}$ for each answer since it is more informative if the $\mathrm{AUM}$ indicates uncertainty for an expert than for a non-expert.

The scores $s^{(j)}$ are obtained \emph{à la} \citet{servajean2017crowdsourcing}: each worker has an estimated confusion matrix $\hat{\pi}^{(j)}\in \sR^{K\times K}$.
Note that the vector $\mathrm{diag}(\hat{\pi}^{(j)}) \in \sR^{K}$ represents the probability for worker $w_j$ to answer correctly to each label.
With a neural network classifier, we estimate the probability for the input $x_i\in\gX_\texttt{train} $ to belong in each category by $\smash{\sigma^{(T)}(x_i)}$, \ie the probability estimate at the last epoch.
As a trust factor, we propose the inner product between the diagonal of the confusion matrix and the softmax vector:
\begin{align}\label{eq:trust_factor}
    s^{(j)}(x_i) = \big\langle \mathrm{diag}(\hat{\pi}^{(j)}) , \sigma^{(T)}(x_i) \big\rangle \in [0,1] \enspace.
\end{align}
The scores control the weight of each worker in \Cref{eq:WAUM}.
This choice of weight is inspired by the bilinear scoring system of $\mathrm{GLAD}$ \citep{whitehill_whose_2009}, as detailed hereafter.
The closer to one, the more we trust the worker for the given task.
The score $s^{(j)}(x_i)$ can be seen as a multidimensional version of $\mathrm{GLAD}$'s trust score.
Indeed, in  $\mathrm{GLAD}$, the trust score is modeled as the product $\alpha_j\beta_i$, with $\alpha_j\in\sR$ (resp. $\beta_i\in (0, +\infty)$) representing worker ability (resp. task difficulty).
In \Cref{eq:trust_factor}, the diagonal of the confusion matrix $\hat{\pi}^{(j)}$ represents the worker's ability and the softmax the task difficulty.

\paragraph{Dataset Pruning.} Our procedure (\Cref{alg:WAUMstack}) proceeds as follows.
We initialize our method by estimating the confusion matrices for all workers.
For each worker $w_j$, the $\mathrm{AUM}$ is computed for its labeled tasks, and so is its worker-dependent trust scores $s^{(j)}(x_i)$ with \Cref{eq:trust_factor}.
The $\mathrm{WAUM}$ in \Cref{eq:WAUM} is then computed for each task.
The most ambiguous tasks, the ones whose $\mathrm{WAUM}$ are below a threshold, are then discarded, and the associated pruned dataset $\gD_{\text{pruned}}$ is output. We consider for the pruning threshold a quantile of order $\alpha\in[0,1]$ of the $\mathrm{WAUM}$ scores.
The hyperparameter $\alpha$ (proportion of training data points pruned) can be chosen on a validation set, yet choosing $\alpha\in\{0.1, 0.05, 0.01\}$ has led to satisfactory results in all our experiments.
Note that the same pruning procedure can be applied to $\mathrm{AUMC}$ for comparison.

\paragraph*{Refined initialization: estimating confusion matrices.}
By default, we rely on the \texttt{Est}$=$DS algorithm to get workers' confusion matrices, but other estimates are possible: DS might suffer from the curse of dimensionality when the number $K$ of classes is large ($K^2$ coefficients needed per worker).

\begin{algorithm}[tb]
\caption{$\mathrm{WAUM}$ (Weighted Area Under the Margin).}
\label{alg:WAUMstack}
\textbf{Input}:  $\mathcal{D}_{\texttt{train}}$: tasks and crowdsourced labels, $\alpha\in[0,1]$: proportion of training points pruned, $T\in \sN$: number of epochs, $\texttt{Est}$: Estimation procedure for the confusion matrices\\
\textbf{Output}: pruned dataset $\mathcal{D}_{\text{pruned}}$
\begin{algorithmic}[1]
\STATE Get confusion matrix $\{\hat{\pi}^{(j)}\}_{j\in[n_\texttt{worker}]}$ from \texttt{Est}
\FOR{$j\in[n_\textrm{worker}]$}
\STATE Get $\mathrm{AUM}(x_i, y_i^{(j)}; \mathcal{D}_{\texttt{train}})$ using \Cref{eq:Margin_WAUM} for $i\in\gT(w_j)$
\STATE Get \textbf{trust scores} $s^{(j)}(x_i)$ using \Cref{eq:trust_factor} for $i\in\gT(w_j)$
\ENDFOR
\FOR{each task $x\in\gX_\texttt{train}$}
\STATE Compute $\mathrm{WAUM}(x)$ using \Cref{eq:WAUM}\;
\ENDFOR
\STATE  Get $q_{\alpha}$ $(\mathrm{WAUM}(x_i))_{i\in[n_\texttt{task}]}$, $\alpha$-\textbf{quantile threshold}
\STATE $\mathcal{D}_{\text{pruned}}\!=\!
        \Big\{
        \big( x_i, \big(y_i^{(j)}\big)_{j\in\gA(x_i)}\big) \! : \!\mathrm{WAUM}(x_i) \geq q_\alpha,  x_i \in \gX_\texttt{train}  \Big
        \}$
\end{algorithmic}
\end{algorithm}

\paragraph{Training on the pruned dataset}
Once a pruned dataset $\gD_{\text{pruned}}$ has been obtained thanks to the $\mathrm{WAUM}$, one can create soft labels through an aggregation step, and use them to train a classifier.
Aggregated soft labels contain information regarding human uncertainty, and could often be less noisy than NS labels.
They can help improve model calibration \citep{wen2020combining, zhong2021improving}, a property useful for interpretation \citep{jiang2012calibrating, kumar2019verified}.
Concerning the classifier training, note that it can differ from the one used to compute the $\mathrm{WAUM}$.
We train a neural network whose architecture is adapted dataset per dataset and that can differ from the one used in \Cref{alg:WAUMstack} (it is the case for instance for the \texttt{LabelMe} dataset).
For an aggregation technique $\texttt{agg}$, we write the full training method on the pruned dataset created from the $\mathrm{WAUM}$: $\texttt{agg}+\mathrm{WAUM}$ and instantiate several choices in \Cref{sec:experiments}.
For comparison, we write $\texttt{agg} + \mathrm{AUMC}$ the training method on the pruned dataset created from the $\mathrm{AUMC}$.

\section{Experiments}
\label{sec:experiments}

Our first experiments focus on multi-class classification datasets with a large number of votes per task.
We consider first a simulated dataset to investigate the $\mathrm{WAUM}$ and the pruning hyperparameter $\alpha$.
Then, with the real \texttt{CIFAR-10H} dataset from \citet{peterson_human_2019} we compare label aggregation-based procedures with and without pruning using the $\mathrm{AUMC}$ or the $\mathrm{WAUM}$.
Finally, we run our experiments on the \texttt{LabelMe} dataset from \citet{rodrigues2018deep} and \texttt{Music} dataset from \citet{rodrigues2014gaussian}, both real crowdsourced datasets with few labels answered per task.
For each aggregation scheme considered, we train a neural network on the soft labels (or hard labels for MV) obtained after the aggregation step.
We compare our $\mathrm{WAUM}$ scheme with several other strategies like $\mathrm{GLAD}$ (feature-blind) or $\mathrm{CoNAL}$ (feature-aware) with and without pruning from the $\mathrm{AUMC}$ identification step.
For $\mathrm{CoNAL}$, two regularization levels are considered: $\lambda=0$ and $\lambda=10^{-4}$ ($\lambda$ controls the distance between the global and the individual confusion matrices).
More simulations and overview of the methods compared are available in \Cref{subsec:Synthetic_dataset}.

\paragraph*{Metrics investigated}
After training, we report two performance metrics on a test set $\mathcal{D}_{\texttt{test}}$: top-$1$ accuracy and expected calibration error (ECE) (with $M=15$ bins as in \citet{guo_calibration_2017}).
The ECE measures the discrepancy between the predicted probabilities and the probabilities of the underlying distribution.
For ease of reporting results, we display the score $1-\mathrm{ECE}$ (hence, the higher the better, and the closer to $1$, the better the calibration); see \Cref{app:calibration} for more details.
Reported errors represent standard deviations over the repeated experiments (10 repetitions on simulated datasets and 3 for real datasets).

\paragraph*{Implementation details}
For simulations, the training is performed with a three dense layers' artificial neural network $(30, 20, 20)$ with batch size set to $64$.
Workers are simulated with \texttt{scikit-learn} \citep{scikit-learn} classical classifiers.
For \texttt{CIFAR-10H} the Resnet-$18$ \citep{he2016deep} architecture is chosen with batch size set to $64$.
We minimize the cross-entropy loss, and use when available a validation step to avoid overfitting.
For optimization, we consider an \texttt{SGD} solver with $150$ training epochs, an initial learning rate of $0.1$, decreasing it by a factor $10$ at epochs $50$ and $100$.
The $\mathrm{WAUM}$ and $\mathrm{AUMC}$ are computed with the same parameters for $T=50$ epochs.
Other hyperparameters for \texttt{Pytorch}'s \citep{pytorch} \texttt{SGD} are \texttt{momentum=0.9} and \texttt{weight\_decay=5e-4}.
For the \texttt{LabelMe} and \texttt{Music} datasets, we use the Adam optimizer with learning rate set to $0.005$ and default hyperparameters.
On these two datasets, the $\mathrm{WAUM}$ and $\mathrm{AUMC}$ are computed using a more classical Resnet-50 for $T=500$ epochs and the same optimization settings.
The architecture used for train and test steps is a pretrained VGG-$16$ combined with two dense layers as described in \citet{rodrigues2018deep} to reproduce original experiments on the \texttt{LabelMe} dataset.
This architecture differs from the one used to recover the pruned set.
Contrary to the modified VGG-$16$, the Resnet-$50$ could be fully pre-trained.
The general stability of pre-trained Resnets, thanks to the residuals connections, allows us to compute the $\mathrm{WAUM}$ and $\mathrm{AUMC}$ with way fewer epochs (each being also with a lower computational cost) compared to VGGs \citep{he2016deep}.
As there are few tasks, we use data augmentation with random flipping, shearing and dropout ($0.5$) for $1000$ epochs.
Experiments were executed with Nvidia RTX 2080 and Quadro T2000 GPUs.
\Cref{sec:aum_and_waum_additional_details} presents more details on the code used with the \texttt{peerannot} library.
Source codes are available at \url{https://github.com/peerannot/peerannot}. Evaluated strategies are at \url{https://github.com/peerannot/peerannot/tree/main/peerannot/models} sorted according to wether they are aggregation-based, learning-based or only for identification.
The $\mathrm{WAUM}$ and $\mathrm{AUMC}$ sources are available in the \texttt{identification} module.

\begin{figure}[tb]
    \centering
    \includegraphics[width=.75\linewidth]{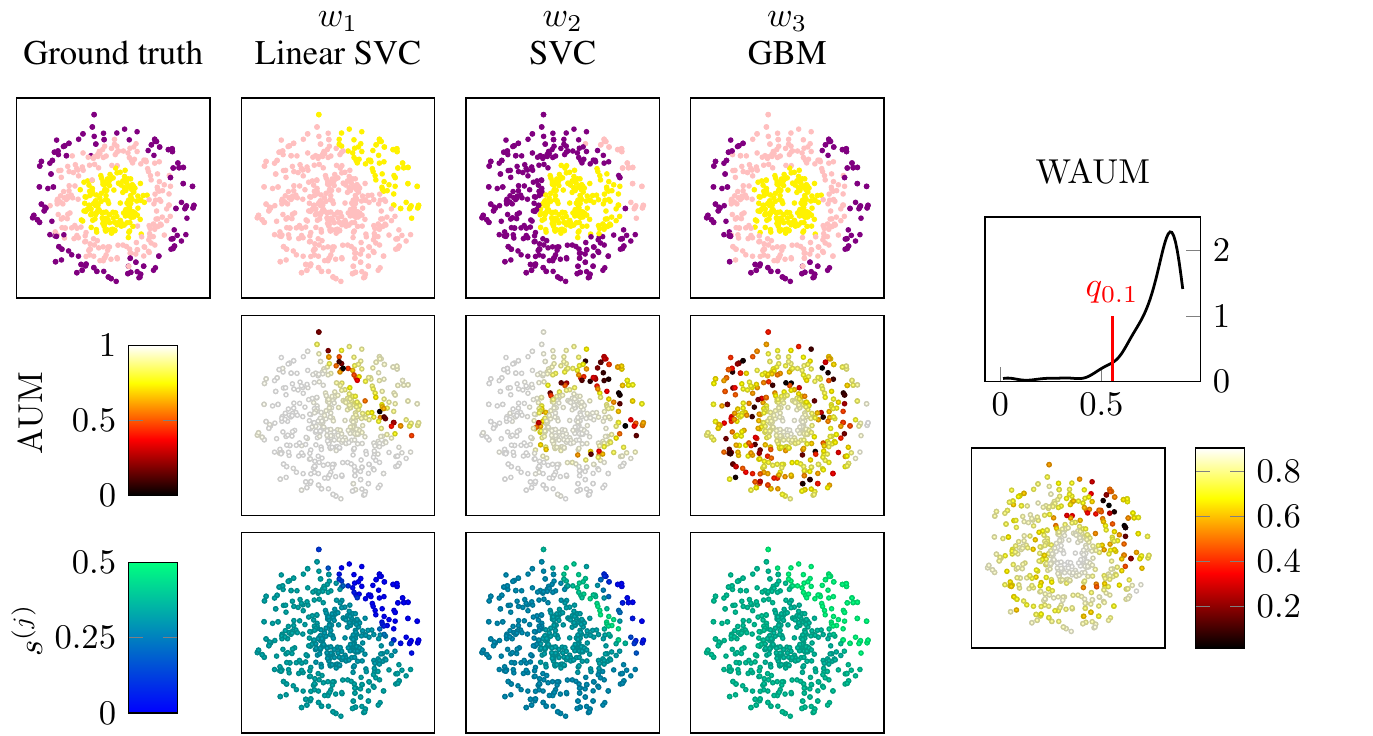}
    \caption{$\texttt{three\_circles}$: one realization of simulated workers $w_1,w_2,w_3$, with their $\mathrm{AUM}$, normalized trust scores $s^{(j)}$ (left) and $\mathrm{WAUM}$ distributions (right) for $\alpha=0.1$. Worker $w_1$ has less impact into the final $\mathrm{WAUM}$ in the disagreement area. Note also that for worker $w_1$ (LinearSCV), the region with low $\mathrm{AUM}$ values recovers the usual classifier's margin around the decision boundary.}
    \label{fig:threecircles_workers}
\end{figure}

\subsection{Simulated multiclass dataset: \texttt{three\_circles}.}
\label{subsec:3circles}
We simulate three cloud points (to represent $K=3$ classes) using \texttt{scikit-learn}'s function \texttt{two\_circles}; see \Cref{fig:threecircles_workers}.
The $n_\texttt{worker}=3$ workers are standard classifiers: $w_1$ is a linear Support Vector Machine Classifier (linear SVC), $w_2$ is an SVM with RBF kernel (SVC), and $w_3$ is a gradient boosted classifier (GBM).
Data is split between train (70\%) and test (30\%) for a total of $750$ points and each simulated worker votes for all tasks, \ie for all $x\in\gX_\texttt{train}$, $|\gA(x)|=n_\texttt{worker}=3$, leading to $n_{\texttt{task}}=525$ tasks (points).
The performance reported in \Cref{tab:res3circles} is averaged over $10$ repetitions.

\begin{table}[t]
    \centering

    \label{tab:res3circles}
    \begin{footnotesize}
        \begin{tabular}{m{3.9cm}cc}
            Strategy              & $\mathrm{Acc}_{\texttt{test}}$ & ECE                      \\ \hline \\[-0.2cm]
            $\mathrm{MV}$                   & $0.73\pm 0.03$                 & $\mathbf{0.13}\pm 0.03$                         \\
            $\mathrm{NS}$                   & $0.70\pm 0.02$                 & $0.18\pm 0.02$                                  \\
            $\mathrm{DS}$                   & $0.75\pm 0.07$                 & $0.22\pm 0.08$                                \\
            $\mathrm{GLAD}$                 & $0.58\pm 0.02$                 & $0.36\pm 0.02$                             \\
              \rowcolor{gray!20}$\mathrm{WDS}$                  & $0.81\pm 0.04$                 & $0.17\pm 0.03$                                \\
              \rowcolor{gray!20}$\mathrm{WDS+WAUM}(\alpha=10^{-2})$ & $0.80\pm 0.04$                 & $0.17\pm 0.01$\\
              \rowcolor{gray!20}$\mathrm{WDS+WAUM}(\alpha=10^{-1})$ & $\mathbf{0.83}\pm 0.03$        & $0.19\pm 0.04$                             \\
              \rowcolor{gray!20}$\mathrm{WDS+WAUM}(\alpha=0.25)$    & $0.69\pm 0.02$                 & $0.19\pm 0.02$         
        \end{tabular}
    \end{footnotesize}
        \caption{\texttt{three\_circles}: Aggregation and learning performance presented in \Cref{fig:threecircles_predictions} ($n_\texttt{task}=525$ tasks, $|\gA(x)|=n_\texttt{worker}=3$, $10$ repetitions). 
        Note that the best worker, $w_3$, reaches
        $0.84$ on test accuracy.
        }
\end{table}

\begin{figure}[t]
    \centering
    \includegraphics[width=.75\linewidth]{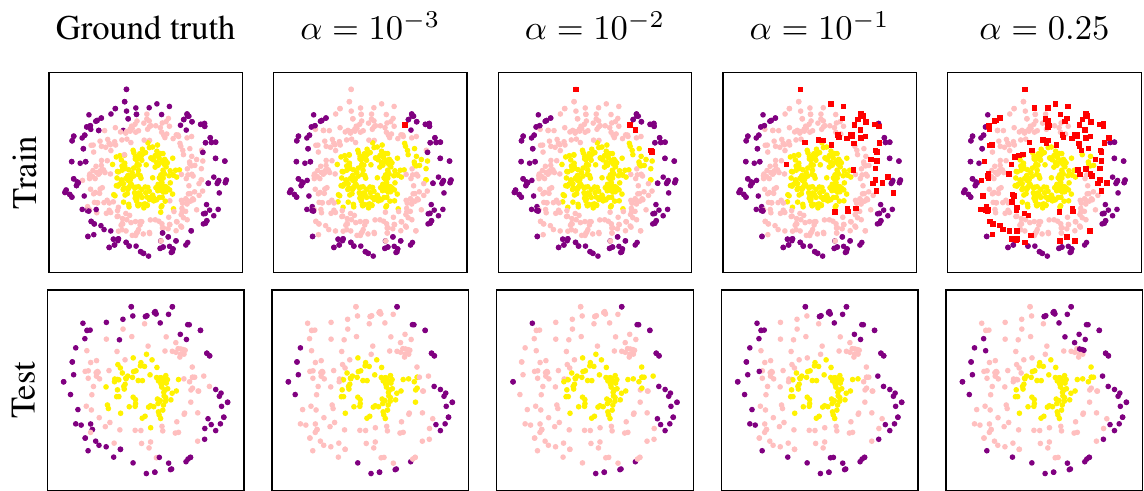}
    \caption{Influence of $\alpha$ on the pruning step. Red dots indicate data points pruned from the training set, at level $q_\alpha$ in the $\mathrm{WAUM}$ (see line 10 in \Cref{alg:WAUMstack}).
        We consider ($\alpha\in\{10^{-3}, 10^{-2}, 10^{-1}, 0.25\}$).
        The neural network used for predictions is three dense layers'
        $(30, 20, 20)$, as for other simulated experiments.
        Training labels are from the $\mathrm{WDS+WAUM}$ strategy with performance reported in \Cref{tab:res3circles}.
        The more we prune data, the worse the neural network can learn from the training dataset.
        However, removing the tasks with high disagreement noise helps to generalize.
        }
    \label{fig:three_circles_alpha_influence}
\end{figure}

A disagreement area is identified in the northeast area of the dataset (see \Cref{fig:threecircles_workers}).
\Cref{tab:res3circles} also shows that pruning too little data ($\alpha$ small) or too much ($\alpha$ large) can mitigate the performance.
In \Cref{fig:three_circles_alpha_influence}, we show the impact of the pruning hyperparameter $\alpha$. The closer $\alpha$ is to $1$, the more training tasks are pruned from the training set (and the worse the performance).

\begin{figure}[t]
    \centering
    \includegraphics[width=.8\textwidth]{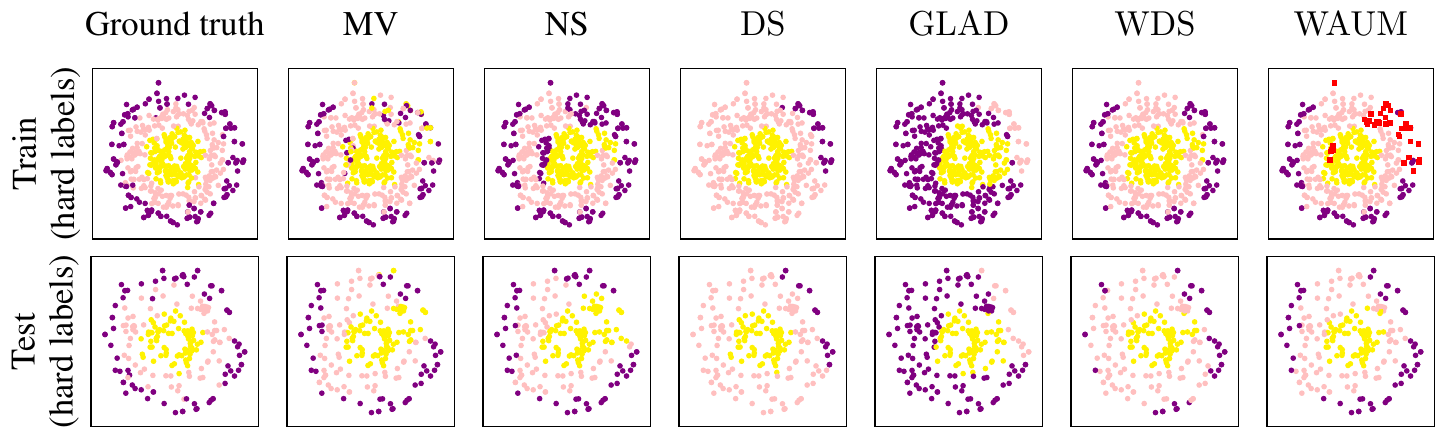}
    \caption{\texttt{three\_circles}: One realization of \Cref{tab:res3circles} varying the aggregation strategy. Training labels are provided from \Cref{fig:threecircles_workers} and predictions on the test set are from three dense layers' artificial neural network $(30, 20, 20)$ trained on the aggregated soft labels. For ease of visualization, the color displayed for each task represents the most likely class.
        Red points are pruned from training by $\mathrm{WAUM}$ with threshold $\alpha=0.1$. Here, we have $n_{\texttt{task}}=525$.
        $\mathrm{WAUM}$ method as in \Cref{tab:res3circles} uses $\mathrm{WDS}$ labels.
        }
    \label{fig:threecircles_predictions}
\end{figure}

\subsection{Real datasets}
\label{subsec:Real datasets}

In this section, we investigate three popular crowdsourced datasets: \texttt{CIFAR-10H}, \texttt{LabelMe} and \texttt{Music}.
The first one, \texttt{CIFAR-10H} \citep{peterson_human_2019}, is a curated dataset with many votes per task while \texttt{LabelMe} \citep{rodrigues2018deep} and \texttt{Music} \citep{rodrigues2014gaussian} datasets are more challenging, having fewer labels per task.
This low number of votes per task, especially for $\texttt{LabelMe}$ can lead to erroneous MV label which then impact the quality of the $\mathrm{AUMC}$. In this context, the label distribution's entropy is also a poor choice to identify hard tasks as can be seen in \Cref{fig:entropy_vs_waum}.
Indeed, with up to three labels, the entropy can only take four different values and thus is no help in ranking the difficulty of $1000$ tasks.

\begin{figure}[th]
  \centering
  \begin{tabular}{m{1.75cm} *{4}{>{\centering\arraybackslash}m{.15\linewidth}}}
    \centering CIFAR-10H & \multicolumn{2}{c}{\raisebox{-.5\height}{\includegraphics[width=0.4\linewidth]{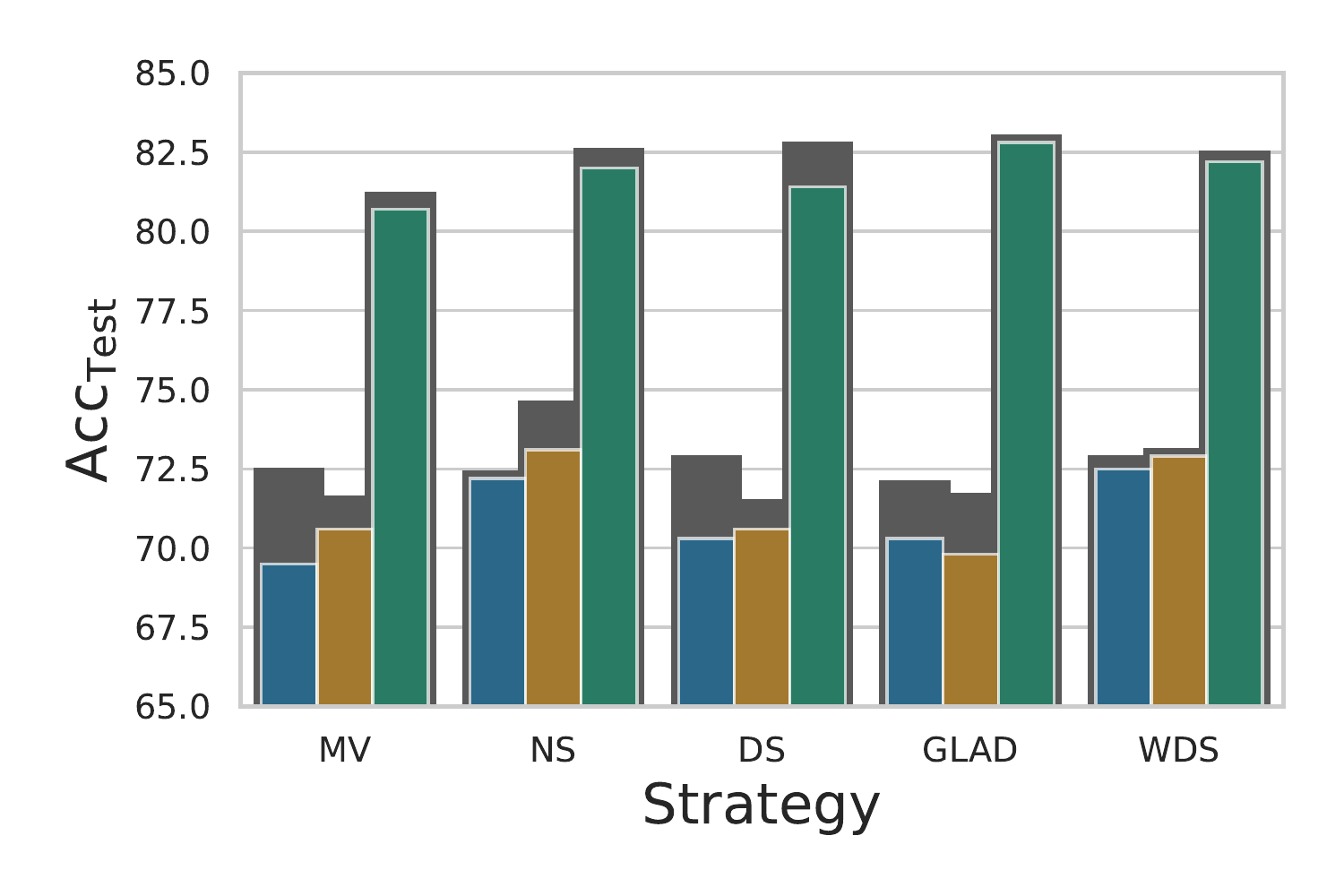}}} &
    \multicolumn{2}{c}{\raisebox{-.5\height}{\includegraphics[width=0.4\linewidth]{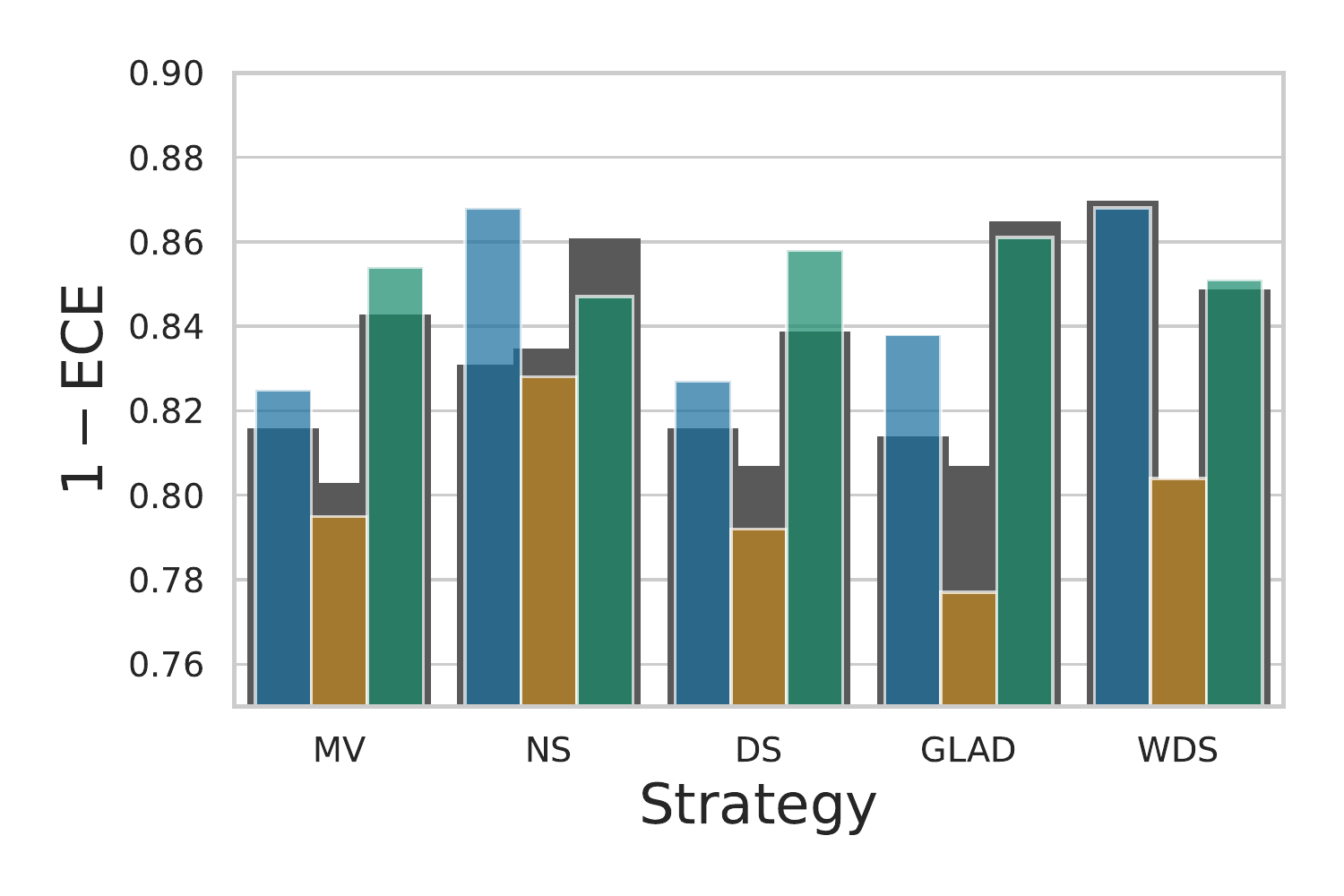}} }\\
    \centering LabelMe & \multicolumn{2}{c}{\raisebox{-.5\height}{\includegraphics[width=0.4\linewidth]{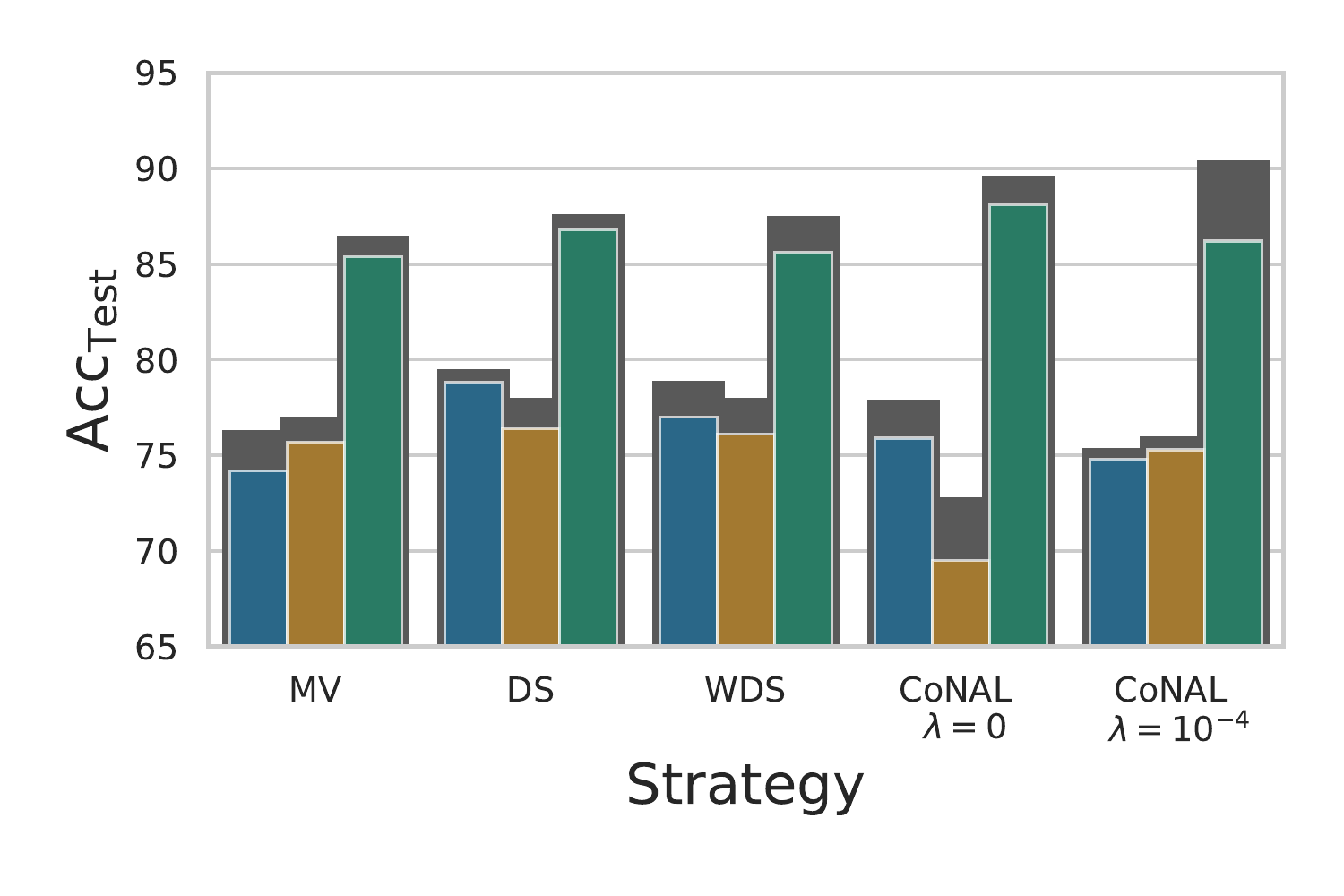} }}&
    \multicolumn{2}{c}{\raisebox{-.5\height}{\includegraphics[width=0.4\linewidth]{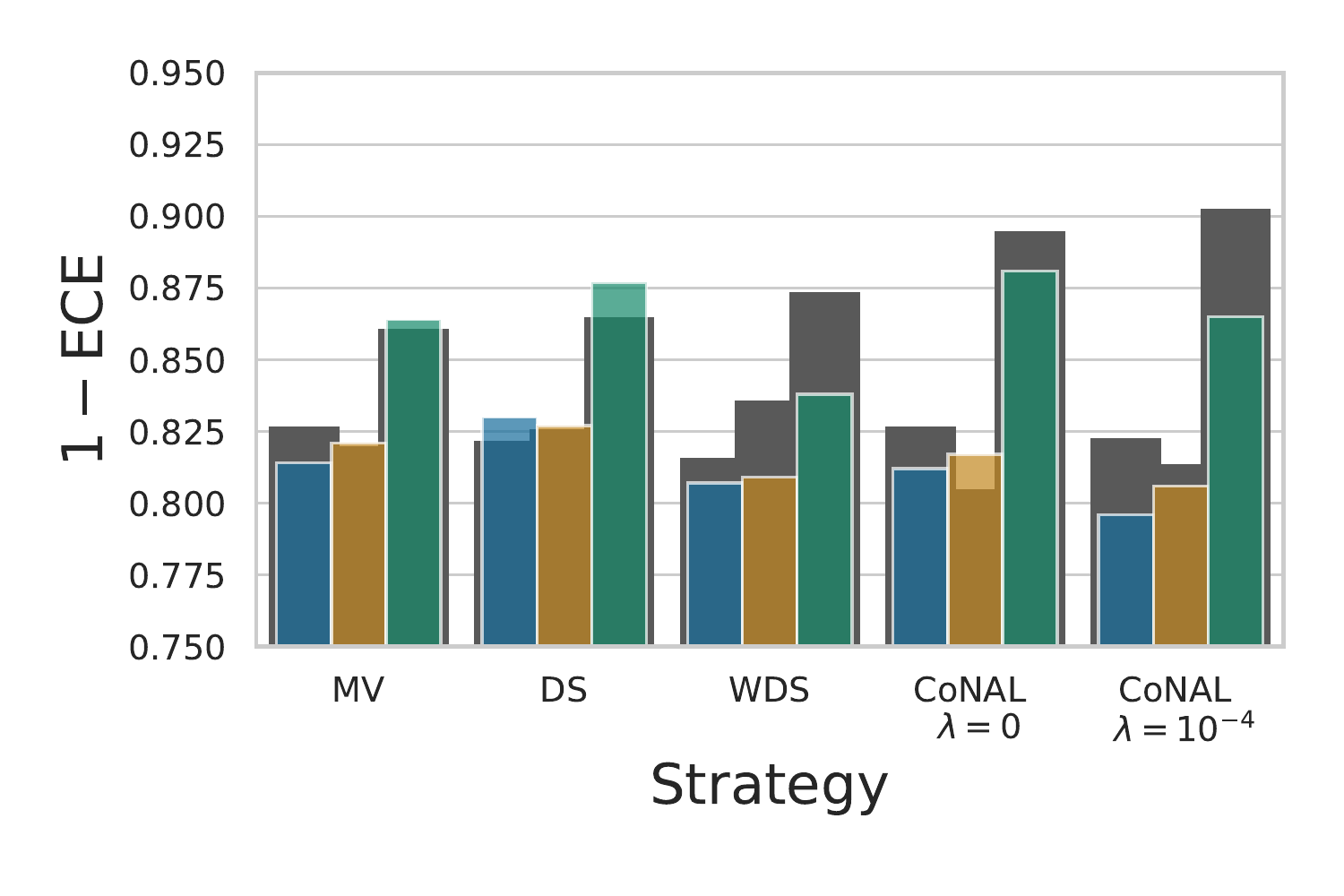}}} \\
    &\multicolumn{4}{c}{\includegraphics[width=0.75\linewidth]{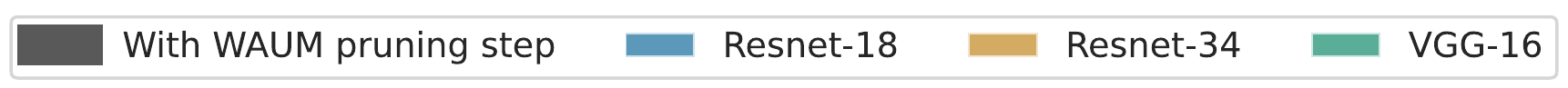}} \\
  \end{tabular}
  \caption{Performance obtained by training on the pruned dataset from the $\mathrm{WAUM}$ preprocessing step on \texttt{CIFAR-10H} and \texttt{LabelMe}. We consider multiple neural network architectures -- ResNet-18, ResNet-34 or VGG-16 with batch normalization and two supplementary dense layers. We show that performance in accuracy are improved in most cases. Calibration performance in term of $\mathrm{ECE}$ fluctuate depending on the architecture considered, especially for the \texttt{CIFAR-10H} dataset. Using the $\mathrm{WAUM}$ with $\mathrm{CoNAL}$ on the \texttt{LabelMe} dataset, we obtain best performance both in accuracy and calibration.}
  \label{fig:tab_arch}
\end{figure}

To prune only a few tasks, we choose $\alpha = 1\%$ for \texttt{CIFAR-10H} and \texttt{LabelMe} datasets.
For the \texttt{Music} dataset, $\alpha=5\%$ leads to better generalization performance; considering the dataset size and complexity, picking $\alpha=0.1$ would be harmful.
Ablation studies by architecture are performed on \texttt{CIFAR-10H} and \texttt{LabelMe} datasets in \Cref{fig:tab_arch} to show consistent improvement in performance by using the $\mathrm{WAUM}$ to prune ambiguous data.

\begin{figure}[tbh]
    \centering
    \includegraphics[width=0.9\columnwidth]{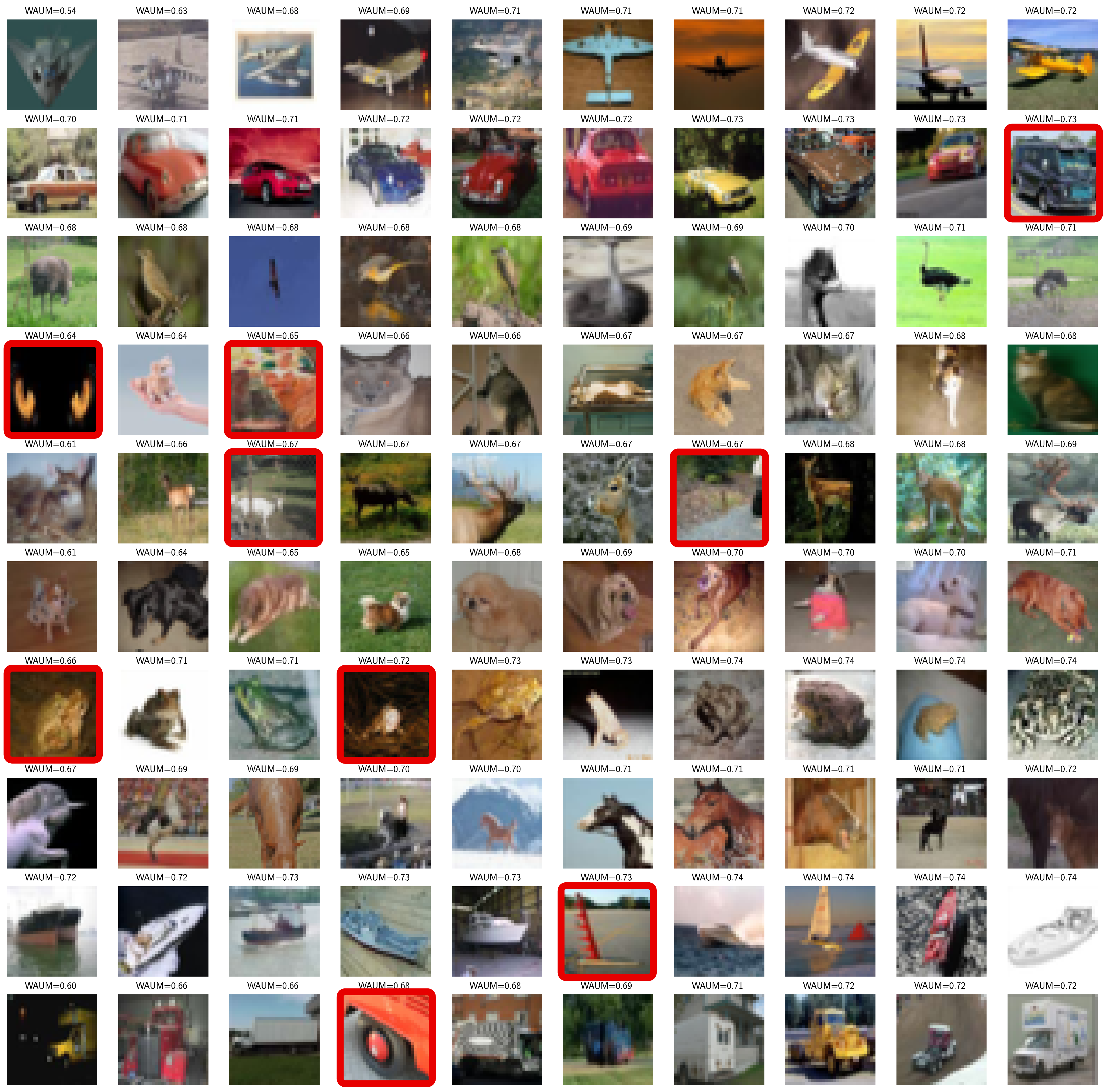}
    \caption{\texttt{CIFAR-10H}: 10 worst images for $\mathrm{WAUM}$ scores, by labels given in \CIFAR. The rows represent the labels \texttt{airplane}, \texttt{automobile}, \texttt{bird}, \texttt{cat}, \texttt{deer}, \texttt{dog}, \texttt{frog}, \texttt{horse}, \texttt{ship}, and \texttt{truck}. Images in red can be particularly hard to classify as they are not typical examples of their label. Comparison with the $\mathrm{AUMC}$ and the $\mathrm{AUM}$ are available in \Cref{fig:comparison_waums_aumc} \Cref{subsubsec:cifar-10h_dataset}.}
    \label{fig:worse_C10H}
\end{figure}

\paragraph*{\texttt{CIFAR-10H} dataset.} The training part of \texttt{CIFAR-10H} consists of the $10000$ tasks extracted from the test set of the classical \CIFAR dataset \citep{krizhevsky2009learning}, and $K=10$.
A total of $n_\texttt{worker}=2571$ workers participated on the Amazon Mechanical Turk platform, each labeling $200$ images ($20$ from each original class), leading to approximately $50$ answers per task.
We have randomly extracted $500$ tasks for a validation set (hence $n_{\texttt{train}}=9500$).
This dataset is notoriously more curated \citep{aitchison2020statistical} than a common dataset in the field: most difficult tasks were identified and removed at the creation of the \CIFAR dataset, resulting in few ambiguities.
\Cref{tab:spam-0-expe} shows that in this simple setting, our data pruning strategy is still relevant, with the choice $\alpha=0.01$. Images with worst $\mathrm{WAUM}$ for each class are presented in \Cref{fig:worse_C10H}.

\begin{table}[t]
    \label{tab:spam-0-expe}
    \begin{footnotesize}
        \begin{center}
            \begin{tabular}{lcc}
                Strategy & $\mathrm{Acc}_{\texttt{test}} (\%)$ & $1-\mathrm{ECE}$          
                \\ \hline \\[-0.2cm]
                MV                 & $69.53\pm 0.84$               & $0.825\pm 0.00$                                 \\
                MV + AUMC & $71.12\pm 1.12$ & $\mathbf{0.836\pm 0.01}$ \\
                MV + WAUM & $\mathbf{72.34\pm 1.01}$ & $0.814\pm 0.02$ \\
                \rowcolor{gray!20}NS                 & $72.14\pm 2.74$               & $\mathbf{0.868\pm 0.03}$          \\
                \rowcolor{gray!20}NS + AUMC                 & $71.80\pm 2.12$               & $0.838\pm 0.00$          \\
                \rowcolor{gray!20}NS + WAUM                 & $\mathbf{72.21\pm 1.82}$               & $0.829\pm 0.00$          \\
                DS                 & $70.26\pm 0.93$               & $0.827\pm 0.00$             \\
                DS + AUMC                 & $70.43\pm 1.10$               & $\mathbf{0.833\pm 0.02}$             \\
                DS + WAUM                & $\mathbf{72.71\pm 0.98}$               & $0.814\pm 0.02$             \\
               \rowcolor{gray!20} $\mathrm{GLAD}$    & $70.28\pm 0.88$               & $\mathbf{0.838\pm 0.01}$\\
                \rowcolor{gray!20} $\mathrm{GLAD}$ + AUMC    & $70.42\pm 1.23$               & $0.830\pm 0.01$\\
               \rowcolor{gray!20} $\mathrm{GLAD}$ + WAUM    & $\mathbf{71.93\pm 1.12}$               & $0.812\pm 0.02$\\
                  WDS                & $72.49\pm 0.48$               & $\mathbf{0.868\pm 0.00}$         \\
                  WDS + AUMC                & $72.47\pm 0.45$               & $0.866\pm 0.00$         \\

                  $\mathrm{WDS} + \mathrm{WAUM}$    & $\mathbf{72.67\pm 0.59}$      & $\mathbf{0.868 \pm 0.00}$ 
            \end{tabular}
        \end{center}
    \end{footnotesize}
    \caption{\texttt{CIFAR-10H}: performance of a \texttt{ResNet-18} by label-aggregation crowdsourcing strategy ($\alpha=0.01$).}
\end{table}

Furthermore, the $\mathrm{WAUM}$ leads to better generalization performance than the vanilla DS model and the pruning with $\mathrm{AUMC}$.
Overall, we show that there is a gain in performance to obtain by using a pruning preprocessing step compared to training the classifier on the aggregated labels for the full training set.
There is consistently an improvement on using the $\mathrm{WAUM}$ pruning -- which is weights the margins by worker and tasks -- over the naive $\mathrm{AUMC}$ which does not use reweighing.

\texttt{CIFAR-10H} is a relatively well-curated dataset, and we observe in \Cref{tab:spam-0-expe} that in this case, simple aggregation methods already perform well, in particular NS.
Over the $2571$ workers, less than $20$ are identified as spammers using \citet{raykar_ranking_2011} but note that most difficult tasks were removed when creating the original \CIFAR dataset.
We refer to the \emph{"labeler instruction sheet"} of \citet[Appendix C]{krizhevsky2009learning} for more information about the directives given to workers.

\begin{figure}[t]
    \label{fig:labelme_worstWAUM}%
    \centering
    \hfill
    \subfloat[Label \texttt{street}.]{\includegraphics[width=0.49583\textwidth]{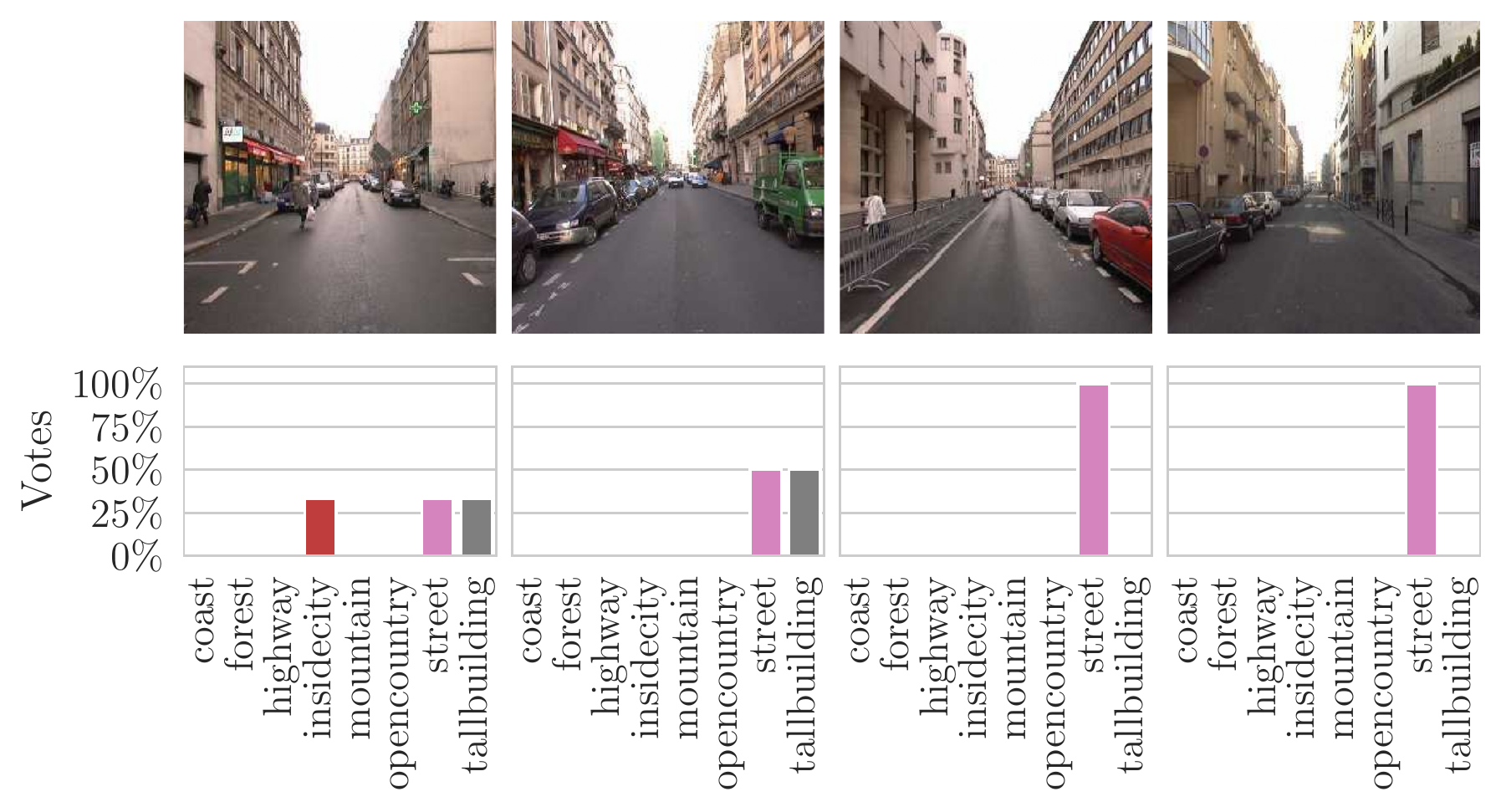}}
    \hfill
    \subfloat[Label \texttt{tallbuilding}.]{\includegraphics[width=0.49583\textwidth]{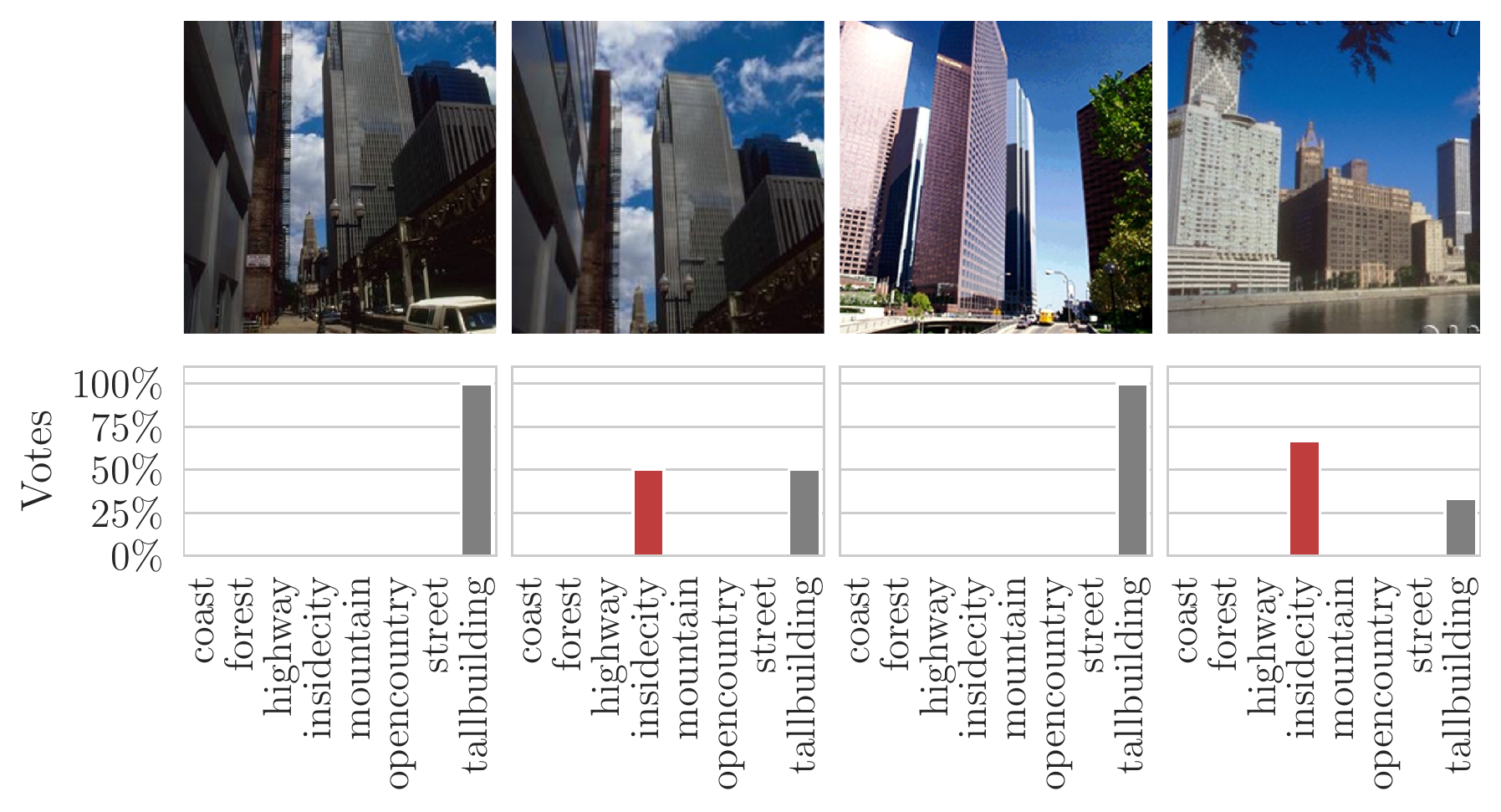}}
    \caption{\texttt{LabelMe} dataset: Worst $\mathrm{WAUM}$ for classes (top) and the associated voting distribution for each image (bottom). (a) Label \texttt{street} (b)  Label \texttt{tallbuilding}.
        Even if the two tasks are very similar, because the workers are different the associated proposed labels can differ and add noise during training.}
\end{figure}

\paragraph{\texttt{LabelMe} dataset.} This dataset consists in classifying $1000$ images in $K=8$ categories.
In total $77$ workers are reported in the dataset (though only $59$ of them answered any task at all).
Each task has between $1$ and $3$ labels. A validation set of $500$ images and a test set of $1188$ images are available.

\begin{table}
\begin{minipage}{0.49\linewidth}
\centering
  \footnotesize
            \begin{tabular}{lcc}
                Strategy                              & $\mathrm{Acc}_{\texttt{test}} (\%)$ & $1-\mathrm{ECE}$ 
                \\  \hline \\[-0.2cm]
                MV                                              & $85.4\pm1.0$                   & $\mathbf{0.864\pm 0.01}$                          \\
                MV + AUMC                                             & $86.0\pm1.1$                   & $0.859\pm 0.01$                          \\
                MV + WAUM                                              & $\mathbf{86.1\pm0.9}$                   & $0.858\pm 0.02$                          \\
               \rowcolor{gray!20} NS                                              & $86.1\pm 1.0$                  & $0.862\pm 0.01$                             \\
               \rowcolor{gray!20} NS + AUMC                                              & $87.2\pm 0.8$                  & $0.882\pm 0.01$                             \\
               \rowcolor{gray!20} NS + WAUM                                             & $\mathbf{88.1\pm 1.0}$                  & $\mathbf{0.890\pm 0.02}$                             \\
                DS                                              & $86.8\pm 0.5$                  & $\mathbf{0.877\pm 0.01}$                           \\
                DS + AUMC                                              & $86.3\pm 0.5$                  & $0.841\pm 0.02$                           \\
                DS + WAUM                                              & $\mathbf{87.2\pm 0.6}$                  & $0.862\pm 0.02$                           \\
               \rowcolor{gray!20}$\mathrm{GLAD}$                                 & $87.1\pm 0.9$                  & $0.881\pm 0.01$                           \\
               \rowcolor{gray!20}$\mathrm{GLAD}$ + AUMC                                 & $87.6\pm 1.1$                  & $0.861\pm 0.03$                           \\
               \rowcolor{gray!20}$\mathrm{GLAD}$ + WAUM                                & $\mathbf{88.2\pm 0.8}$                  & $\mathbf{0.885\pm 0.02}$                           \\
                $\mathrm{WDS}$ & $85.6\pm 0.7$ & $0.838\pm 0.02$  \\
                $\mathrm{WDS}$ + AUMC & $86.7\pm 0.7$ & $0.862\pm 0.02$  \\
                $\mathrm{WDS}$ + WAUM& $\mathbf{87.1\pm 0.8}$ & $\mathbf{0.871\pm 0.01}$  \\
                \rowcolor{gray!20}$\mathrm{CrowdLayer}$                           & $85.4\pm 4.2$                  & $0.858\pm 0.04$                                    \\
                \rowcolor{gray!20}$\mathrm{CrowdLayer}$ + AUMC                          & $87.1\pm 3.5$                  & $0.809\pm 0.05$                                    \\
                \rowcolor{gray!20}$\mathrm{CrowdLayer}$ + WAUM                           & $\mathbf{87.5\pm 3.2}$                  & $\mathbf{0.860\pm 0.03}$                                    \\
                $\mathrm{CoNAL (\lambda=0)}$                    & $88.1\pm 1.0$                  & $0.881\pm 0.01$                      \\
                $\mathrm{CoNAL (0)}$ + AUMC                   & $89.1\pm 1.1$                  & $\mathbf{0.903\pm 0.02}$                      \\
                $\mathrm{CoNAL (0) + WAUM }$        & $\mathbf{89.2\pm 1.0}$                  & $0.892\pm 0.01$ \\
                 \rowcolor{gray!20} $\mathrm{CoNAL (\lambda=10^{-4})}$              & $86.2\pm 6.4$                  & $0.865\pm 0.06$                 \\
                 \rowcolor{gray!20} $\mathrm{CoNAL (10^{-4})}$ + AUMC              & $88.4\pm 2.3$                  & $0.884\pm 0.04$                 \\
                 \rowcolor{gray!20} $\mathrm{CoNAL (10^{-4}) + WAUM}$ & $\mathbf{90.0\pm 0.8}$         & $\mathbf{0.901\pm 0.01}$ 
            \end{tabular}
  \captionof{table}{Ablation study on \texttt{LabelMe} using the VGG backbone: $\alpha=0.01$}
      \label{tab:labelme}
\end{minipage}\hfill
\begin{minipage}{0.49\linewidth}
  \centering
  \footnotesize
            \begin{tabular}{lcc}
                Strategy                              & $\mathrm{Acc}_{\texttt{test}} (\%)$ & $1-\mathrm{ECE}$ 
                \\  \hline \\[-0.2cm]
                MV                                              & $59.9\pm1.23$                   & $0.631\pm 0.01$                          \\
                MV + AUMC                                             & $\mathbf{62.0\pm1.23}$                   & $\mathbf{0.650\pm 0.02}$                          \\
                MV + WAUM                                              & $61.1\pm2.35$                   & $0.624\pm 0.02$                          \\
               \rowcolor{gray!20} NS                                              & $59.9\pm 1.40$                  & $0.624\pm 0.02$                             \\
               \rowcolor{gray!20} NS + AUMC                                              & $59.9\pm 1.41$                  & $0.640\pm 0.02$                             \\
               \rowcolor{gray!20} NS + WAUM                                             & $\mathbf{62.1\pm 2.18}$                  & $\mathbf{0.642\pm 0.01}$                             \\
                DS                                              & $\mathbf{62.9\pm 1.72}$                  & $\mathbf{0.661\pm 0.01}$                           \\
                DS + AUMC                                              & $61.5\pm2.22$                  & $0.659\pm 0.01$                           \\
                DS + WAUM                                              & $62.1\pm 2.81$                  & $0.640\pm 0.02$                           \\
               \rowcolor{gray!20}$\mathrm{GLAD}$                                 & $61.5\pm 1.72$                  & $0.639\pm 0.01$                           \\
               \rowcolor{gray!20}$\mathrm{GLAD}$ + AUMC                                 &      $\mathbf{61.6\pm 0.93}$             & $\mathbf{0.664\pm 0.01}$                            \\
               \rowcolor{gray!20}$\mathrm{GLAD}$ + WAUM                                & $61.5\pm 1.23$                  & $0.645\pm 0.01$                           \\
                $\mathrm{WDS}$ & $60.2\pm 1.66$ & $0.652\pm 0.01$  \\
                $\mathrm{WDS}$ + AUMC & $62.9\pm 2.67$ & $0.647\pm 0.03$  \\
                $\mathrm{WDS}$ + WAUM& $\mathbf{63.1\pm 0.91}$ & $\mathbf{0.660\pm 0.02}$  \\
                \rowcolor{gray!20}$\mathrm{CrowdLayer}$                           & $63.2\pm 1.34$                  & $0.615\pm 0.02$                                    \\
                \rowcolor{gray!20}$\mathrm{CrowdLayer}$ + AUMC             &  $\mathbf{63.3\pm 2.54}$    &          $0.617\pm 0.04$         \\
                \rowcolor{gray!20}$\mathrm{CrowdLayer}$ + WAUM                           & $63.2\pm 2.46$                 &$\mathbf{0.680\pm 0.03}$                                    \\
                $\mathrm{CoNAL (\lambda=0)}$                    & $64.2\pm 0.91$                  & $0.660\pm 0.02$                      \\
                $\mathrm{CoNAL (0)}$ + AUMC                   &         $64.3\pm 0.88$          &  $\mathbf{0.735\pm 0.01}$                     \\
                $\mathrm{CoNAL (0) + WAUM }$        & $\mathbf{64.5\pm 0.76}$                  & $\mathbf{0.735\pm 0.01}$ \\
                 \rowcolor{gray!20} $\mathrm{CoNAL (\lambda=10^{-4})}$              & $64.2\pm 0.55$                  & $0.639\pm 0.06$                 \\
                 \rowcolor{gray!20} $\mathrm{CoNAL (10^{-4})}$ + AUMC              &     $64.1\pm 0.74$              &  $\mathbf{0.745\pm 0.02}$     \\
                 \rowcolor{gray!20} $\mathrm{CoNAL (10^{-4}) + WAUM}$ & $\mathbf{64.4\pm 0.78}$         & $0.726\pm 0.02$ 
            \end{tabular}
\captionof{table}{Ablation study on \texttt{Music} using the VGG backbone: $\alpha=0.05$}
    \label{tab:music}
\end{minipage}
\end{table}

\begin{figure}[t]
    \centering
    \includegraphics[width=0.9\columnwidth]{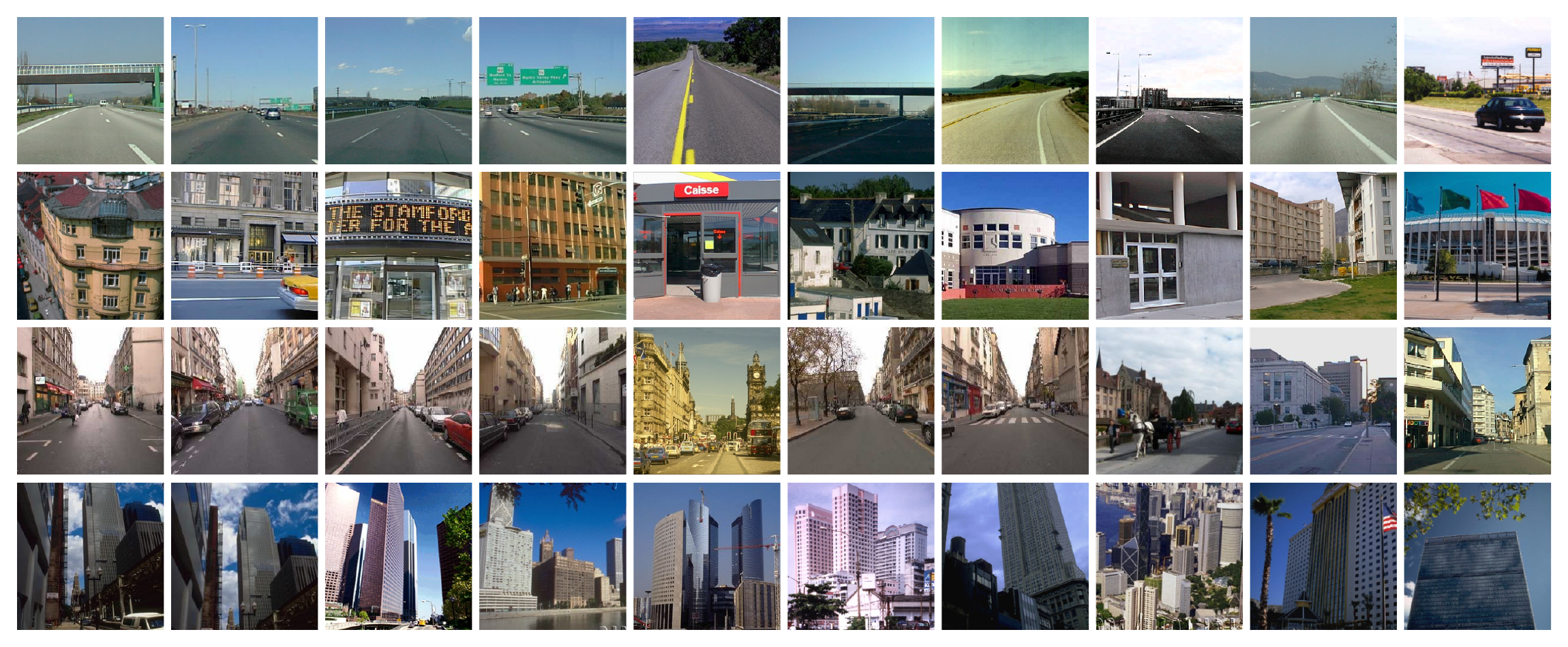}
    \caption{\texttt{LabelMe}: top-$10$ worst images  detected by the $\mathrm{WAUM}$ (with labels row-ordered from top to bottom: \texttt{highway}, \texttt{insidecity}, \texttt{street}, \texttt{tallbuilding}). Overlapping classes lead to labeling confusion and learning difficulties for both the workers and the neural network.}
    \label{fig:worse_labelme}
\end{figure}

We observe in \Cref{tab:labelme} that the $\mathrm{WAUM}$ improves the final test accuracy when combined with the CoNAL network with regularization.
Note that the \texttt{LabelMe} dataset has classes that overlap and thus lead to intrinsic ambiguities.
This is the reason why the CoNAL strategy was introduced by \citet{chu2021learning}: modeling common confusions help the network's decision, so it was expected for the CoNAL to perform well.
Combined with our $\mathrm{WAUM}$, additional gains are obtained on both metrics.
The vanilla strategy, either for aggregation or learning, can be improved using a pruning preprocessing step.
However, between the $\mathrm{AUMC}$ and the $\mathrm{WAUM}$, we show a consistent improvement on using the $\mathrm{WAUM}$ that considers weights for the workers individually.
For example, the classes \texttt{highway}, \texttt{insidecity}, \texttt{street} and \texttt{tallbuilding} (in rows) are overlapping for some tasks: some cities have streets with tall buildings, leading to confusion as shown in \Cref{fig:worse_labelme}.

\paragraph*{\texttt{Music} dataset.}

This dataset differs from \texttt{LabelMe} and \texttt{CIFAR-10H} as it consists in classifying $1000$ recordings of $30$ seconds into $K=10$ music genres.
All the $44$ workers involved voted for at least one music, resulting in up to $7$ labels per task.
Instead of classifying the original audio files, we use the associated Mel spectrograms following the methodology considered by \citet{dong2018convolutional} to retrieve an image classification setting.
Though the benefits are not as striking as before on test accuracy, the ECE is slightly improved by combining our $\mathrm{WAUM}$ with $\mathrm{CoNAL}$ as can be seen in \Cref{tab:music}.
Moreover, we show constant improvement of the test generalization performance using the $\mathrm{WAUM}$ preprocessing either in accuracy or in calibration.

Among other interesting discoveries, the $\texttt{WAUM}$ helped us detect that the music \emph{Zydeco Honky Tonk} by Buckwheat Zydeco was labeled as \texttt{classical}, \texttt{country} or \texttt{pop} by the workers, though it is a \texttt{blues} standard.
Another example is \emph{Caught in the middle} by Dio classified (with the same number of votes) as \texttt{rock}, \texttt{jazz}, or \texttt{country} though it is a \texttt{metal} song.
One last example detected: the music \emph{Patches} by Clarence Carter is stored in the \texttt{disco00020.wav} file.
The true label is supposed to be \texttt{disco}, while the workers have provided the following labels: two have chosen \texttt{rock}, two \texttt{blues}, one \texttt{pop} and another one proposed \texttt{country}.
The actual genre of this music is \texttt{country}-soul, so both the true label and five out of six workers are incorrect.

\paragraph*{WAUM sensitivity to the neural network architecture}

In the following, we explore the architecture's impact on the generalization performance using the $\mathrm{WAUM}$ preprocessing. We compare three architectures, a VGG-$16$ with two dense layers added from \citet{rodrigues2018deep}, a Resnet-$18$ and a Resnet-$34$. We show in \Cref{fig:tab_arch} that depending on the network used, performance vary, but the $\mathrm{WAUM}$ step improves generalization performance in most cases (and does not worsen it).

\paragraph{Limitations: computing the weights with many classes}
First, concerning the weights $s_i^{(j)}$ (reflecting the trust in the image/worker interaction), we rely on confusion matrices $\{\hat{\pi}^{(j)}\}_{j\in [n_\texttt{worker}]}$.
The DS model \citep{dawid_maximum_1979} can be naturally used to estimate such matrices $\pi^{(j)}\in\sR^{K\times K}$ for each worker $w_j$.
Yet, the quadratic number of parameters (w.r.t. $K$) to be estimated for each worker can create convergence issues for the vanilla DS model when $K$ is large.
But as stated in \Cref{sec:diff_aware}, any model that can estimate confusion matrices can be considered for the $\mathrm{WAUM}$'s computation.
We detail below some possible variants, that could help computing the confusion matrices used in the $\mathrm{WAUM}$ for the trust score computation.
\begin{itemize}
    \item \citet{sinha2018fast} accelerated the vanilla DS by constraining the estimated labels' distribution to be a Dirac mass. Hence, predicted labels are hard labels. This leads to worse calibration errors than vanilla DS but preserves the same accuracy.
    \item \citet{passonneau-carpenter-2014-benefits} introduced Dirichlet priors on the confusion matrices' rows and the prevalence $\rho$ to incorporate previously known information on the workers in the model (\eg from other experiments).
    \item \citet{servajean2017crowdsourcing} exploited the sparsity of the confusion matrices to cope with a large $K$.
    \item \citet{imamura2018analysis} estimated with variational inference $L \ll n_{\texttt{worker}}$ clusters of workers, constraining at most $L$ different confusion matrices. This reduces the number of parameters required  from $K^2\times n_{\texttt{worker}}$ to $K^2\times L$.
\end{itemize}

\paragraph{Pruning and \emph{i.i.d} assumption}
For the pruning at preprocessing can induce a distortion in the training data distribution.
A usual assumption made on learning problems is that the task/label pairs are \emph{i.i.d}.
However, by removing some of the hardest tasks, the new dataset $\mathcal{D}_{\text{pruned}}$ contains tasks that are not independent anymore.
We should also keep in mind that \citet{ilyas2022datamodels} have shown that in the standard datasets, the data is not \emph{i.i.d} to begin with.

\section{Conclusion}
\label{sec:conclusions}

In this paper, we investigate crowdsourcing aggregation models and how judging systems may impact generalization performance.
Most models consider the ambiguity from the workers' perspective (very few consider the difficulty of the task itself) and evaluate workers on hard tasks that might be too ambiguous to be relevant, leading to a performance drop.
Using a popular model (DS), we develop the $\mathrm{WAUM}$, a flexible feature-aware metric that can identify hard tasks and improves generalization performance over vanilla strategies and naive pruning $\mathrm{AUMC}$.
It also yields a fair evaluation of workers' abilities and supports recent research on data pruning in supervised datasets.
Independently of pruning, the $\mathrm{WAUM}$ allows identifying early the images that need extra labeling efforts or that are impossible to correctly label.

Extension of the $\mathrm{WAUM}$ to more general learning tasks (\eg top-$k$ classification) would be natural, including sequential label.
Indeed, the $\mathrm{WAUM}$ could help to identify tasks requiring additional expertise and guide how to allocate more experts/workers for such identified tasks.
Future works could adapt the $\mathrm{WAUM}$ to imbalanced crowdsourced datasets to identify potentially too ambiguous images that naturally occur in open platforms like Pl$@$ntNet\footnote{\url{https://plantnet.org/en/}}.
And in this case, a class-dependent pruning threshold quantile could be used to avoid a learning bias for classes with very few instances.

Last but not least, on the dataset side, we believe that the community would benefit from releasing a challenging dataset (such as the one by \citet{Garcin_Joly_Bonnet_Affouard_Lombardo_Chouet_Servajean_Lorieul_Salmon2021} for instance) tailored to learn in crowdsourcing settings.
Indeed, a dataset with the following properties could greatly foster future research in the field: a varying number of labels per worker, a high number of classes, and a subset with ground truth labels to test generalization performance.

\subsubsection*{Broader Impact Statement}
As this work proposes a method to prune tasks from training datasets based on human-derived data, we remind that pruning based on learning difficulty can induce a learning bias for the model.
To mitigate this, only pruning a small portion of the dataset can help avoid any class with a small number of representatives to be removed of the dataset.
Also, in this paper, we only remove tasks that are difficult to classify, we do not remove workers from the dataset.
In particular, there is no repercussion on their pay, and by only evaluating them on tasks that are not detected as ambiguous, we evaluate their abilities on fairer tasks.
Finally, during the entire procedure, all anonymity is conserved for workers, no other data than their anonymous identification number is used.


\bibliography{main}

\providecommand{\AC}{{A.-C}}\providecommand{\AM}{{A.-M}}\providecommand{\CA}{{C.-A}}\providecommand{\CH}{{C.-H}}\providecommand{\CC}{{C.-C}}\providecommand{\CJ}{{C.-J}}\providecommand{\CN}{{C.-N}}\providecommand{\CP}{{C.-P}}\providecommand{\DY}{{D.-Y}}\providecommand{\HJ}{{H.-J}}\providecommand{\HG}{{H.-G}}\providecommand{\HT}{{H.-T}}\providecommand{\HY}{{H.-Y}}\providecommand{\JA}{{J.-A}}\providecommand{\JC}{{J.-C}}\providecommand{\JF}{{J.-F}}\providecommand{\JJ}{{J.-J}}\providecommand{\JL}{{J.-L}}\providecommand{\JM}{{J.-M}}\providecommand{\JP}{{J.-P}}\providecommand{\JS}{{J.-S}}\providecommand{\JY}{{J.-Y}}\providecommand{\KC}{{K.-C}}\providecommand{\KR}{{K.-R}}\providecommand{\KW}{{K.-W}}\providecommand{\LJ}{{L.-J}}\providecommand{\MR}{{M.-R}}\providecommand{\PL}{{P.-L}}\providecommand{\RE}{{R.-E}}\providecommand{\SJ}{{S.-J}}\providecommand{\TB}{{T.-B}}\providecommand{\XR}{{X.-R}}\providecommand{\WX}{{W.-X}}\providecommand{\YX}{{Y.-X}}
\begin{thebibliography}{62}
\providecommand{\natexlab}[1]{#1}
\providecommand{\url}[1]{\texttt{#1}}
\expandafter\ifx\csname urlstyle\endcsname\relax
  \providecommand{\doi}[1]{doi: #1}\else
  \providecommand{\doi}{doi: \begingroup \urlstyle{rm}\Url}\fi

\bibitem[Aitchison(2021)]{aitchison2020statistical}
L.~Aitchison.
\newblock A statistical theory of cold posteriors in deep neural networks.
\newblock In \emph{ICLR}, 2021.

\bibitem[Albert et~al.(2012)Albert, Donnet, Guihenneuc-Jouyaux, Low-Choy,
  Mengersen, and Rousseau]{albert2012combining}
I.~Albert, S.~Donnet, C.~Guihenneuc-Jouyaux, S.~Low-Choy, K.~Mengersen, and
  J.~Rousseau.
\newblock Combining expert opinions in prior elicitation.
\newblock \emph{Bayesian Analysis}, 7\penalty0 (3):\penalty0 503--532, 2012.

\bibitem[Angelova(2004)]{angelova2004data}
A.~Angelova.
\newblock \emph{Data pruning}.
\newblock PhD thesis, California Institute of Technology, 2004.

\bibitem[Angelova et~al.(2005)Angelova, Abu{-}Mostafa, and
  Perona]{Angelova_AbuMostafa_Perona05}
A.~Angelova, Y.~S. Abu{-}Mostafa, and P.~Perona.
\newblock Pruning training sets for learning of object categories.
\newblock In \emph{CVPR}, volume~1, pp.\  494--501 vol. 1, 2005.

\bibitem[Birnbaum(1968)]{birnbaum1968some}
A.~Lord Birnbaum.
\newblock Some latent trait models and their use in inferring an examinee's
  ability.
\newblock \emph{Statistical theories of mental test scores}, 1968.

\bibitem[Camilleri \& Williams(2019)Camilleri and
  Williams]{camilleri2019extended}
M.~PJ. Camilleri and C.~KI. Williams.
\newblock The extended {Dawid-Skene} model.
\newblock In \emph{Joint European Conference on Machine Learning and Knowledge
  Discovery in Databases}, pp.\  121--136. Springer, 2019.

\bibitem[Chu et~al.(2021)Chu, Ma, and Wang]{chu2021learning}
Z.~Chu, J.~Ma, and H.~Wang.
\newblock Learning from crowds by modeling common confusions.
\newblock In \emph{AAAI}, pp.\  5832--5840, 2021.

\bibitem[Dawid \& Skene(1979)Dawid and Skene]{dawid_maximum_1979}
AP. Dawid and AM. Skene.
\newblock Maximum likelihood estimation of observer error-rates using the {EM}
  algorithm.
\newblock \emph{J. R. Stat. Soc. Ser. C. Appl. Stat.}, 28\penalty0
  (1):\penalty0 20--28, 1979.

\bibitem[Dempster et~al.(1977)Dempster, Laird, and
  Rubin]{Dempster_Laird_Rubin77}
AP. Dempster, NM. Laird, and DB. Rubin.
\newblock Maximum likelihood from incomplete data via the em algorithm.
\newblock \emph{J. R. Stat. Soc. Ser. B Stat. Methodol.}, 39\penalty0
  (1):\penalty0 1--22, 1977.

\bibitem[Dong(2018)]{dong2018convolutional}
M.~Dong.
\newblock Convolutional neural network achieves human-level accuracy in music
  genre classification.
\newblock \emph{arXiv preprint arXiv:1802.09697}, 2018.

\bibitem[Draws et~al.(2021)Draws, Rieger, Inel, Gadiraju, and
  Tintarev]{draws2021checklist}
T.~Draws, A.~Rieger, O.~Inel, Y.~Gadiraju, and N.~Tintarev.
\newblock A checklist to combat cognitive biases in crowdsourcing.
\newblock In \emph{AAAI Conference on Human Computation and Crowdsourcing},
  volume~9, pp.\  48--59, 2021.

\bibitem[Gao \& Zhou(2013)Gao and Zhou]{gao2013minimax}
G.~Gao and D.~Zhou.
\newblock Minimax optimal convergence rates for estimating ground truth from
  crowdsourced labels.
\newblock \emph{arXiv preprint arXiv:1310.5764}, 2013.

\bibitem[Garcin et~al.(2021)Garcin, Joly, Bonnet, Affouard, Lombardo, Chouet,
  Servajean, Lorieul, and
  Salmon]{Garcin_Joly_Bonnet_Affouard_Lombardo_Chouet_Servajean_Lorieul_Salmon2021}
C.~Garcin, A.~Joly, P.~Bonnet, A.~Affouard, J.-C. Lombardo, M.~Chouet,
  M.~Servajean, T.~Lorieul, and J.~Salmon.
\newblock Pl@ntnet-300k: a plant image dataset with high label ambiguity and a
  long-tailed distribution.
\newblock In \emph{Proceedings of the Neural Information Processing Systems
  Track on Datasets and Benchmarks}, 2021.

\bibitem[Garcin et~al.(2022)Garcin, Servajean, Joly, and
  Salmon]{Garcin_Servajean_Joly_Salmon22}
C.~Garcin, M.~Servajean, A.~Joly, and J.~Salmon.
\newblock Stochastic smoothing of the top-k calibrated hinge loss for deep
  imbalanced classification.
\newblock In \emph{ICML}, 2022.

\bibitem[Germain et~al.(2015)Germain, Lacasse, Laviolette, Marchand, and
  Roy]{germain2015risk}
P.~Germain, A.~Lacasse, F.~Laviolette, M.~Marchand, and JF. Roy.
\newblock Risk bounds for the majority vote: from a pac-bayesian analysis to a
  learning algorithm.
\newblock \emph{J. Mach. Learn. Res.}, 16:\penalty0 787--860, 2015.

\bibitem[Guan et~al.(2017)Guan, Gulshan, Dai, and Hinton]{doctornet}
MY. Guan, V.~Gulshan, AM. Dai, and GE. Hinton.
\newblock Who said what: Modeling individual labelers improves classification.
\newblock \emph{CoRR}, abs/1703.08774, 2017.

\bibitem[Guo et~al.(2017)Guo, Pleiss, Sun, and
  Weinberger]{guo_calibration_2017}
C.~Guo, G.~Pleiss, Y.~Sun, and KQ. Weinberger.
\newblock On calibration of modern neural networks.
\newblock In \emph{ICML}, pp.\  1321, 2017.

\bibitem[Han et~al.(2019)Han, Luo, and Wang]{han2019deep}
J.~Han, P.~Luo, and X.~Wang.
\newblock Deep self-learning from noisy labels.
\newblock In \emph{ICCV}, pp.\  5138--5147, 2019.

\bibitem[He et~al.(2016)He, Zhang, Ren, and Sun]{he2016deep}
K.~He, X.~Zhang, S.~Ren, and J.~Sun.
\newblock Deep residual learning for image recognition.
\newblock In \emph{CVPR}, pp.\  770--778, 2016.

\bibitem[Hoang et~al.(2021)Hoang, Faucon, Jungo, Volodin, Papuc, Liossatos,
  Crulis, Tighanimine, Constantin, Kucherenko, Maurer, Grimberg, Nitu, Vossen,
  Rouault, and El{-}Mhamdi]{hoang2021tournesol}
LN. Hoang, L.~Faucon, A.~Jungo, S.~Volodin, D.~Papuc, O.~Liossatos, B.~Crulis,
  M.~Tighanimine, I.~Constantin, A.~Kucherenko, A.~Maurer, F.~Grimberg,
  V.~Nitu, C.~Vossen, S.~Rouault, and EM~El{-}Mhamdi.
\newblock Tournesol: A quest for a large, secure and trustworthy database of
  reliable human judgments.
\newblock \emph{arXiv preprint arXiv:2107.07334}, 2021.

\bibitem[Ibrahim et~al.(2019)Ibrahim, Fu, Kargas, and
  Huang]{ibrahim2019crowdsourcing}
S.~Ibrahim, X.~Fu, N.~Kargas, and K.~Huang.
\newblock Crowdsourcing via pairwise co-occurrences: Identifiability and
  algorithms.
\newblock \emph{Advances in neural information processing systems}, 32, 2019.

\bibitem[Ilyas et~al.(2022)Ilyas, Park, Engstrom, Leclerc, and
  Madry]{ilyas2022datamodels}
A.~Ilyas, SM. Park, L.~Engstrom, G.~Leclerc, and A.~Madry.
\newblock Datamodels: Predicting predictions from training data.
\newblock \emph{arXiv preprint arXiv:2202.00622}, 2022.

\bibitem[Imamura et~al.(2018)Imamura, Sato, and Sugiyama]{imamura2018analysis}
H.~Imamura, I.~Sato, and M.~Sugiyama.
\newblock Analysis of minimax error rate for crowdsourcing and its application
  to worker clustering model.
\newblock In \emph{ICML}, pp.\  2147--2156, 2018.

\bibitem[James(1998)]{james1998majority}
GM. James.
\newblock \emph{Majority vote classifiers: theory and applications}.
\newblock PhD thesis, Stanford University, 1998.

\bibitem[Jamison \& Gurevych(2015)Jamison and Gurevych]{jamison2015noise}
E.~Jamison and I.~Gurevych.
\newblock Noise or additional information? leveraging crowdsource annotation
  item agreement for natural language tasks.
\newblock In \emph{Conference on Empirical Methods in Natural Language
  Processing}, pp.\  291--297, 2015.

\bibitem[Jiang et~al.(2012)Jiang, Osl, Kim, and
  Ohno-Machado]{jiang2012calibrating}
X.~Jiang, M.~Osl, J.~Kim, and L.~Ohno-Machado.
\newblock Calibrating predictive model estimates to support personalized
  medicine.
\newblock \emph{J. Am. Med. Inform. Assoc.}, 19\penalty0 (2):\penalty0
  263--274, 2012.

\bibitem[Ju et~al.(2018)Ju, Bibaut, and Van~der Laan]{ju2018relative}
C.~Ju, A.~Bibaut, and M.~Van~der Laan.
\newblock The relative performance of ensemble methods with deep convolutional
  neural networks for image classification.
\newblock \emph{J. Appl. Stat.}, 45\penalty0 (15):\penalty0 2800--2818, 2018.

\bibitem[Kamar et~al.(2015)Kamar, Kapoor, and Horvitz]{kamar2015identifying}
E.~Kamar, A.~Kapoor, and E.~Horvitz.
\newblock Identifying and accounting for task-dependent bias in crowdsourcing.
\newblock In \emph{Third AAAI Conference on Human Computation and
  Crowdsourcing}, 2015.

\bibitem[Khattak(2017)]{khattak_toward_2017}
FK. Khattak.
\newblock \emph{Toward a Robust and Universal Crowd Labeling Framework}.
\newblock PhD thesis, Columbia University, 2017.

\bibitem[Kim \& Ghahramani(2012)Kim and Ghahramani]{pmlr-v22-kim12}
H-C. Kim and Z.~Ghahramani.
\newblock Bayesian classifier combination.
\newblock In \emph{AISTATS}, volume~22, pp.\  619--627, 2012.

\bibitem[Krizhevsky \& Hinton(2009)Krizhevsky and
  Hinton]{krizhevsky2009learning}
A.~Krizhevsky and G.~Hinton.
\newblock Learning multiple layers of features from tiny images.
\newblock Technical report, University of Toronto, 2009.

\bibitem[Kumar et~al.(2019)Kumar, Liang, and Ma]{kumar2019verified}
A.~Kumar, PS. Liang, and T.~Ma.
\newblock Verified uncertainty calibration.
\newblock In \emph{NeurIPS}, volume~32, 2019.

\bibitem[Lapin et~al.(2016)Lapin, Hein, and Schiele]{lapin2016loss}
M.~Lapin, M.~Hein, and B.~Schiele.
\newblock Loss functions for top-k error: Analysis and insights.
\newblock In \emph{CVPR}, pp.\  1468--1477, 2016.

\bibitem[Li \& Varshney(2017)Li and Varshney]{li2017does}
Q.~Li and PK. Varshney.
\newblock Does confidence reporting from the crowd benefit crowdsourcing
  performance?
\newblock In \emph{International Workshop on Social Sensing}, pp.\  49--54,
  2017.

\bibitem[Ma \& Olshevsky(2020)Ma and Olshevsky]{ma2020adversarial}
Q.~Ma and A.~Olshevsky.
\newblock Adversarial crowdsourcing through robust rank-one matrix completion.
\newblock In \emph{NeurIPS}, volume~33, pp.\  21841--21852, 2020.

\bibitem[Ma et~al.(2020)Ma, Olshevsky, Saligrama, and
  Szepesvari]{ma2020gradient}
Y.~Ma, A.~Olshevsky, V.~Saligrama, and C.~Szepesvari.
\newblock Gradient descent for sparse rank-one matrix completion for
  crowd-sourced aggregation of sparsely interacting workers.
\newblock \emph{J. Mach. Learn. Res.}, 21\penalty0 (1):\penalty0 5245--5280,
  2020.

\bibitem[M{\"u}ller et~al.(2019)M{\"u}ller, Kornblith, and
  Hinton]{muller2019does}
R.~M{\"u}ller, S.~Kornblith, and GE. Hinton.
\newblock When does label smoothing help?
\newblock \emph{NeurIPS}, 32, 2019.

\bibitem[Müller \& Markert(2019)Müller and Markert]{muller2019}
NM. Müller and K.~Markert.
\newblock Identifying mislabeled instances in classification datasets.
\newblock In \emph{2019 International Joint Conference on Neural Networks
  (IJCNN)}, pp.\  1--8, 2019.

\bibitem[Northcutt et~al.(2021{\natexlab{a}})Northcutt, Athalye, and
  Mueller]{northcutt_pervasive_2021}
C.~Northcutt, A.~Athalye, and J.~Mueller.
\newblock Pervasive label errors in test sets destabilize machine learning
  benchmarks.
\newblock In \emph{Proceedings of the Neural Information Processing Systems
  Track on Datasets and Benchmarks}, 2021{\natexlab{a}}.

\bibitem[Northcutt et~al.(2021{\natexlab{b}})Northcutt, Jiang, and
  Chuang]{northcutt_confident_2021}
C.~Northcutt, L.~Jiang, and I.~Chuang.
\newblock Confident learning: Estimating uncertainty in dataset labels.
\newblock \emph{J. Artif. Intell. Res.}, 70:\penalty0 1373--1411,
  2021{\natexlab{b}}.

\bibitem[Oyama et~al.(2013)Oyama, Baba, Sakurai, and
  Kashima]{oyama2013accurate}
S.~Oyama, Y.~Baba, Y.~Sakurai, and H.~Kashima.
\newblock Accurate integration of crowdsourced labels using workers'
  self-reported confidence scores.
\newblock In \emph{IJCAI}, 2013.

\bibitem[Park \& Caragea(2022)Park and Caragea]{park2022calibration}
SY. Park and C.~Caragea.
\newblock On the calibration of pre-trained language models using mixup guided
  by area under the margin and saliency.
\newblock In \emph{ACML}, pp.\  5364--5374, 2022.

\bibitem[Passonneau \& Carpenter(2014)Passonneau and
  Carpenter]{passonneau-carpenter-2014-benefits}
RJ. Passonneau and B.~Carpenter.
\newblock The benefits of a model of annotation.
\newblock \emph{Transactions of the Association for Computational Linguistics},
  2:\penalty0 311--326, 2014.

\bibitem[Paszke et~al.(2019)Paszke, Gross, Massa, Lerer, Bradbury, Chanan,
  Killeen, Lin, Gimelshein, Antiga, Desmaison, Kopf, Yang, DeVito, Raison,
  Tejani, Chilamkurthy, Steiner, Fang, Bai, and Chintala]{pytorch}
A.~Paszke, S.~Gross, F.~Massa, A.~Lerer, J.~Bradbury, G.~Chanan, T.~Killeen,
  Z.~Lin, N.~Gimelshein, L.~Antiga, A.~Desmaison, A.~Kopf, E.~Yang, Z.~DeVito,
  M.~Raison, A.~Tejani, S.~Chilamkurthy, B.~Steiner, L.~Fang, J.~Bai, and
  S.~Chintala.
\newblock Pytorch: An imperative style, high-performance deep learning library.
\newblock In \emph{NeurIPS}, pp.\  8024--8035, 2019.

\bibitem[Paul et~al.(2021)Paul, Ganguli, and Dziugaite]{NEURIPS2021_ac56f8fe}
M.~Paul, S.~Ganguli, and GK. Dziugaite.
\newblock Deep learning on a data diet: Finding important examples early in
  training.
\newblock In \emph{NeurIPS}, volume~34, pp.\  20596--20607, 2021.

\bibitem[Pedregosa et~al.(2011)Pedregosa, Varoquaux, Gramfort, Michel, Thirion,
  Grisel, Blondel, Prettenhofer, Weiss, Dubourg, Vanderplas, Passos,
  Cournapeau, Brucher, Perrot, and Duchesnay]{scikit-learn}
F.~Pedregosa, G.~Varoquaux, A.~Gramfort, V.~Michel, B.~Thirion, O.~Grisel,
  M.~Blondel, P.~Prettenhofer, R.~Weiss, V.~Dubourg, J.~Vanderplas, A.~Passos,
  D.~Cournapeau, M.~Brucher, M.~Perrot, and E.~Duchesnay.
\newblock Scikit-learn: Machine learning in {P}ython.
\newblock \emph{J. Mach. Learn. Res.}, 12:\penalty0 2825--2830, 2011.

\bibitem[Peterson et~al.(2019)Peterson, Battleday, Griffiths, and
  Russakovsky]{peterson_human_2019}
JC. Peterson, RM. Battleday, TL. Griffiths, and O.~Russakovsky.
\newblock Human uncertainty makes classification more robust.
\newblock In \emph{ICCV}, pp.\  9617--9626, 2019.

\bibitem[Pleiss et~al.(2020)Pleiss, Zhang, Elenberg, and
  Weinberger]{pleiss_identifying_2020}
G.~Pleiss, T.~Zhang, ER. Elenberg, and KQ. Weinberger.
\newblock Identifying mislabeled data using the area under the margin ranking.
\newblock In \emph{NeurIPS}, 2020.

\bibitem[Raju et~al.(2021)Raju, K., and HL.]{ada_pruning}
Ravi~S. Raju, Daruwalla K., and Mikko HL.
\newblock Accelerating deep learning with dynamic data pruning.
\newblock \emph{CoRR}, abs/2111.12621, 2021.

\bibitem[Raykar \& Yu(2011)Raykar and Yu]{raykar_ranking_2011}
VC. Raykar and S.~Yu.
\newblock Ranking annotators for crowdsourced labeling tasks.
\newblock In \emph{NeurIPS}, pp.\  1809--1817, 2011.

\bibitem[Rodrigues \& Pereira(2018)Rodrigues and Pereira]{rodrigues2018deep}
F.~Rodrigues and F.~Pereira.
\newblock Deep learning from crowds.
\newblock In \emph{AAAI}, volume~32, 2018.

\bibitem[Rodrigues et~al.(2014)Rodrigues, Pereira, and
  Ribeiro]{rodrigues2014gaussian}
F.~Rodrigues, F.~Pereira, and B.~Ribeiro.
\newblock Gaussian process classification and active learning with multiple
  annotators.
\newblock In \emph{ICML}, pp.\  433--441. PMLR, 2014.

\bibitem[Servajean et~al.(2017)Servajean, Joly, Shasha, Champ, and
  Pacitti]{servajean2017crowdsourcing}
M.~Servajean, A.~Joly, D.~Shasha, J.~Champ, and E.~Pacitti.
\newblock Crowdsourcing thousands of specialized labels: A {Bayesian} active
  training approach.
\newblock \emph{IEEE Transactions on Multimedia}, 19\penalty0 (6):\penalty0
  1376--1391, 2017.

\bibitem[Sinha et~al.(2018)Sinha, Rao, and Balasubramanian]{sinha2018fast}
VB. Sinha, S.~Rao, and VN. Balasubramanian.
\newblock Fast {Dawid-Skene}: A fast vote aggregation scheme for sentiment
  classification.
\newblock \emph{arXiv preprint arXiv:1803.02781}, 2018.

\bibitem[Snow et~al.(2008)Snow, O'Connor, and Jurafsky]{snow_cheap_2008}
R.~Snow, B.~O'Connor, and A.~Jurafsky, D.and~Ng.
\newblock Cheap and fast - but is it good? evaluating non-expert annotations
  for natural language tasks.
\newblock In \emph{Conference on Empirical Methods in Natural Language
  Processing}, pp.\  254--263. Association for Computational Linguistics, 2008.

\bibitem[Sorscher et~al.(2022)Sorscher, Geirhos, Shekhar, Ganguli, and
  Morcos]{sorscher2022beyond}
B.~Sorscher, R.~Geirhos, S.~Shekhar, S.~Ganguli, and AS. Morcos.
\newblock Beyond neural scaling laws: beating power law scaling via data
  pruning.
\newblock \emph{arXiv preprint arXiv:2206.14486}, 2022.

\bibitem[Wen et~al.(2021)Wen, Jerfel, Muller, Dusenberry, Snoek,
  Lakshminarayanan, and Tran]{wen2020combining}
Y.~Wen, G.~Jerfel, R.~Muller, WM. Dusenberry, J.~Snoek, B.~Lakshminarayanan,
  and D.~Tran.
\newblock Combining ensembles and data augmentation can harm your calibration.
\newblock In \emph{ICLR}, 2021.

\bibitem[Whitehill et~al.(2009)Whitehill, Wu, Bergsma, Movellan, and
  Ruvolo]{whitehill_whose_2009}
J.~Whitehill, T.~Wu, J.~Bergsma, J.~Movellan, and P.~Ruvolo.
\newblock Whose vote should count more: Optimal integration of labels from
  labelers of unknown expertise.
\newblock In \emph{NeurIPS}, volume~22, 2009.

\bibitem[Yang \& Koyejo(2020)Yang and Koyejo]{yang2020consistency}
F.~Yang and S.~Koyejo.
\newblock On the consistency of top-k surrogate losses.
\newblock In \emph{ICML}, pp.\  10727--10735, 2020.

\bibitem[Zhang et~al.(2018)Zhang, Ciss{\'{e}}, Dauphin, and
  Lopez{-}Paz]{zhang2017mixup}
H.~Zhang, M.~Ciss{\'{e}}, YN. Dauphin, and D.~Lopez{-}Paz.
\newblock mixup: Beyond empirical risk minimization.
\newblock In \emph{ICLR}, 2018.

\bibitem[Zhong et~al.(2021)Zhong, Cui, Liu, and Jia]{zhong2021improving}
A.~Zhong, J.~Cui, S.~Liu, and J.~Jia.
\newblock Improving calibration for long-tailed recognition.
\newblock In \emph{CVPR}, pp.\  16489--16498, 2021.

\bibitem[Zhou et~al.(2015)Zhou, Liu, Platt, Meek, and
  Shah]{zhou2015regularized}
D.~Zhou, Q.~Liu, JC. Platt, C.~Meek, and N.~Shah.
\newblock Regularized minimax conditional entropy for crowdsourcing.
\newblock \emph{arXiv preprint arXiv:1503.07240}, 2015.

\end{thebibliography}
\bibliographystyle{tmlr}

\appendix
\clearpage
\section{Popular label aggregation techniques}
\label{subsec:aggregation_techniques}

Several aggregation techniques can transform crowdsourced labels into probability distributions (soft labels).
For any $d\in \mathbb{N}$ and $z \in (0,\infty)^d$, let $\mathrm{Norm}(z) \in (0,\infty)^d$ be the vector defined by $\forall i \in [d]$, $\mathrm{Norm}(z)_i = {z_i}/{\sum_{i'=1}^d z_{i'}}$.

\subsection{Naive soft (NS)}
\label{app:soft}
The naive soft (NS) labeling is simply the empirical distribution of the answered votes:
\begin{align}
    \forall x_i \in \gD_{\text{train}}, \quad \hat y_i^{\textrm{NS}} = \mathrm{Norm}(\tilde y_i), \quad \text{where } \tilde y_i = \Big(\sum\limits_{j\in\gA(x_i)} \ind_{\{y^{(j)}_i=k\}} \Big)_{k\in[K]} \enspace.
    \label{eq:naive_soft}
\end{align}
\subsection{Majority voting (MV)}
\label{app:mv}

Majority voting (MV) outputs the most answered label:
\begin{align}
    \forall x_i \in \gD_{\text{train}},\quad
    \hat y_i^{\textrm{MV}} = \argmax_{k\in [K]}
    \Big(\sum\limits_{j\in\gA(x_i)} \ind_{\{y^{(j)}_i=k\}} \Big) \enspace.
    \label{eq:majority_vote}
\end{align}
\subsection{Dawid and Skene (DS)}
\label{app:ds}

The Dawid and Skene \citep{dawid_maximum_1979} model aggregates answers and evaluates the workers' confusion matrix to observe where their expertise lies exactly.
Let us introduce $\rho_\ell$ the prevalence of each label in the dataset (\ie $\mathbb{P}(y_i^\star=\ell)$), the probability that a task drawn at random is labeled $\ell\in[K]$.
Following standard notations, we also write $\{T_{i,\ell},\ i\in[n_\texttt{task}]\}$ the indicator variables for task $i$, that is $T_{i,\ell} = 1$ if the true label for task $i$ is $\ell$ (\ie $y_i^\star=\ell$) and zero otherwise.
Finally, let $\pi^{(j)}_{\ell, k}$ be the probability for worker $j$ to select label $k$ when $y^\star=\ell$.
The model's likelihood reads:
\begin{align}\label{eq:ds}
    \prod_{i\in [n_{\texttt{task}}]}\prod_{\ell \in [K]}\bigg[\rho_\ell\prod_{j\in [n_{\texttt{worker}}]}
    \prod_{k\in [K]}\big(\pi^{(j)}_{\ell, k}\big)^{\mathds{1}_{\{y_i^{(j)}=k\}}}
    \bigg]^{T_{i\ell} \phantom{\mathds{1}_{\{y_i^\star=\ell\}}}} \enspace.
\end{align}
To maximize the likelihood, we use the EM algorithm \citep{Dempster_Laird_Rubin77} to estimate the parameters $\pi^{(j)}_{\ell, k}$ and $\rho_\ell$, using $(T_{i,\cdot})_{i \in [n_{\texttt{task}}]}$ as latent variables.
Our implementation of the EM algorithm is given in \Algref{alg:DS_EM}.
The convergence criterion we use in practice is that the likelihood has not decreased more than $\epsilon>0$ between two iterations.
By default, $\epsilon$ is set to $10^{-6}$, and the EM algorithm stops at iteration $t\in\mathbb{N}$ if $\big|\text{Likelihood}_{t} - \text{Likelihood}_{t+1}\big| < \varepsilon.$

\begin{algorithm}[tb]
    \caption{DS (EM version)}\label{alg:DS_EM}
\textbf{Input}: $\gD_{\text{train}}$: crowdsourced dataset \\
\textbf{Output}: $(\hat{y}_i^{\textrm{DS}})_{i\in[n_\texttt{task}]} = (\hat{T}_{i,\cdot})_{i\in[n_\texttt{task}]}$: estimated soft labels and $\{\hat{\pi}^{(j)}\}_{j\in[n_\texttt{worker}]}$: estimated confusion matrices
\begin{algorithmic}[1]
\STATE \textbf{Initialization:} $\forall i\in[n_\texttt{task}], \forall \ell\in[K],\ \hat{T}_{i,\ell} = \frac{1}{|\gA(x_i)|} {\sum_{j \in \gA(x_i)} \ind_{\{y_i^{(j)}=\ell\}} }$
\WHILE{Likelihood not converged}
    \STATE Get $\hat{\pi}$ and $\hat\rho$ assuming $\hat{T}$s are known%
    \STATE$ \forall (\ell, k)\in [K]^2,\ \hat{\pi}^{(j)}_{\ell, k} \gets
        \frac{\sum_{i\in[n_\texttt{task}]}\hat{T}_{i,\ell} \cdot \ind_{\{y_i^{(j)}=k\}}}{\sum_{k'\in[K]}\sum_{i'\in[n_\texttt{task}]}\hat{T}_{i',\ell} \cdot \ind_{\{y_{i'}^{(j)}= k'\}}}$

    \STATE $\forall \ell \in [K],\ \hat\rho_\ell\gets \frac{1}{n_\texttt{task}} {\sum_{i \in [n_\texttt{task}]} \hat{T}_{i,\ell}}$

    \STATE Estimate $\hat T$s knowing $\hat\pi$ and $\hat\rho$
    \STATE $\forall (i,\ell),
        \in[n_{\texttt{task}}] \times [K],
        \hat T_{i \ell}
        \gets
        \frac{\prod_{j\in \mathcal{A}(x_i)}\prod_{k\in[K]} \hat\rho_\ell \cdot \big(\hat\pi^{(j)}_{\ell, k}\big)^{\mathds{1}_{\{y_i^{(j)}=k\}}}}
        {\sum_{\ell' \in [K]}\prod_{j'\in \mathcal{A}(x_i)}\prod_{k' \in [K]} \hat\rho_{\ell'}\cdot \big(\hat\pi^{(j')}_{\ell' k'}\big)^{\mathds{1}_{\{y_i^{(j')}=k'\}}}}$
\ENDWHILE
\end{algorithmic}
\end{algorithm}

\subsection{Weighted Dawid and Skene (WDS)}
\label{app:WDS}

Let us run the DS model to get estimated confusion matrices $\hat{\pi}^{(j)} \in \R^{K\times K}$ for $j\in [n_\texttt{worker}]$.
Now, remind that for a given worker $j\in[n_{\texttt{worker}}]$ and a class $k\in[K]$, the term $\hat{\pi}^{(j)}_{k,k}$ estimate the probability for worker $w_{j}$ to recognize a task whose true label is $k$.
We use this term as a trust score and define the WDS soft label as
\begin{align}\label{eq:wds}
    \forall x_i \in \gD_{\text{train}}, \quad \hat y_i^{\textrm{WDS}} = \mathrm{Norm}(\tilde y_i), \quad \text{ with } \quad \tilde y_i = \Big(\sum\limits_{j\in\gA(x_i)} \hat{\pi}^{(j)}_{k,k}\ind_{\{y^{(j)}_i=k\}} \Big)_{k\in[K]} \enspace.
\end{align}

\subsection{Generative model of Labels, Abilities, and Difficulties (GLAD)}
\label{app:glad}

We recall the $\mathrm{GLAD}$ \citep{whitehill_whose_2009} algorithm in the binary setting.
A modeling assumption is that the $j$-th worker labels correctly the $i$-th task with probability given by
\begin{align}\label{eq:glad}
    \sP(y_i^{(j)}=y_i^\star|\alpha_j,\beta_i) = \frac{1}{1+e^{-\alpha_j\beta_i}} \enspace,
\end{align}
with $\alpha_j\in\sR$ the worker's expertise: $\alpha_j < 0$ implies misunderstanding, $\alpha_j=0$ an impossibility to separate the two classes and $\alpha_j>0$ a valuable expertise.
The coefficient $1/\beta_i\in\sR_+$ represents the task's intrinsic difficulty: if $1/\beta_i\to 0$ the task is trivial; on the other side when $1/\beta_i\to +\infty$ the task is very ambiguous.
Parameters $(\alpha_j)_{j\in[n_\texttt{worker}]}$ and $(\beta_i)_{i\in[n_\texttt{task}]}$ are estimated using an EM algorithm as described in \Algref{ag:GLAD}.

The auxiliary function for the binary $\mathrm{GLAD}$ model is:
\begin{align}
    \label{eq:auxi_glad}
    Q(\alpha,\beta) & = \E[\log \sP(\{y_i^{(j)}\}_{ij}, \{y^\star_i\}_{i})]
    = \sum_i \E[\log \sP(y_i^\star)] + \sum_{ij} \E[\log \sP(y_i^{(j)}|y_i^\star,\alpha_j,\beta_i)] \enspace.
\end{align}
An extension to the multiclass setting is given by \citet{whitehill_whose_2009} under the following assumption: the distribution over all incorrect labels is supposed uniform.
In this setting, the model assumption from \Cref{eq:glad} still holds and
\begin{align*}
    \forall k\neq y_i^\star,\ \mathbb{P}(y_i^{(j)}=k|\alpha_j,\beta_i) = \frac{1}{K-1}\left(1-\frac{1}{1+e^{-\alpha_j\beta_i}}\right) \enspace.
\end{align*}
However, this is not verified in many practical cases, as can be seen for example in \Cref{subfig:example_cat} where the \texttt{cat} label is only mistaken \texttt{deer} and not with other ones.
We have used the implementation from \url{https://github.com/notani/python-glad} to evaluate the $\mathrm{GLAD}$ performance in our experiments.
The maximization of the function $Q$ \wrt $\alpha$ and $\beta$ is performed using a conjugate gradient solver.
The initial parameters are all set to $1$.

\begin{algorithm}[tb]
   \caption{$\mathrm{GLAD}$ (EM version)}\label{ag:GLAD}
\textbf{Input}: $\gD_{\text{train}}$: crowdsourced dataset \\
\textbf{Output}:$\alpha=\{\alpha_j\}_{j\in [n_\texttt{worker}]}$: worker abilities, $\beta=\{\beta_i\}_{i\in [n_\texttt{task}]}$: task difficulties, aggregated labels
\begin{algorithmic}[1]
\WHILE{Likelihood not converged}
    \STATE Estimate probability of $y_i^\star$
    \STATE $\forall i \in [n_{\texttt{task}}],\ \sP(y_i^\star|\{y_i^{(j)}\}_{i},\alpha,\beta_i)\propto \sP(y_i^\star)\prod_j \sP(y_i^{(j)}|y_i^\star,\alpha_j,\beta_i)$
    \STATE Maximization step
    \STATE Maximize auxiliary function $Q(\alpha,\beta)$ in \Cref{eq:auxi_glad} \wrt $\alpha$ and $\beta$
\ENDWHILE
\end{algorithmic}
\end{algorithm}

\subsection{CrowdLayer and its matrix weights strategy (MW)}
\label{subsec:crowdlayer}

From \citep{rodrigues2018deep}, CrowdLayer is an end-to-end strategy in the crowdsourcing setting.
From the output of a neural network, a new layer called \emph{crowd layer} is added to take into account worker specificities.
The main classifier thus becomes globally shared, and the new layer is the only worker-aware layer.
As multiple variants of CrowdLayer can exist, we only considered in this paper the matrix weights (MW) strategy that is akin to the DS model.
Denoting $z=f(x_i)$ the output of the neural network classifier $f$ for a given task $x_i$ labeled by a worker $w_j$, the added layer multiplies $z$ by a matrix of weights $W^j \in\mathbb{R}^{K\times K}$.
This matrix of weights per worker takes into account the local confusion of each worker.
In practice, the forward pass $F$ on a task $x_i$ annotated by worker $w_j$ using CrowdLayer computes $F(x_i, w_j) = W^j\sigma(f(x_i))$.

\subsection{Common Noise Adaptation Layers (CoNAL)}
\label{subsec:conal}

CrowdLayer takes into account worker-specific confusion matrices.
CoNAL \citep{chu2021learning} generalizes this setting by creating a global confusion matrix $W^g\in\mathbb{R}^{K\times K}$ in addition to the local ones $W^j\in\mathbb{R}^{K\times K}$ for $j\in[n_{\texttt{worker}}]$ working all together with the classifier $f$.
Given a worker $w_j$, the confusion is global with weight $\omega_i^j$ and local with weight $1-\omega_i^j$.
The final distribution output used to compute the loss is given by:
\begin{align*}
    p_{\text{out}}(x_i, w_j) = \omega _i^j W^g f(x_i) + (1-\omega_i^j)W^j f(x_i) \enspace.
\end{align*}
As is, CoNAL local matrices tend to aggregate themselves onto the global matrix.
To avoid this phenomenon, a regularization term in the loss can be added as leading to the final loss:
\begin{align*}
    \mathcal{L}(W^g, \{W^j\}_{j\in [n_\texttt{worker}]}) = \frac{1}{n_\texttt{task}} \sum_{i\in [n_\texttt{task}]}\sum_{j\in [n_\texttt{worker}]} \mathrm{H}\Big(y_i^{(j)}, p_{\text{out}}(x_i, w_j)\Big) - \lambda \sum_{j\in [n_\texttt{worker}]} \| W^g - W^j\|_2 \enspace,
\end{align*}
with $\lambda$ the regularization hyperparameter and $\mathrm{H}$ the crossentropy loss. The larger $\lambda$, the farther local confusion weights are from the shared confusion.

\section{AUM and WAUM additional details}
\label{sec:aum_and_waum_additional_details}

\begin{algorithm}[h]
    \caption{$\textrm{worker-wise WAUM}$.}
    \label{alg:WAUM}
    \begin{algorithmic}[1]
        
    \STATE \textbf{Input:} $\mathcal{D}_{\text{train}}$: tasks and crowdsourced labels, $\alpha\in[0,1]$: proportion of training points pruned, $T\in \sN$: number of epochs, $\texttt{Est}$: Estimation procedure for the confusion matrices
    
    \STATE\textbf{Initialization:} Get confusion matrices $\{\hat{\pi}^{(j)}\}_{j\in[n_\texttt{worker}]}$ from \texttt{Est} ($=\mathrm{DS}$ by default)

    \FOR{$j\in [n_\texttt{worker}]$}\label{alg:loop_worker_begin}
    \FOR{$T$ \text{epochs}}
    \STATE\textbf{Train} a neural network for $T$ epochs on $\mathcal{D}_{\text{train}}^{(j)}=\left\{\big( x_i, y_i^{(j)}\big)
        \text{ for } i \in\gT(w_j)\right\}$
    \ENDFOR

    \STATE Get $\mathrm{AUM}(x_i, y_i^{(j)}; \mathcal{D}_{\text{train}}^{(j)})$
    using \Cref{eq:Margin_WAUM} \;

    \STATE Get \textbf{trust scores} $s^{(j)}(x_i)$ using
    \Cref{eq:trust_factor}
    \ENDFOR

    \FOR{each task $x\in\gX_\mathrm{train}$}
        \STATE Compute $\mathrm{WAUM}(x)$ using \Cref{eq:WAUM}
    \ENDFOR
    \STATE Get $q_{\alpha}$ the \textbf{quantile threshold} of order
    $\alpha$ of $(\mathrm{WAUM}(x_i))_{i\in[n_\texttt{task}]}$

    \STATE Define $\mathcal{D}_{\text{pruned}} =
        \left\{\Big( x_i, \big(y_i^{(j)}\big)_{j\in\gA(x_i)}\Big):
        \mathrm{WAUM}(x_i) \geq q_\alpha \text{ for }
        i\in[n_\texttt{task}]\right\}$

    \end{algorithmic}
\end{algorithm}

\subsection{Unstacking workers answers in the WAUM: the worker-wise WAUM}
\label{subsec:limitation_WAUM}

In \Algref{alg:WAUMstack}, the $\mathrm{WAUM}$ requires training a classifier directly from all votes.
If the crowdsourcing experiment generates many answers per worker, for example when each worker answers all the tasks, we can modify \Algref{alg:WAUMstack} to train one classifier per worker for $T$ epochs instead of a single one \emph{à la} \citet{doctornet}.
This means that each classifier is only trained on $\mathcal{D}^{(j)}:=\{(x_i, y_i^{(j)})\}_{i\in [n_{\texttt{task}}]}$ to compute the $\mathrm{AUM}$ of the tasks answered.
We refer to this as the \emph{worker-wise} $\mathrm{WAUM}$ and give the full algorithm in \Algref{alg:WAUM}.
By doing so, the network trained for a given worker is not influenced by the answers of the other workers.
Hence, the $\mathrm{AUM}$ computed by this $\textrm{worker-wise WAUM}$ is independent across workers (assuming workers are answering independently).
One downside of this worker-wise application is its training cost that increases drastically.
Where the vanilla $\mathrm{WAUM}$ adds a cost of $T$ epochs before training to identify ambiguous tasks, \emph{worker-wise} $\mathrm{WAUM}$ adds a cost of $T\times n_{\texttt{worker}}$ epochs.

In the simulated examples we propose, we provide the results for the $\textrm{worker-wise WAUM}$, yet in such simulated cases with many labels per task, the results do not differ much from the $\textrm{WAUM}$; see for instance \Cref{tab:res_4classes_upto5}.

\subsection{AUM computation in practice.}
\label{subsec:AUM_in_practice}

We recall in \Algref{alg:AUM} how to compute the $\mathrm{AUM}$ in practice for a given training set $\mathcal{D}_{\text{train}}$.
This step is used within the $\mathrm{WAUM}$ (label aggregation step).
Overall, \wrt training a model, computing the $\mathrm{AUM}$ requires an additional cost: $T$ training epochs are needed to record the margins' evolution for each task.
This usually represents less than twice the original time budget.
We recall that $\sigma^{(t)}(x_i)$ is the softmax output of the predicted scores for the task $x_i$ at iteration $t$.

\begin{algorithm}[htb]
    \caption{$\mathrm{AUM}$ algorithm}\label{alg:AUM}
    \begin{algorithmic}
    \STATE \textbf{Input:} $\mathcal{D}_{\text{train}}=(x_i, y_i)_{i\in[n_\texttt{task}]}$: training set with $n_\texttt{task}$ task/label couples, $T\in \sN$: number of epochs 
    \FOR{$t=1,\dots,T$}
    \STATE \textbf{Train} the neural network for the $t^{th}$ epoch, using $\mathcal{D}_{\text{train}}$\;
    \FOR{$i \in [n_{\texttt{task}}]$}
    \STATE\textbf{Record softmax} output $\sigma^{(t)}(x_i)\in\Delta_{K-1}$\;
    \STATE\textbf{Compute margin} $M^{(t)}(x_i, y_i)=\sigma^{(t)}_{y_i}(x_i) - \sigma^{(t)}_{[2]}(x_i)$\;
    \ENDFOR
    \ENDFOR
    \STATE $\forall i\in[n_{\texttt{task}}],\ \mathrm{AUM}(x_i, y_i;\mathcal{D}_{\text{train}})=\frac{1}{T}\sum_{t\in[T]}M^{(t)}(x_i, y_i)$
\end{algorithmic}
\end{algorithm}

\section{Reminder on the calibration of neural networks}
\label{app:calibration}
Hereafter, we propose a reminder on neural networks calibration metric defined in \citet{guo_calibration_2017}.
Calibration measures the discrepancy between the accuracy and the confidence of a network.
In this context, we say that a neural network is perfectly calibrated if it is as accurate as it is confident.
For each task $x \in \gX_\text{train} = \{x_1,\dots,x_{n_\texttt{task}}\}$, let us recall that an associated predicted probability distribution is provided by $\sigma(x)\in \Delta_{K-1}$.
Let us split the prediction interval $[0,1]$ into $M=15$ bins $I_1,\dots,I_M$ of size $1/M$: $I_m=(\tfrac{m-1}{M}, \tfrac{m}{M}]$, where $m=1,\dots,M$.
Following \citet{guo_calibration_2017}, we denote $B_m=\{x\in \gX_\text{train} :\ \sigma_{[1]}(x)\in I_m\}$ the task whose predicted probability is in the $m$-th bin\footnote{Remember that with our notation $\sigma_{[1]}(x)=\argmax_{k\in[K]} \left(\sigma(x)\right)_{k}$, with ties broken at random.}.
We recall that the accuracy of the network for the samples in $B_m$ is given by $\mathrm{acc}(B_m)$ the empirical confidence by $\mathrm{conf}(B_m)$:
\begin{align*}
    \mathrm{acc}(B_m) = \frac{1}{|B_m|} \sum_{i\in B_m} \ind_{\{\sigma_{[1]}(x_i) = y_i\}} \quad \text{ and } \quad
    \mathrm{conf}(B_m) = \frac{1}{|B_m|} \sum_{i\in B_m} \sigma_{[1]}(x_i)\enspace.
\end{align*}
Finally, the expected calibration error ($\mathrm{ECE}$) reads:
\begin{align}\label{eqapp:calibration}
    \mathrm{ECE}=\sum_{m=1}^M \frac{|B_m|}{n_\texttt{task}} \left| \mathrm{acc}(B_m) - \mathrm{conf}(B_m)\right| \enspace.
\end{align}
A neural network is said \emph{perfectly calibrated} if $\mathrm{ECE}=0$, thus if the accuracy equals the confidence for each subset $B_m$.

\section{Datasets description}
\label{sec:datasets}
\subsection{Synthetic dataset}
\label{subsec:Synthetic_dataset}

In this section, we present simulated datasets to showcase the specificities and possible limitations of the $\mathrm{WAUM}$. Here is a summary of the experiments detailed in the following sub-sections:
\begin{enumerate}
    \item \textbf{The \texttt{three\_circles} dataset}: we explain further how the simulations in \Cref{sec:experiments} were conducted
    \item \textbf{The \texttt{two\_moons} dataset}: we showcase a setting where the ambiguous tasks should be kept and not pruned. No simulated worker was able to get past the intrinsic difficulty of the dataset.
    \item \textbf{The \texttt{make\_classication\_many\_workers} dataset}: we showcase a setting with many workers and few labels per task. In this case, it is more relevant to consider the $\mathrm{WAUM}$ instead of the $\textrm{worker-wise WAUM}$.
\end{enumerate}

\subsubsection{The \texttt{three\_circles} dataset}
\label{subsubsec:3_circles_dataset}

This dataset was presented in \Cref{sec:experiments}, we give additional details here.
We simulate three cloud points using \texttt{scikit-learn}'s function \texttt{two\_circles}. Each of the $n_{\texttt{task}}=525$ points represents a task. The $n_\texttt{worker}=3$ workers are standard classifiers: $w_1$ is a linear Support Vector Machine Classifier (linear SVC), $w_2$ is an SVM with RBF kernel (SVC), and $w_3$ is a gradient boosted classifier (GBM) with five estimators.
To induce more ambiguity (and avoid too similar workers), the SVC has a maximum iteration set to $1$ in the learning phase.
Other hyperparameters are set to \texttt{scikit-learn}'s default values\footnote{For instance, the squared-hinge is penalized with an $\ell^2$ regularization parameter set to $1$ for linear SVC and SVC, GBM uses as loss the multinomial deviance, and the maximum depth equals to $3$ (default).}.
Data is split between train (70\%) and test (30\%) and each simulated worker votes for each task, \ie for all $x\in\gX_\text{train}$, $|\gA(x)|=n_\texttt{worker}=3$.
The disagreement area is identified in the northeast area of the dataset as can be seen in \Cref{fig:threecircles_workers}. \Cref{tab:res3circles} also shows that pruning too little data ($\alpha$ small) or too much ($\alpha$ large) can mitigate the performance.

\subsubsection{The \texttt{two\_moons} dataset}
\label{subsubsec:2-moons}

This dataset is introduced as a case where pruning is not recommended, to illustrate the limitations of the $\textrm{worker-wise WAUM}$ method.
The \texttt{two\_moons} simulation framework showcases the difference between relevant ambiguity in a dataset and an artificial one.
This dataset is created using \texttt{make\_moons} function from \texttt{scikit-learn}.
We simulate $n_\texttt{task}=500$ points, a noise $\varepsilon=0.2$ and use a test split of $0.3$.

\begin{figure}[h]
    \centering
    \includegraphics[width=0.75\textwidth]{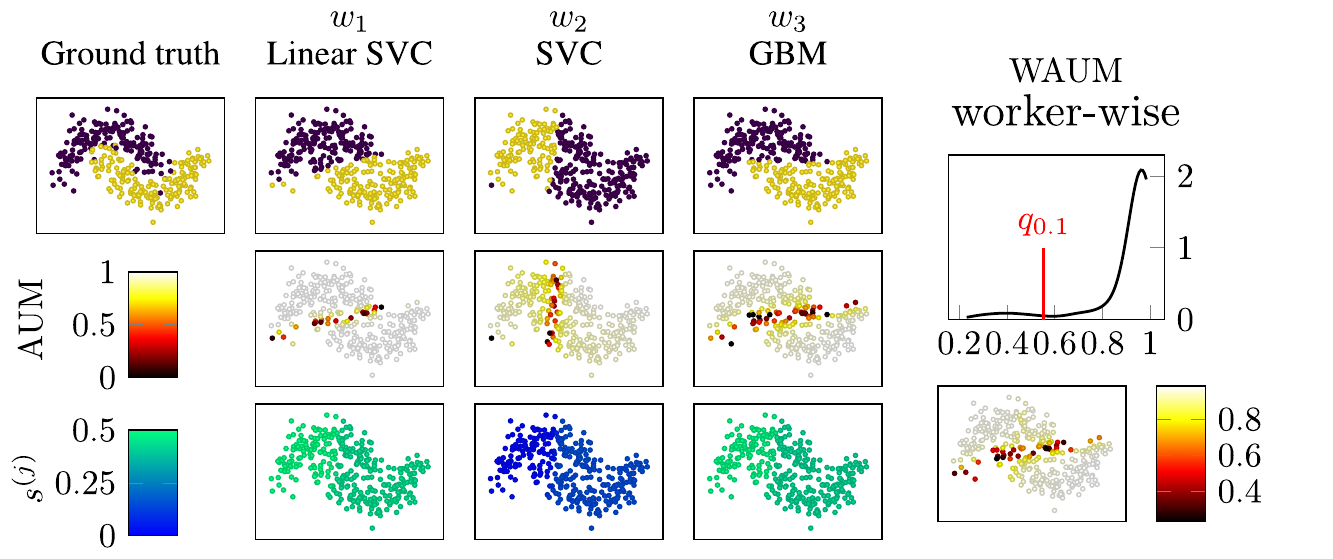}
    \caption{$\texttt{two\_moons}$ dataset: simulated workers with associated $\mathrm{AUM}$ and normalized trust scores. The hyperparameter $\alpha$ is set to $0.1$ for the $\textrm{worker-wise WAUM}$.
        Notice that the $\mathrm{SVC}$ classifier is mostly wrong (since we only train for one epoch for this worker), inducing a lower trust score overall.}
    \label{fig:2moons_workers}
\end{figure}

\begin{figure}[th!]
    \centering
    \includegraphics[width=0.75\textwidth]{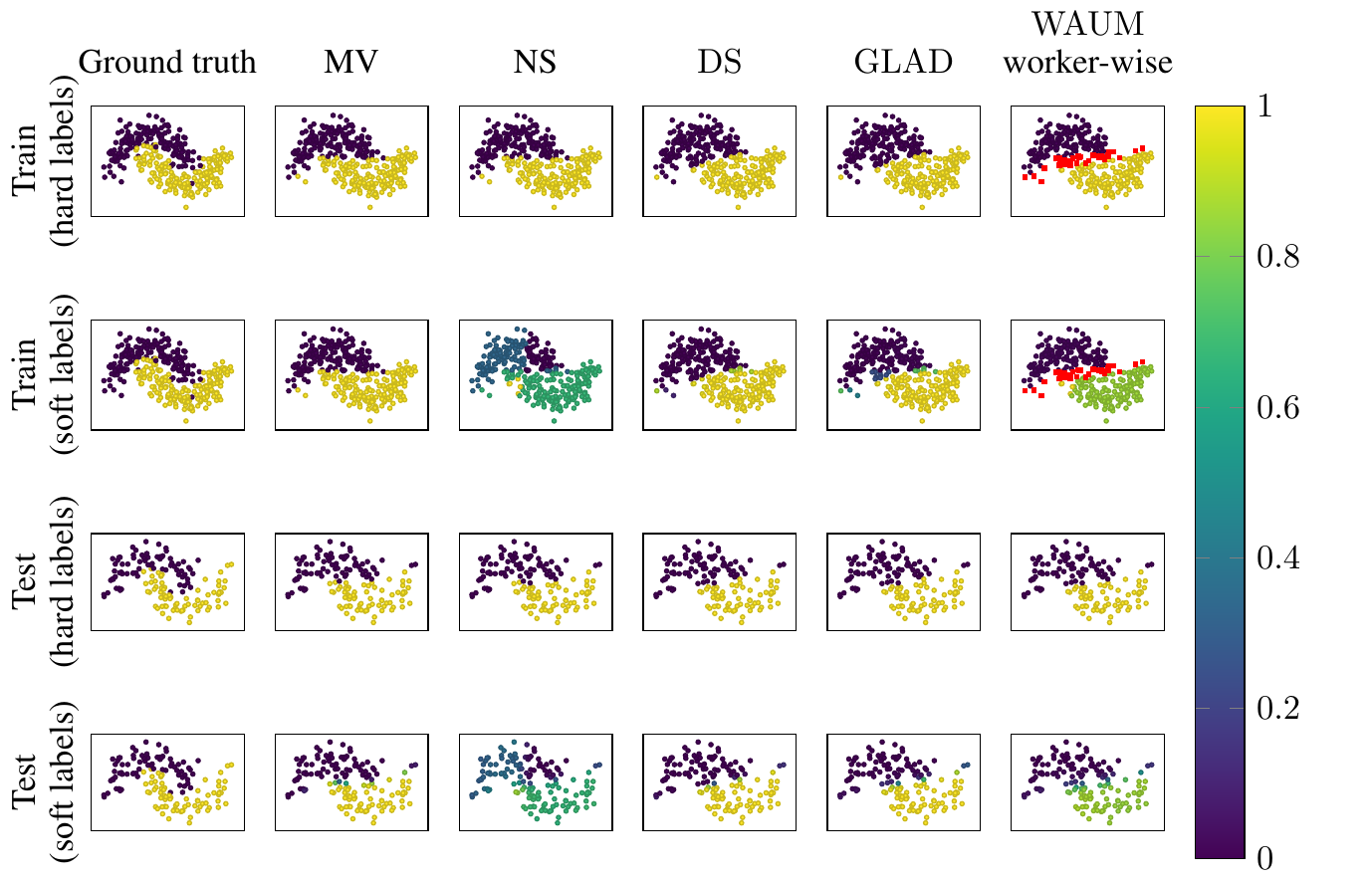}
    \caption{ \texttt{two\_moons} dataset: One realization of \Cref{tab:results_simu_moons} varying the aggregation strategy. Label predictions on train/test sets provided by a three dense layers' artificial neural network $(30, 20, 20)$ trained on smooth labeled obtained by after aggregating the crowdsourced labels (as in \Cref{fig:2moons_workers}).
        Points in red are pruned from the training set in the $\textrm{worker-wise WAUM}$ aggregation.
        The $\alpha$ hyperparameter is set to $0.1$. Each point represents a task $x_i$, and its color is the probability to belong in class $1$.
        One can visualize the ambiguity in the soft training aggregated labels, but also in the resulting predictions by the neural network.}
    \label{fig:twomoons_predictions_soft}
\end{figure}

\begin{table}[th!]
    \centering
    \caption{Training and test accuracy depending on the aggregation method used for the \texttt{two\_moons}'s dataset with $n_\texttt{task}=500$ points used for training a three dense layers' artificial neural network $(30, 20, 20)$. For reference, the best worker is $w_3$ with a training accuracy of $0.923$ and a test accuracy of $0.900$. }
        \begin{tabular}{lccc}
            Aggregation                                  & $\mathrm{Acc}_{\text{test}}$ & ECE
            \\ \hline \\
            MV                                                           & $\mathbf{0.894\pm 0.002}$    & $\mathbf{0.098}\pm 0.004$ \\
            NS                                                             & $0.887\pm 0.002$    & $0.217\pm 0.010$          \\
            DS                                                         & $0.867\pm 0.000$             & $0.126\pm 0.001$          \\
            $\mathrm{GLAD}$                                                  & $0.872\pm 0.006$             & $0.107\pm 0.004$          \\
            worker-wise $\mathrm{WAUM}(\alpha=10^{-3}) $                     & $0.875\pm 0.002$             & $0.088\pm 0.012$ \\
            worker-wise $\mathrm{WAUM}(\alpha=10^{-2})$                  & $0.874\pm 0.002$             & $0.092\pm 0.011$ \\
            worker-wise $\mathrm{WAUM}(\alpha=10^{-1})$                    & $0.870\pm 0.003$             & $0.101\pm 0.020$          \\
            worker-wise $\mathrm{WAUM}(\alpha=0.25)$                     & $0.829\pm 0.006$             & $0.135\pm 0.011$          \\
        \end{tabular}
    \label{tab:results_simu_moons}
\end{table}
As can be observed with \Cref{fig:2moons_workers} and \Cref{fig:twomoons_predictions_soft}, the difficulty of this dataset comes from the two shapes leaning into one another.
However, this intrinsic difficulty is not due to noise but is inherent to the data.
In this case, removing the hardest tasks means removing points at the edges of the crescents, and those are important in the data's structure.
From \Cref{tab:results_simu_moons}, we observe that learning on naive soft labeling leads to better performance than other aggregations.
But with these workers, no aggregation produced labels capturing the shape of the data.

\subsubsection{The \texttt{make\_classification\_many\_workers} dataset}
\label{subsubsec:make_classification_multiclass}

We simulate $n_w=150$ workers who answer tasks from a dataset with $K=4$ classes simulated using \texttt{scikit-learn}'s function \texttt{make\_classification}.
In this setting, the $\mathrm{WAUM}$ has the same performance as the $\textrm{worker-wise WAUM}$, with a \textbf{much lower computational cost} (as we do not train $n_\textrm{worker}$ networks but a single one).
All simulated tasks are labeled by up to five workers among Linear SVCs, SVCs or Gradient Boosted Classifiers (GBM) chosen uniformly.
To simulate multiple workers with some dissimilarities, we
randomly assign hyperparameters for each classifier as follows.

Each Linear SVC has a margin $\texttt{C}$ chosen in a linear grid of $20$ points from $10^{-3}$ to $3$, a maximum number of iterations between $1$ and $100$, and either \texttt{hinge} or \texttt{squared\_hinge} as $\texttt{loss}$ function.
Each SVC has a \texttt{poly} (with degree $3$), \texttt{rbf} or \texttt{sigmoid} kernel and a maximum number of iterations between $1$ and $100$.
Finally, each GBM has a learning rate of $0.01$, $0.1$ or $0.5$, a given number of base estimators in $\{1, 2, 5, 10, 15, 20, 30, 50, 100\}$ and a maximum number of iterations between $1$ and $100$.
All simulated workers are also initialized using different seeds.
All hyperparameters are drawn uniformly at random from their respective set of possible values.

\begin{table}[ht]
    \centering
        \centering
        \caption{The \texttt{make\_classification\_many\_workers} dataset: Performance metrics by aggregation method. The number of tasks is $n_\texttt{task} = 250$ tasks per classes and $1\leq|\gA(x)|\leq 5$.}
            \begin{tabular}{lccc}
                Aggregation                                 &  $\mathrm{Acc}_{\text{test}}$ & ECE                       \\ \hline \\
                NS                                                            & $0.851\pm 0.00$              & $0.146\pm 0.023$          \\
                $\mathrm{DS}$                                                   & $0.849\pm 0.004$             & $0.242\pm 0.011$          \\
                $\mathrm{GLAD}$                                                & $0.842\pm 0.002$             & $0.196\pm 0.004$          \\
                $\textrm{worker-wise WAUM}(\alpha=10^{-1})$                      & $0.849\pm0.006 $             & $\mathbf{0.137}\pm 0.034$ \\
                $\mathrm{WAUM}(\alpha=10^{-1})$                             & $\mathbf{0.861}\pm 0.007$    & $0.156\pm 0.023$
            \end{tabular}
        \label{tab:res_4classes_upto5}
\end{table}

\subsection{Real datasets}
\label{subsec:Real_datasets}

\begin{figure}[th]
    \centering
    \subfloat[Feedback effort per task distribution]{\includegraphics[trim={0 0 0 0.8cm},clip,width=0.3\textwidth]{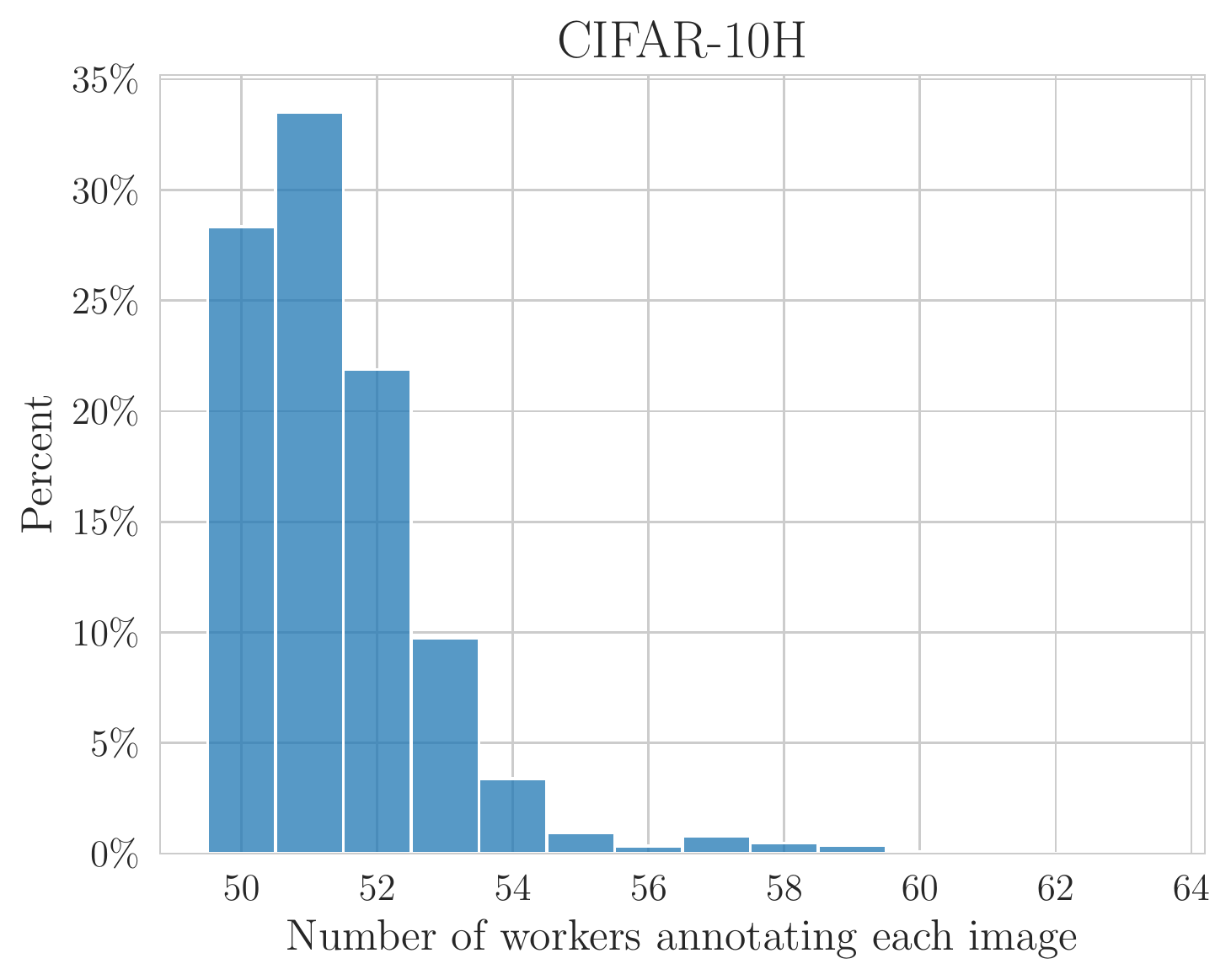}\label{subfig:cifar10h_feedback_effort}}
    \subfloat[Load per worker distribution]{\includegraphics[trim={0 0 0 0.8cm},clip,width=0.3\textwidth]{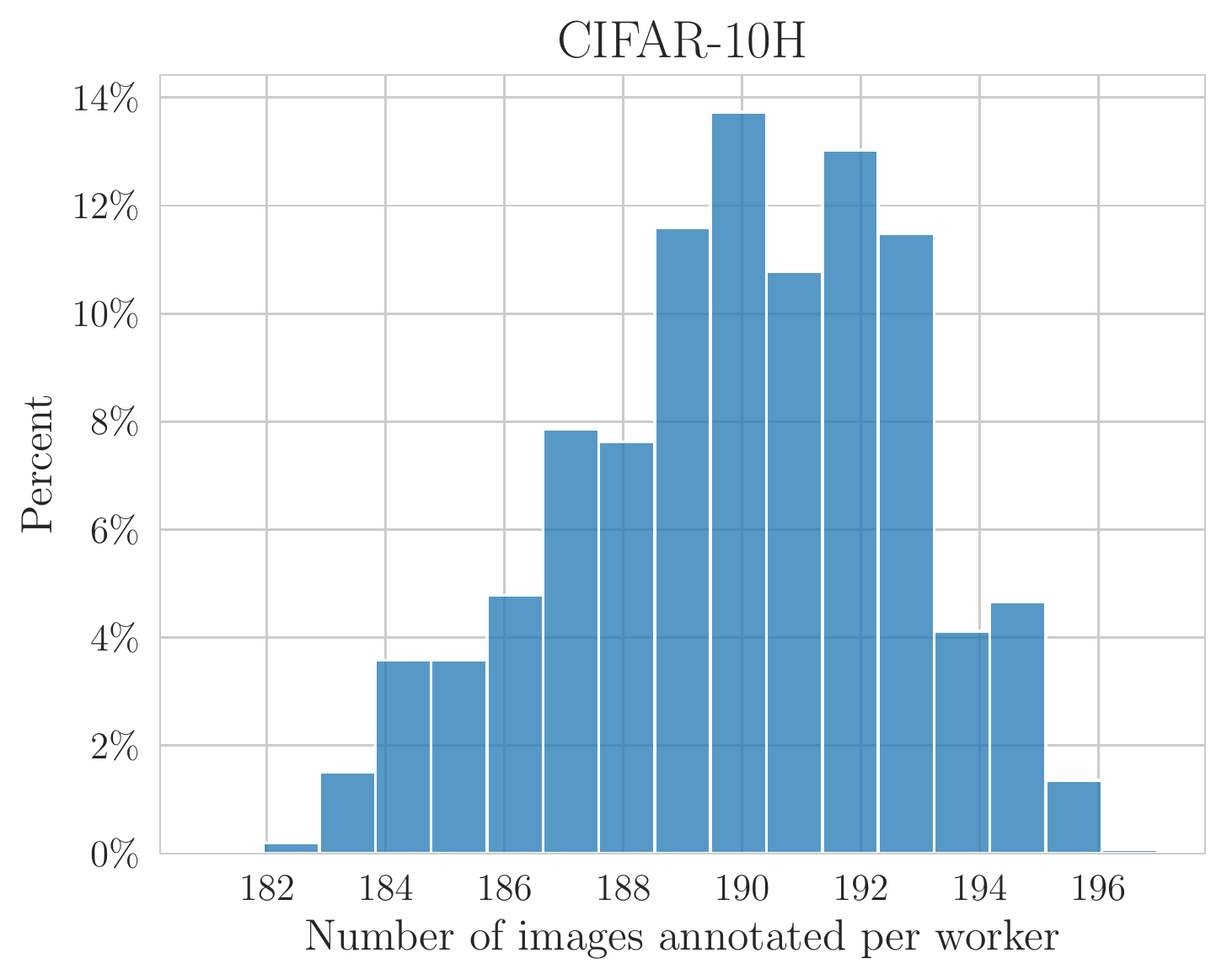}\label{subfig:cifar10h_worker_load}}
    \subfloat[Naive soft labels, entropy distribution]{\includegraphics[trim={0 0 0 0.8cm},clip,width=0.3\textwidth]{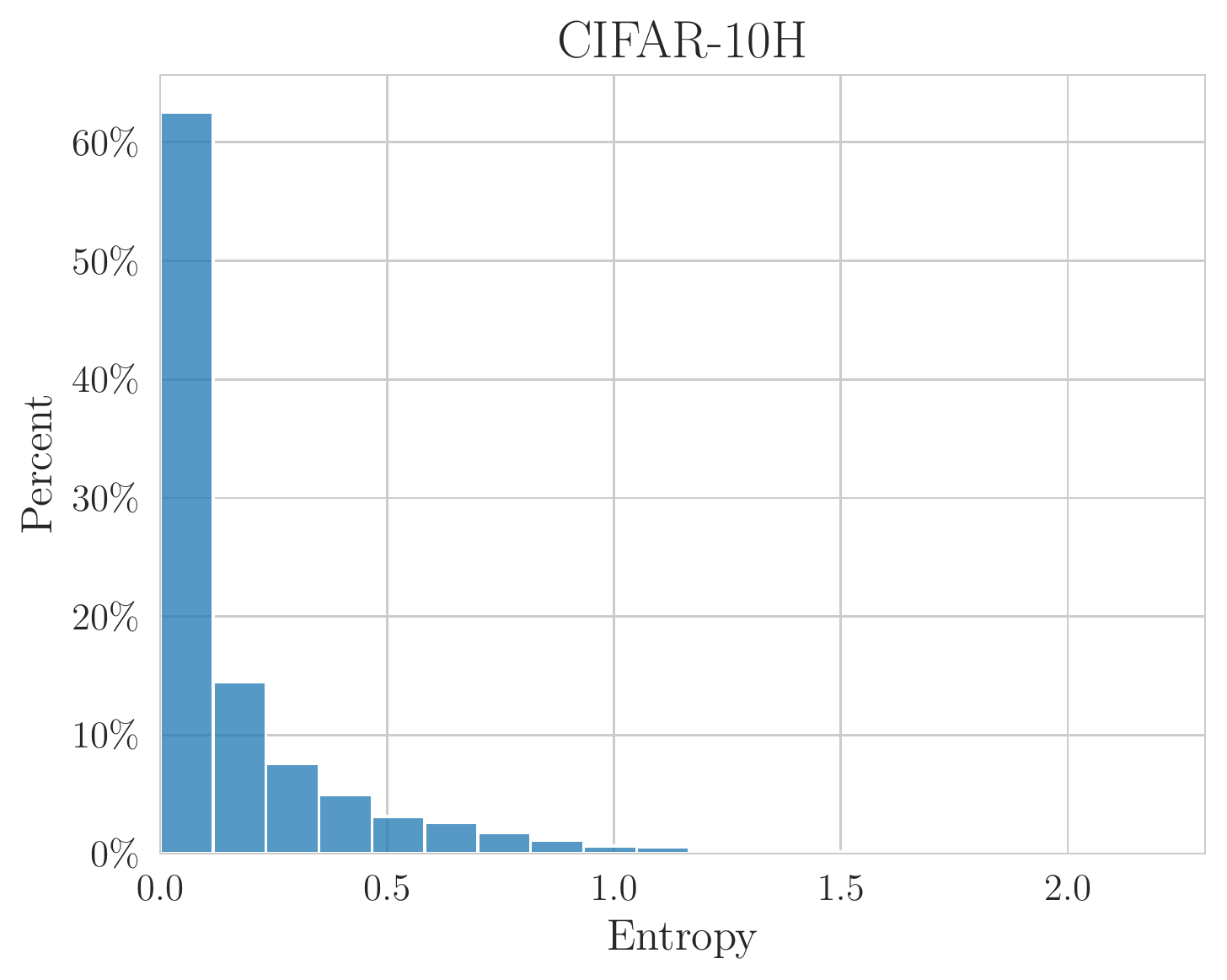}\label{subfig:cifar10h_entropy}}
    \caption{\texttt{CIFAR-10H}: dataset visualization}
    \label{fig:CIFAR-10H_dataset_visualization}
\end{figure}

\begin{figure}[th!]
    \centering
    \subfloat[Feedback effort per task]{\includegraphics[trim={0 0 0 0.8cm},clip,width=0.3\textwidth]{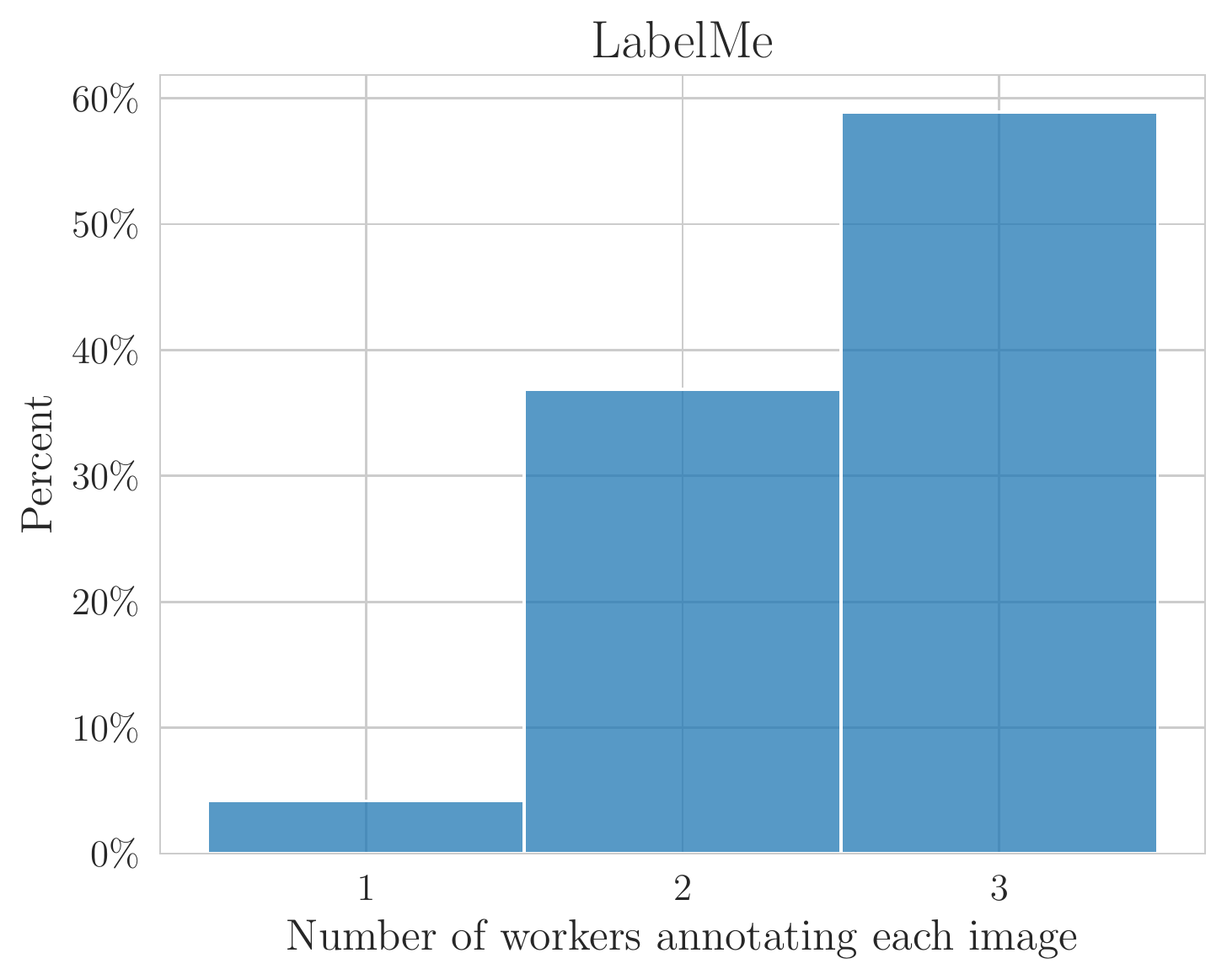}\label{subfig:labelme_feedback_effort}}
    \subfloat[Load per worker distribution]{\includegraphics[trim={0 0 0 0.8cm},clip,width=0.3\textwidth]{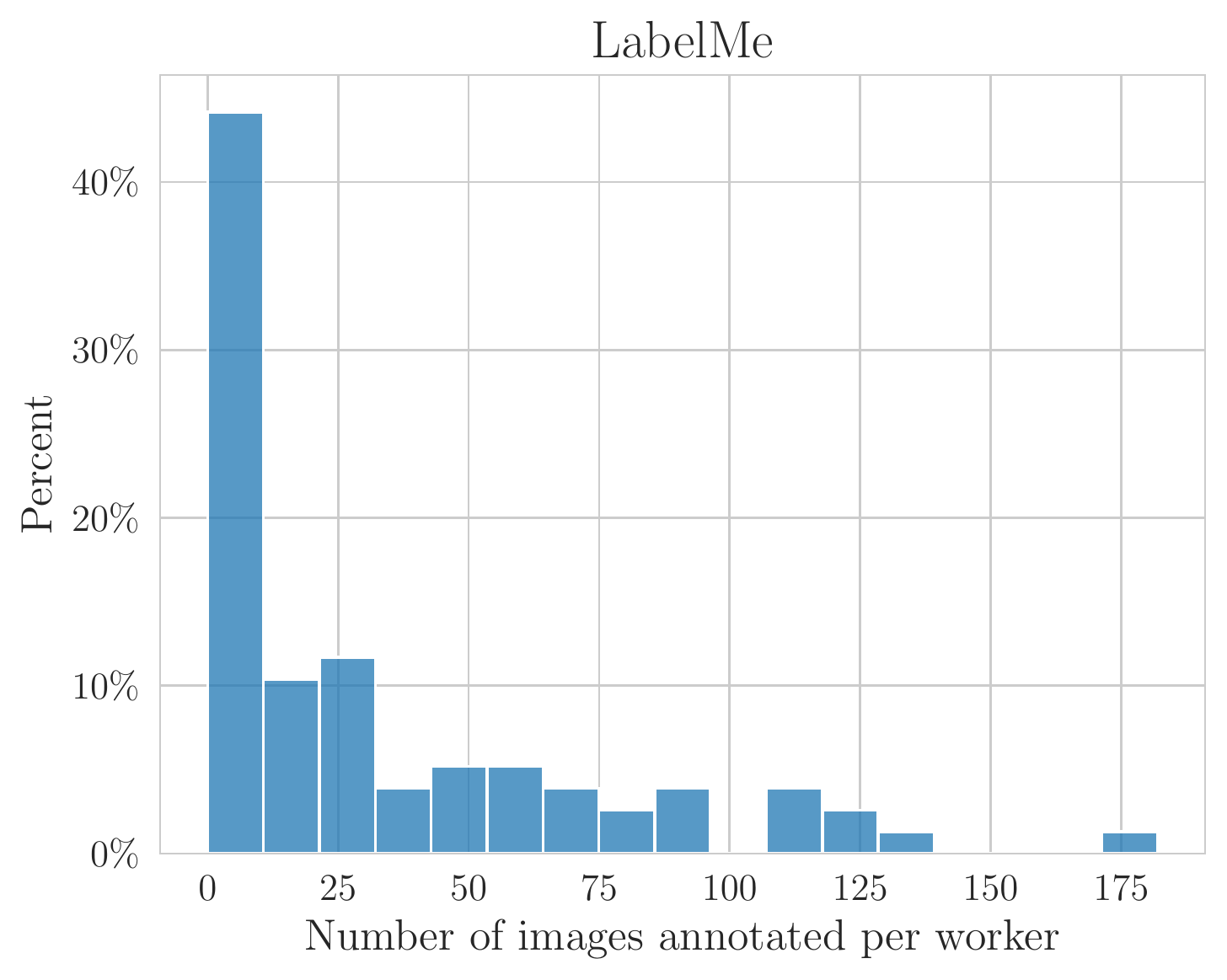}\label{subfig:labelme_worker_load}}
    \subfloat[Naive soft labels, entropy distribution]{\includegraphics[trim={0 0 0 0.8cm},clip,width=0.3\textwidth]{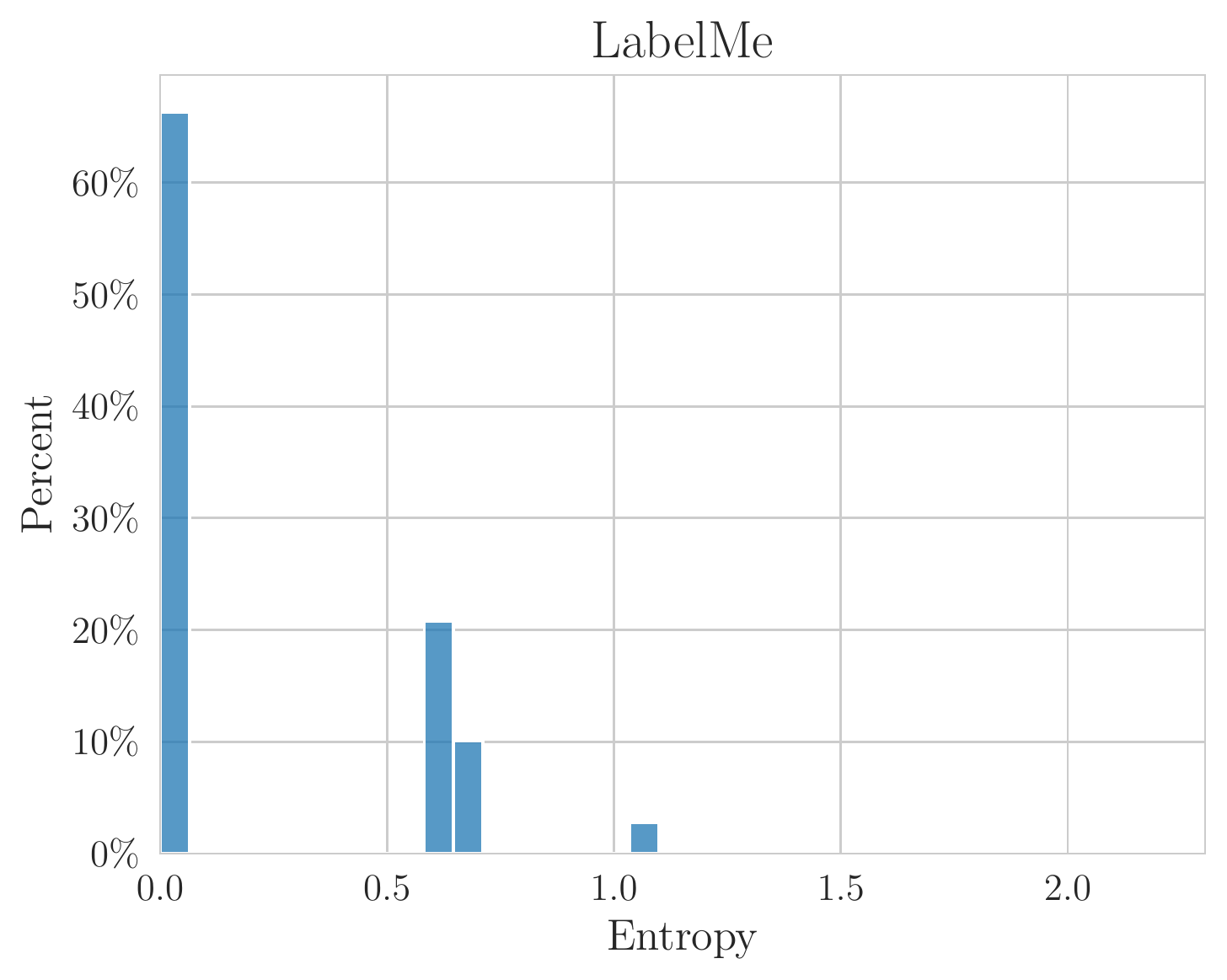}\label{subfig:labelme_entropy}}
    \caption{\texttt{LabelMe}: dataset visualization}
    \label{fig:LabelMe_dataset_visualization}
\end{figure}

\begin{figure}[th!]
    \centering
    \subfloat[Feedback effort per task distribution]{\includegraphics[trim={0 0 0 0.8cm},clip,width=0.3\textwidth]{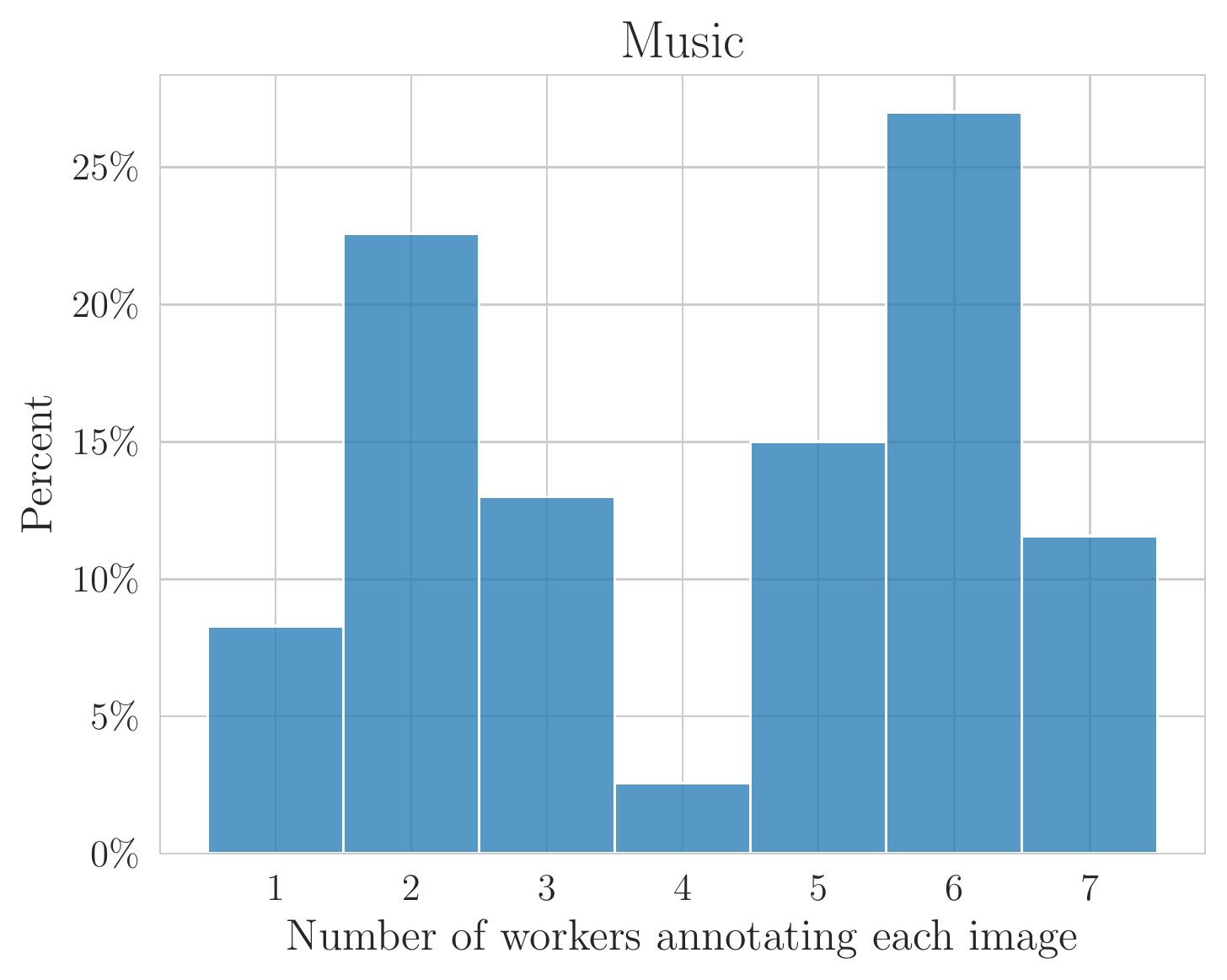}\label{subfig:music_feedback_effort}}
    \subfloat[Load per worker distribution]{\includegraphics[trim={0 0 0 0.8cm},clip,width=0.3\textwidth]{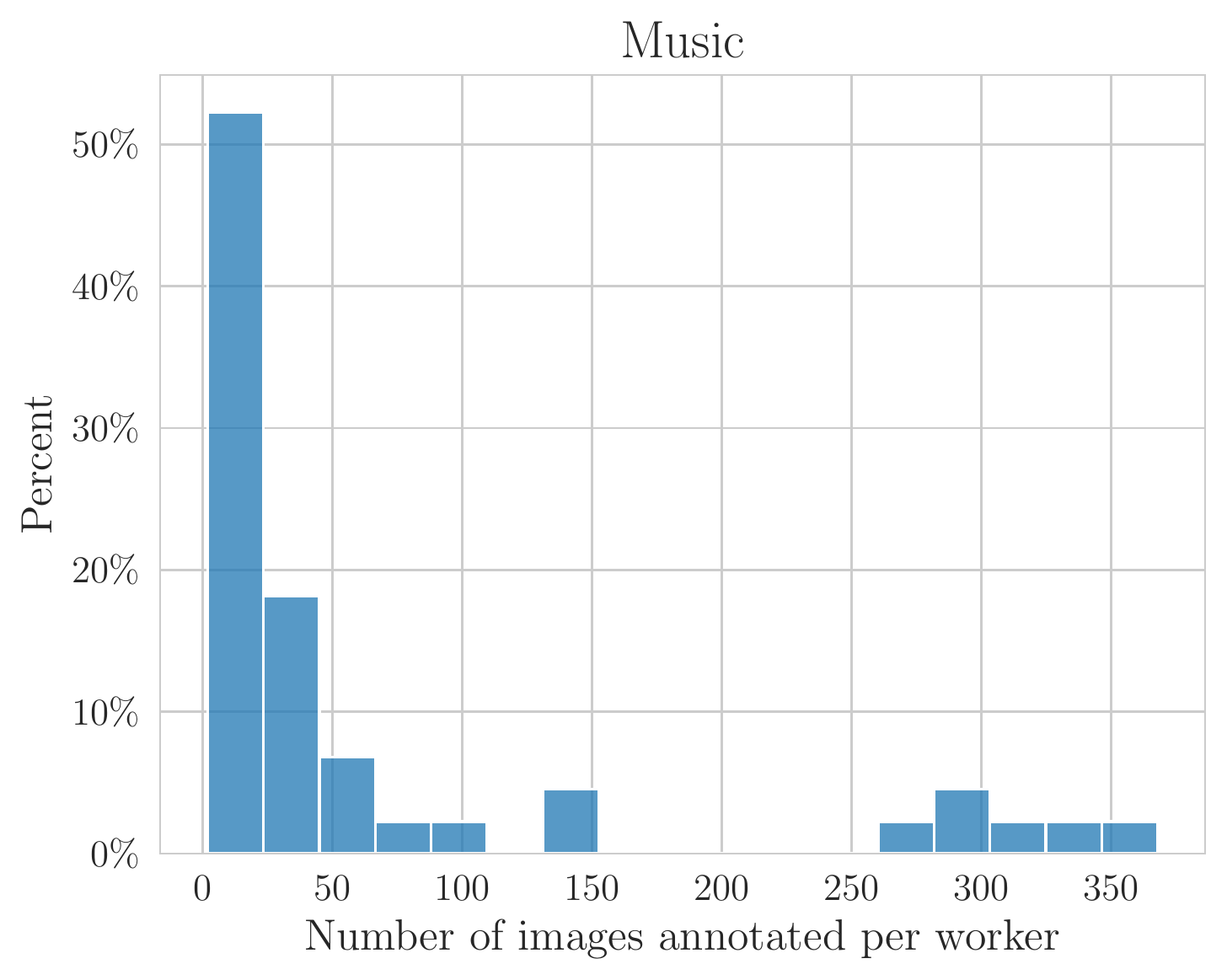}\label{subfig:music_worker_load}}
    \subfloat[Naive soft labels, entropy distribution]{\includegraphics[trim={0 0 0 0.8cm},clip,width=0.3\textwidth]{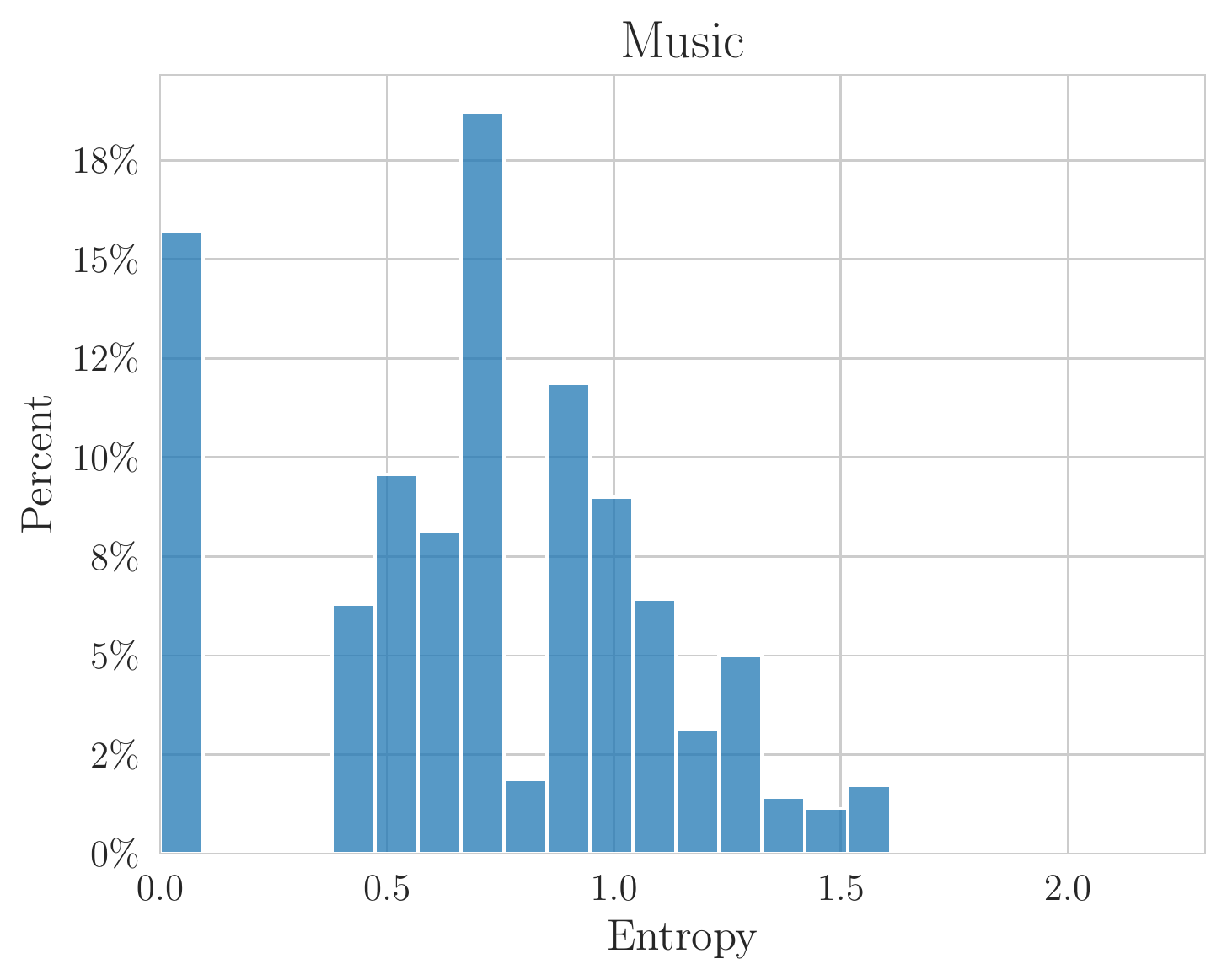}\label{subfig:music_entropy}}
    \caption{Music: dataset visualization}
    \label{fig:Music_dataset_visualization}
\end{figure}

The datasets we consider are all decomposed into three parts: train $(\mathcal{D}_{\text{train}})$, validation $(\mathcal{D}_{\text{val}})$, and test $(\mathcal{D}_{\text{test}})$.
They are described in the following subsections.
In particular, we provide for the training set of each dataset (see \Cref{fig:CIFAR-10H_dataset_visualization,fig:LabelMe_dataset_visualization,fig:Music_dataset_visualization}) three visualizations: the feedback effort per task distribution ($|\gA(x)|$), the load per worker distribution ($|\gW(x)|$), and the naive soft labels entropy distribution, \ie the entropy distribution for each task in the training set, defined by:
$\forall x_i \in\gX_\text{train}    ,\ \mathrm{Ent}(x_i) = -\sum_{k\in[K]}  (\hat{y}_i^{\textrm{NS}})_k\log((\hat {y}_i^{\textrm{NS}})_k)$.

We have conducted experiments on three real datasets.
The \texttt{CIFAR-10H} dataset has been proposed to reflect human perceptual uncertainty in (a subpart of) the classical \texttt{CIFAR-10} dataset.
Each worker has annotated a large number of (seemingly easy) tasks, thus leading to few disagreements.
The \texttt{LabelMe} and \texttt{Music} datasets have very few votes per task, leading to more ambiguous votes distributions.

\subsubsection{The \texttt{CIFAR-10H} dataset}
\label{subsubsec:cifar-10h_dataset}

Introduced by \citet{peterson_human_2019}, the crowdsourced dataset \texttt{CIFAR-10H} attempts to recapture the human labeling noise present when creating the dataset.
We have transformed this dataset, mainly by creating a validation set.
Hence, the training set for our version of \texttt{CIFAR-10H} consists of the first $9500$ test images from \CIFAR, hence
$|\mathcal{D}_{\text{train}}| = 9500$.
The validation set is then composed of the last $500$ images from the training set of \texttt{CIFAR-10} meaning  $|\mathcal{D}_{\text{test}}| = 500 $.
The test set consists of the whole training set from \CIFAR, so  $|\mathcal{D}_{\text{test}}| = 50000$.
The crowdsourcing experimentation involved $n_\texttt{worker}=2571$ workers on Amazon Mechanical Turk.
Workers had to choose one label for each presented image among the $K=10$ labels of \CIFAR: \texttt{airplane}, \texttt{automobile}, \texttt{bird}, \texttt{cat}, \texttt{deer}, \texttt{dog}, \texttt{frog}, \texttt{horse}, \texttt{ship} and \texttt{truck}.
Each worker labeled $200$ tasks (and was paid $\$1.50$ for that): $20$ for each original category.
Answering time was also measured for each worker\footnote{Note that attention checks occurred every $20$ trial for each worker, for tasks whose labels were known.
They have been removed from the dataset since the corresponding images are not available.}.
The \texttt{CIFAR-10H} annotating effort is balanced: each task has been labeled by $50$ workers on average.

\begin{figure}[th]
    \centering
    \subfloat[$\mathrm{WAUM}$ crowdsourced identification]{\includegraphics[width=0.3\linewidth]{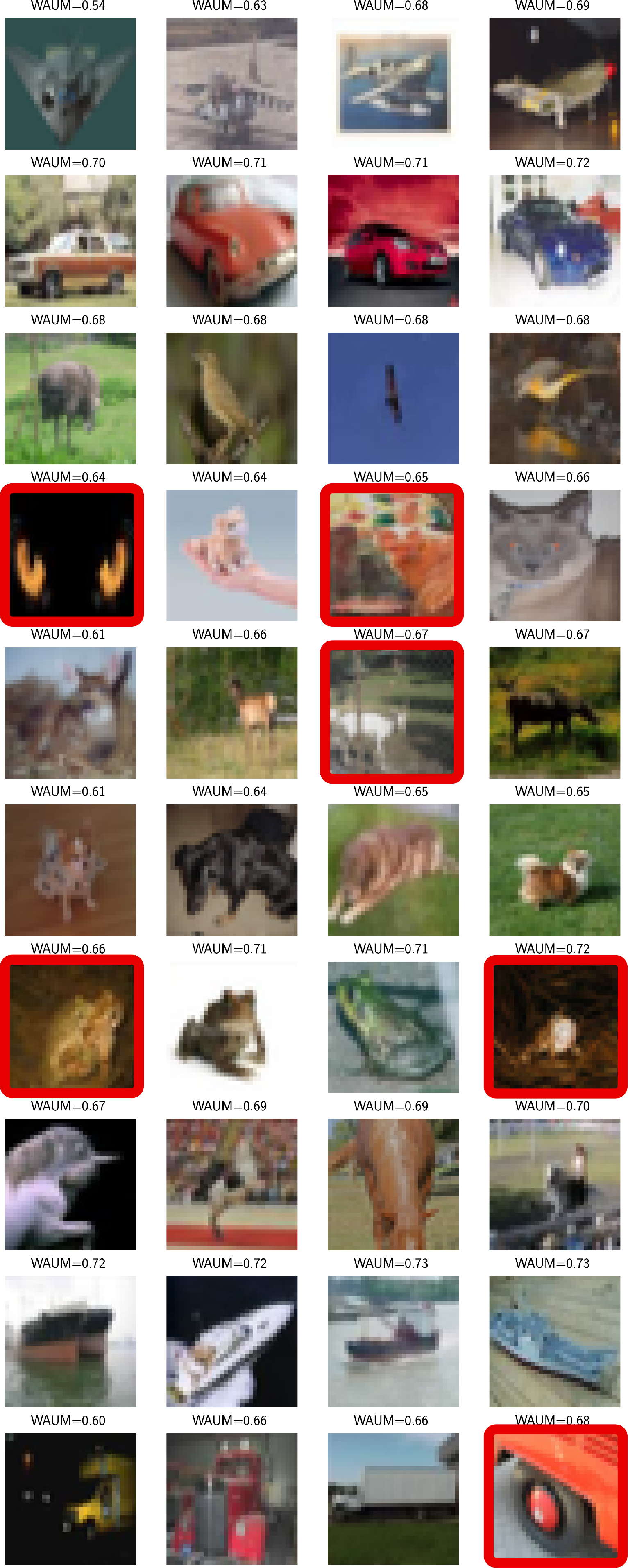}}
    \hfill
    \subfloat[$\mathrm{AUMC}$ crowdsourced identification] {\includegraphics[width=0.3\linewidth]{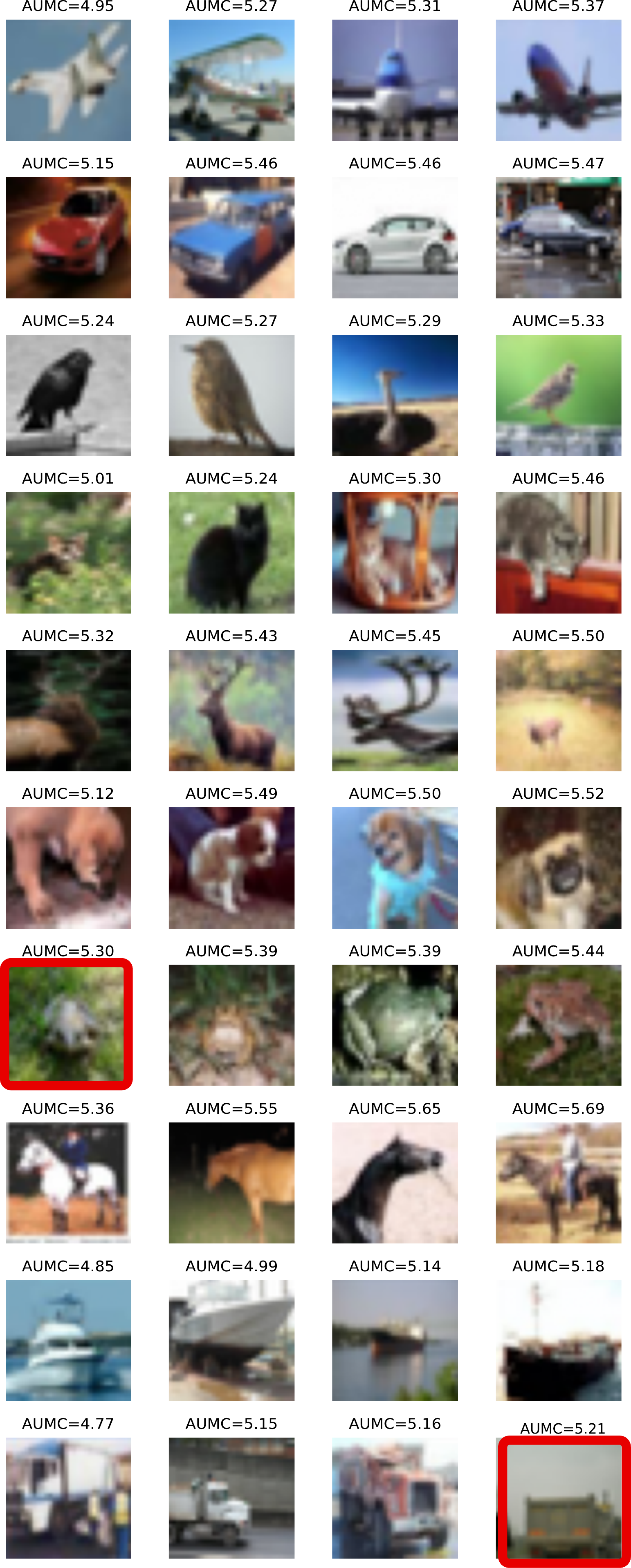}}
    \hfill
    \subfloat[$\mathrm{AUM}$ ground truth identification]{\includegraphics[width=0.3\linewidth]{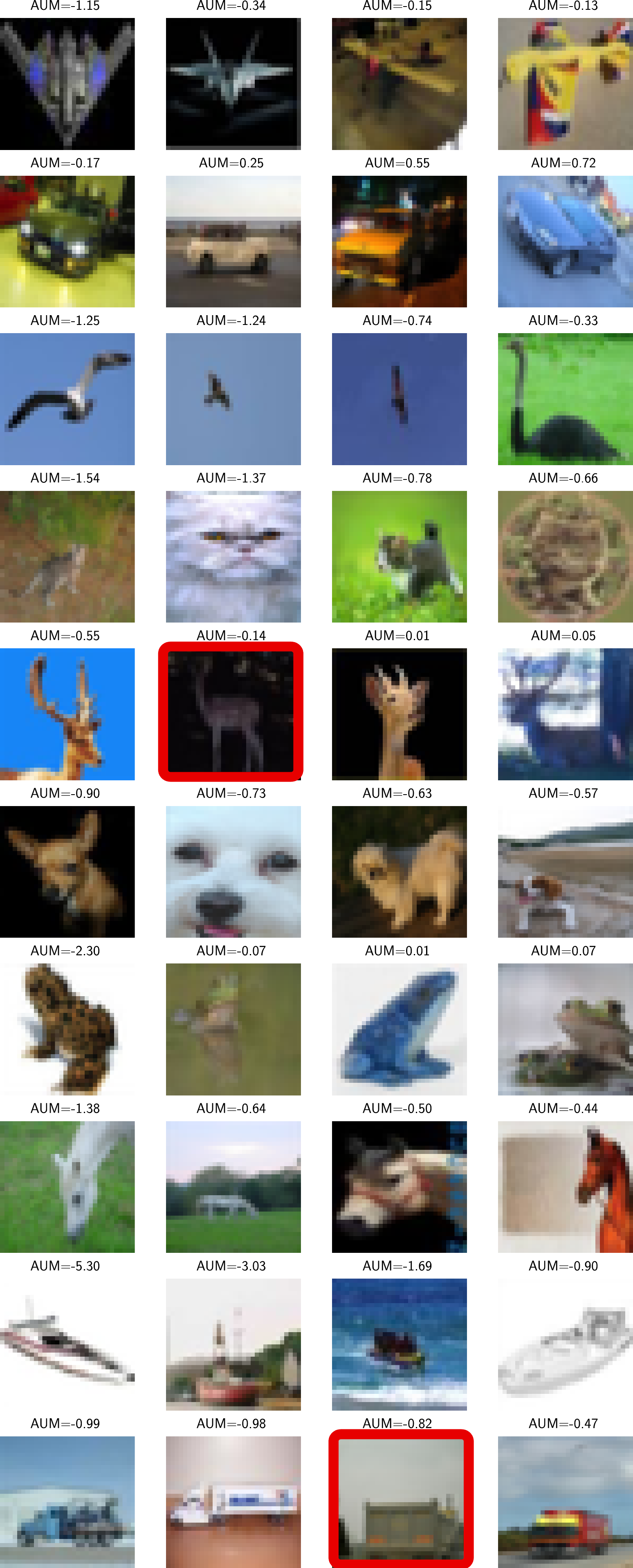}}
    \caption{Comparison of the worse images detected by the $\mathrm{WAUM}$, $\mathrm{AUMC}$ and classical $\mathrm{AUM}$ preprocessing step. Identification was computed with a ResNet-18 for $50$ epochs using the parameters described in \Cref{sec:experiments}. Each row represents the class given by the unobserved ground truth label from the \texttt{CIFAR-10} dataset. Only the $\mathrm{AUM}$ uses the ground truth label, other methods are based on the crowdsourced labels only. Images framed in red can be hard to classify.}
    \label{fig:comparison_waums_aumc}
\end{figure}

\subsubsection{The \texttt{LabelMe} dataset}
\label{subsubsec:labelme}

Another real dataset in the crowdsourced image classification field that can be used is the \texttt{LabelMe} crowdsourced dataset created by \citet{rodrigues2018deep}.
This dataset consists of $n_\text{task}=1000$ training images dispatched among $K=8$ classes:
\texttt{highway}, \texttt{insidecity}, \texttt{tallbuilding}, \texttt{street}, \texttt{forest}, \texttt{coast}, \texttt{mountain} or \texttt{open country}.
The validation set has $500$ images and the test set has $1188$ images.
The whole training tasks have been labeled by $n_\texttt{worker}=59$ workers, each task having between one and three given (crowdsourced) labels.
In particular, $42$ tasks have been labeled only once, $369$ tasks have been labeled twice and $589$ received three labels.
This is a way sparser labeling setting than the \texttt{CIFAR-10H} dataset.

Also, note that the \texttt{LabelMe} dataset has classes that overlap and thus lead to intrinsic ambiguities.
This is the reason why the CoNAL strategy was introduced by \citet{chu2021learning}, see details in \Cref{subsec:conal}.
For example, the classes \texttt{highway}, \texttt{insidecity}, \texttt{street} and \texttt{tallbuilding} (in rows) are overlapping for some tasks:
some cities have streets with tall buildings, leading to confusion as shown in \Cref{fig:worse_labelme}.
The proposed feature aware aggregation using the $\mathrm{WAUM}$ leads to better performance in test accuracy and calibration as illustrated in \Cref{tab:labelme}.

\subsubsection{The \texttt{Music} dataset}
\label{subsubsec:music}

\citet{rodrigues2014gaussian} released a crowdsourced dataset of audio files.
The goal of this classification task was to decide the genre of $30$ seconds musical excerpts. Number of tasks is $n_{\texttt{task}}=700$. 
The $n_{\texttt{worker}}=44$ workers had $K=10$ possible labels: \texttt{blues}, \texttt{classical}, \texttt{country}, \texttt{disco}, \texttt{hiphop}, \texttt{jazz}, \texttt{metal}, \texttt{pop} and \texttt{reggae}.
Each audio file was labeled by between $1$ and $7$ workers.
To test the results, a dataset of $299$ labeled clips is used (originally $300$, but one file is known to be corrupted).
Instead of working with the original audio files, we have used Mel spectrograms, openly available\footnote{ \scriptsize \url{https://www.kaggle.com/datasets/andradaolteanu/gtzan-dataset-music-genre-classification?datasetId=568973}}, to rely on standard neural networks architecture for image classification.

\section{Algorithmic details on the neural network training}
\label{sec:Details_on_the_neural_network}
Experiments can be reproduced using the code available at \url{https://github.com/peerannot/peerannot} from the \texttt{peerannot} library, which is briefly described below:
\begin{itemize}
\item The \texttt{identification} module is used to explore datasets tasks and workers. Tasks can be explored thanks to the entropy of the label distribution, the $\mathrm{WAUM}$ or the $\mathrm{AUMC}$. Workers can be evaluated thanks to the Spam-score of \citet{raykar_ranking_2011}, the trace of the DS estimated matrices, GLAD's parameters among other.
\item The \texttt{aggregate} module is used to produce aggregated labels from multiple answered labels. The labels can then be used for training a neural network architecture from \texttt{Pytorch} using the \texttt{train} module.
\item The \texttt{aggregate-deep} module is used for the $\mathrm{CoNAL}$ and $\mathrm{CrowdLayer}$ strategies. A neural network is directly learning from the crowdsourced tasks and labels without the aggregation step.
\item Multiple datasets are ready to use, including \texttt{CIFAR-10H}, \texttt{LabelMe} and \texttt{Music}.
\end{itemize}
The documentation of the library is at \url{https://peerannot.github.io/}.
\end{document}